\documentclass[preprint,12pt]{elsarticle}

\usepackage{amssymb}
\usepackage{amsmath,amsfonts}
\usepackage{algorithmicx}
\usepackage{algpseudocode}
\usepackage{algorithm}
\usepackage{array}
\usepackage{adjustbox}
\usepackage{mathptmx}
\usepackage{amsmath,amssymb}
\usepackage{xcolor}
\usepackage{booktabs}
\usepackage[caption=false,font=normalsize,labelfont=sf,textfont=sf]{subfig}
\usepackage{textcomp}
\usepackage{stfloats}
\usepackage{url}
\usepackage{verbatim}
\usepackage{graphicx}
\usepackage{pdfpages}
\newtheorem{theorem}{Theorem}

\journal{---}

\begin{document}

\begin{frontmatter}



\title{Multi-Label Feature Selection Using Adaptive and Transformed Relevance}


\author[inst1]{Sadegh Eskandari}

\affiliation[inst1]{organization={Department of Computer Science},
            addressline={University of Guilan, Rasht, Iran}}

\author[inst2]{Sahar Ghassabi}

\affiliation[inst2]{organization={Department of Computer Engineering},
            addressline={Ferdowsi University of Mashhad,
Mashhad , Iran}}

\begin{abstract}

Multi-label learning has emerged as a crucial paradigm in data analysis, addressing scenarios where instances are associated with multiple class labels simultaneously. With the growing prevalence of multi-label data across diverse applications, such as text and image classification, the significance of multi-label feature selection has become increasingly evident. This paper presents a novel information-theoretical filter-based multi-label feature selection, called ATR, with a new heuristic function.  Incorporating a combinations of algorithm adaptation and problem transformation approaches, ATR ranks features considering individual labels as well as abstract label space discriminative powers. Our experimental studies encompass twelve benchmarks spanning various domains, demonstrating the superiority of our approach over ten state-of-the-art information-theoretical filter-based multi-label feature selection methods across six evaluation metrics. Furthermore, our experiments affirm the scalability of ATR for benchmarks characterized by extensive feature and label spaces. \\
The codes are available at \textcolor{blue}{https://github.com/Sadegh28/ATR}  
\end{abstract}
\newpage
\begin{graphicalabstract}
\fbox{\includegraphics[width=1.\textwidth]{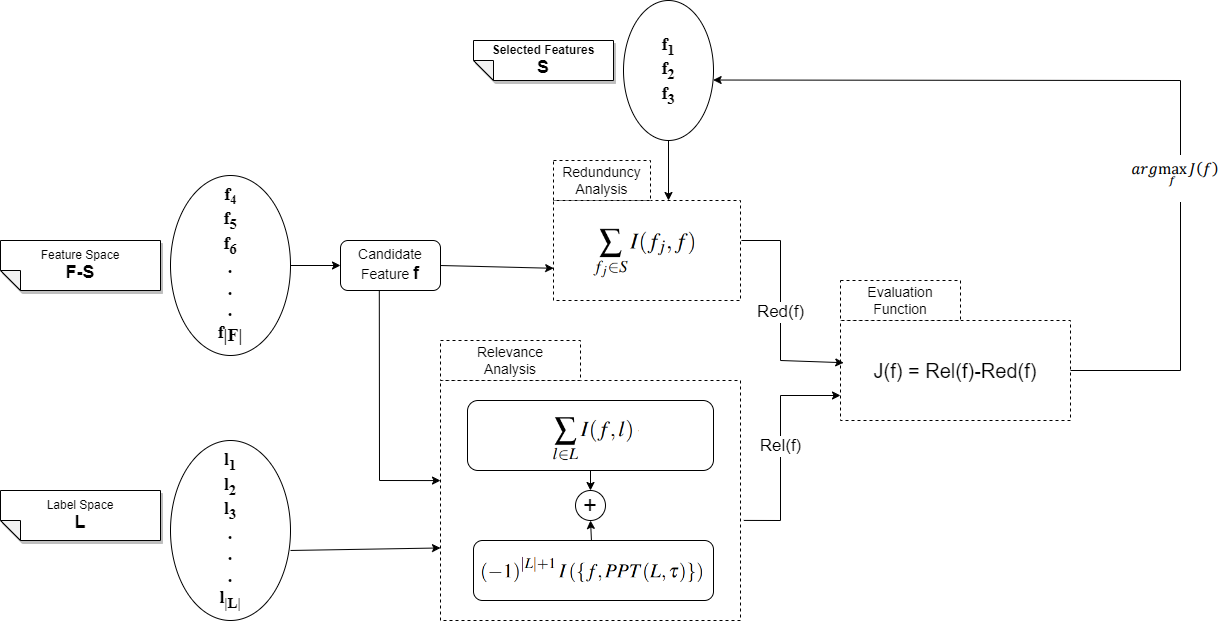}}
\end{graphicalabstract}



\begin{keyword}
Multi-label Learning \sep Multi-Label Feature Selection \sep Feature Selection \sep Mutual Information \sep Dimension Reduction 
\end{keyword}

\end{frontmatter}


\section{Introduction}
\label{sec:Introduction}
In recent years, multi-label learning has gained significant traction, finding applications in various domains such as text sentiment analysis \cite{lin2015multi, liu2015multi}, music emotion recognition \cite{yang2012machine}, and image annotation \cite{hong2013image}.  In multi-label data, an instance may simultaneously belong to more than one class label 
\cite{zhang2021multia,zhang2021multib}. For example, a tweet about covid-19 can be associated with "vaccine", "social health", "pandemic" and "technology" simultaneously.

Formally, we denote a set of instances $\mathbb{X} = \{\mathbf{x_1}, \mathbf{x_2}, \dots, \mathbf{x_M} \}$, constructed from a feature set $F = \{f_1, f_2, \dots, f_{|F|} \}$. In the context of multi-label data, each instance $\mathbf{x_i} \in \mathbb{X}$ is associated with a subset of labels $\mathbf{y_i} \subseteq L$, where $L = \{l_1, l_2, \dots, l_{|L|}\}$ represents a finite label set. Notably, the conventional binary and multi-class classification problems can be viewed as special cases of multi-label learning, with the constraint that $|\mathbf{y_i}| = 1$ for all $i = 1, 2, \dots, |F|$.

In practical multi-label datasets, a frequent scenario involves an extensive feature space encompassing numerous irrelevant or redundant features \cite{qian2021label,lv2021semi,hashemi2020mfs}. This abundance of features directly contributes to a proliferation of learning parameters, consequently diminishing the overall generalization performance of classifiers. To mitigate this concern, Multi-label Feature Selection (MLFS) emerges as a pivotal approach. MLFS aims to address this challenge by identifying an optimal subset of features, $S^* \subseteq F$, that maximizes a given evaluation measure $E$. This can be formulated mathematically as follows:
\begin{equation}\label{equation:MLFS_problem}
	S^* = arg\,min_{|S|} \left( arg\,max_{S\subseteq F}E\left( S;L  \right)  \right)
\end{equation}

MLFS algorithms are classified into three main categories, namely wrapper, embedded, and filter methods, based on the varying interpretations of the evaluation measure $E$ \cite{zhang2009feature,gharroudi2014comparison}. In the context of wrapper methods \cite{zhang2009feature,gharroudi2014comparison}, $E$ corresponds to the accuracy of a learning model. Here, for each potential feature subset, a model is trained, and the subset yielding the highest accuracy is chosen. Embedded methods \cite{yu2005multi,zhang2010multilabel,gu2011correlated,jian2016multi,huang2017joint}, on the other hand, define $E$ as a measure of a learning model's complexity, often incorporated as part of the regularization term within the loss function. Throughout model training, the objective is to strike a balance between model accuracy and complexity by selectively retaining or eliminating features. In contrast, filter methods \cite{read2008pruned,doquire2011feature,zhang2021multia,zhang2021multib,hashemi2020mfs,lee2015mutual,li2014multi,zhang2019distinguishing,lin2015multi,lee2013feature,lee2017scls,bidgoli2021reference,hu2020multi} leverage $E$ from a diverse range of model-independent criteria, including information theory, relevance, and dependency. Among these categories, filter methods have garnered particular interest due to their simplicity and lack of bias toward any specific learning model. This makes filters suitable for standalone data pre-processing, without concern for the ultimate learning task.


The primary focus of this paper is on filter-based MLFS algorithms that utilize information theory as a means to evaluate feature subsets.  This choice is substantiated by the robust mathematical and empirical foundation of information theory \cite{peng2005feature,khinchin2013mathematical,vergara2014review}. Moreover, it is shown that information theoretic criteria, such as mutual information, are able to measure any kind of dependency (linear and non-linear) between random variables \cite{dionisio2004mutual}. For the context of information-theoretic filter-based MLFS, the problem represented in Eq. \ref{equation:MLFS_problem} can be restated as:

\begin{equation}\label{equation:MLFS_problem_information}
	S^* = arg\,min_{|S|} \left( arg\,max_{S\subseteq F}I\left( S;L  \right)  \right)
\end{equation}

The main challenge in solving Eq. \ref{equation:MLFS_problem_information} lies in the computational infeasibility of evaluating all possible subsets of the feature space. To address this complexity, a prevalent approach is to reduce the feature subset selection problem into a feature ranking problem. This is achieved through the use of a heuristic function that assesses the relevance and redundancy of features. Given $S \subset F$ as a set of already evaluated features and $f \in F-S$, the general form of the heuristic function is:

\begin{equation}\label{equation:MLFS_problem_information_heuristic}
J\left(f\right) = Rel\left(f\right) - Red\left(f\right)
\end{equation}

Where, $Rel\left(f\right)$ and $Red\left(f\right)$ represent the relevance and redundancy of feature $f$, respectively. The relevance measures the contribution of $f$ in determining $L$, while the redundancy ensures that $f$ does not convey information already provided by $S$. Typically, an information-theoretical filter-based MLFS algorithm commences with an empty $S$ and iteratively updates it using the following rule:

\begin{equation}\label{equation:MLFS_problem_update_rule}
S \leftarrow S\cup { arg,max_{f\in F-S} J\left(f\right) }
\end{equation}




Several information-theoretical filter-based MLFS methods have been proposed in the literature. These methods can be categorized into three distinct groups based on the consideration of label correlations within their heuristic functions.
The initial group of methods assumes complete independence among all labels. Consequently, these methods solely rely on first-order label correlations for assessing feature relevance. While these approaches exhibit relatively swift computation, they often compromise accuracy by overlooking crucial information embedded within label subsets.
The second group encompasses methods that incorporate second-order label correlations by evaluating mutual information between pairs of labels. Generally, these techniques provide a more  precise evaluation of feature relevance. However, they necessitate the calculation of pairwise mutual information for all label combinations, leading to computational inefficiencies of $O(|L|^2)$ in scenarios with extensive label spaces.
The third  group of methods adopts an abstract perspective on the label space.  By leveraging correlations of order $|L|$, they extract information carried by the entire label set. These algorithms stand out for their remarkable computational efficiency, as each feature evaluation requires only a single multi-variate information calculation. Nonetheless, these methods face the challenge of high-dimensional joint probability estimation. This issue arises primarily due to the large label space and the limited number of available patterns. As a result, the accuracy of relevance evaluation may be degraded by this issue.

This paper introduces a novel and efficient information-theoretical filter-based MLFS approach, employing a new heuristic function. The proposed method incorporates both first-order and $|L|$th order correlations to estimate feature relevance. This allows for the consideration of within-label as well as within-label-space discriminative powers of  features in their ranking. The unique contributions that distinguish our work from existing approaches are (1) It advances the evaluation metrics in filter-based MLFS one step further. (2) A novel MLFS algorithm is introduced that combines first and $|L|$th order correlations to  accurately estimate the relevance of features.  (3) Our approach is highly scalable, addressing challenges in problems featuring both extensive feature and label spaces.

The remainder of the paper is organized as follows: In Section 2, a foundational understanding of information theory is provided, serving as a preliminary for the subsequent discussions. Related works  are also reviewed within the same section. In Section 3, our proposed MLFS algorithm is presented. In Section 4, the performance of our method is  evaluated through extensive experimental results, with comparisons made against existing approaches. Finally, in Section 5, the paper is concluded. 

\section{Preliminaries and Related Work}\label{section:Preliminaries and Related Work}
\subsection{Information Theory}\label{Information Theory}
Information theory is widely acknowledged as a reliable indicator for measuring the significance of random variables \cite{cover1999elements}. It offers effective tools for quantifying correlations between features and labels, making it widely employed in feature selection methods \cite{vergara2014review}. In this subsection, we provide a definition of the theory with a specific emphasis on entropy and mutual information.

For a discrete random variable $X$, the entropy of $X$, denoted $H(X)$, is defined as: 
\begin{equation}\label{Entropy}
	\begin{split}
		H(X) = E\left[ -\log p\left(X \right)  \right]
		&= -\sum_{x\in X}p(x)\log p(x)\\
	\end{split}
\end{equation}
$H(X)\in [0,1]$ quantifies the uncertainty present in the distribution of $X$. If $H(X) = 1$, it implies that there is no certainty regarding the outcome of the random variable $X$. In contrast, if $H(X) = 0$, it means that the outcome of $X$ can be predicted with perfect certainty. The conditional entropy of $X$ given $Y$, denoted as $H(X|Y)$, measures the remaining uncertainty in $X$ when the random variable $Y$ is known.

\begin{equation}\label{Conditioanl Entropy}
	\begin{split}
		H(X|Y) &= E_{Y \sim p(y)}\{ H\left(X|Y=y\right)\}\\
		&= \sum_{y\in Y}p(y)H\left(X|Y=y\right) \\		
		&=-\sum_{y\in Y}p(y) \sum_{x\in X}p\left(x|y\right)log\left(p\left(x|y\right)\right)\\
		&=-\sum_{y\in Y}\sum_{x\in X}p\left(x,y\right)\log p\left(x|y\right)\\
	\end{split}	
\end{equation}
$H(X|Y) \in [0,1]$, and the greater the value of $H(X|Y)$, the less predictable the value of $X$ becomes when the value of $Y$ is known. If $H(X|Y) = H(X)$, it indicates that $X$ and $Y$ are independent variables. 

Similarly, the joint entropy of a pair of random variables $(X,Y)$ with a joint distribution $p(x,y)$ is defined as:
\begin{equation}
\label{eq:joint_entropy}
    \begin{split}
        H(X,Y) &= E\left[ -\log p(X,Y) \right] \\
        &= -\sum_{x\in X}\sum_{y\in Y}p(x,y)\log p(x,y)
    \end{split}
\end{equation}

Mutual information is defined as the reduction in entropy of one random variable due to the knowledge of the other random variable. Mutual information of two random variables
$X$ and $Y$, denoted by $I(X;Y)$, is:
\begin{equation}\label{MI}
	I(X;Y)  = H(X) - H(X|Y)
\end{equation}
The conditional mutual information of two variables $X$ and $Y$, given another variable $Z$, is defined as:
 \begin{equation}\label{Conditional MI}
 	I(X;Y|Z)  = H(X|Z) - H(X|Y,Z)
 \end{equation}
The amount of information among more than two variables is called interaction information. The conditional mutual information can be used to inductively define the interaction information as follows:
 \begin{equation}
\label{Interaction Information}
 	I\left(X_1, X_2, \dots, X_n\right) = I\left(X_1, X_2, \dots, X_{n-1}\right)-I\left(X_1, X_2, \dots, X_{n-1}|X_n\right) 
\end{equation}

\subsection{Related work}
\label{subsec:Related work}

In recent years, numerous MLFS algorithms have been proposed in the literature. These algorithms can be classified into two primary trends, characterized by the degree of abstraction in the label space: 1) problem transformation methods and 2) algorithm adaptation methods.


Problem transformation methods involve converting the label space into a single label, followed by applying a single-label feature selection approach to address the resultant problem. The label powerset (LP) approach, proposed by Trohidis et al. \cite{trohidis2008multi}, treats each distinct label subset as an individual label. While LP considers label correlations, it grapples with an imbalance in the number of examples associated with numerous label subsets.
To mitigate this concern, Tsoumakas and Vlahavas \cite{tsoumakas2007random} introduced RAKEL, an ensemble method. RAKEL constructs each ensemble member by training a single-label classifier on a randomly chosen subset of labels.
Another strategy is the pruned problem transformation (PPT) method proposed by Read \cite{read2008pruned}. PPT eliminates samples featuring label combinations occurring less frequently than a predefined minimum threshold $\tau$. Doquire and Verleysen \cite{doquire2011feature} utilize PPT and employ a greedy approach based on mutual information to develop an MLFS algorithm for feature selection. Their algorithm, known as PPT-MI, utilizes the following heuristic function:
\begin{equation}\label{PPT-MI}
J\left(f\right) = I\left(F-S;PPT(L)\right) - I\left(F-S-{f},PPT(L)\right)
\end{equation}
PPT-MI, demonstrates high computational efficiency by assessing each feature through a singular multi-variate information calculation. However, it faces challenges with high-dimensional joint probability estimation, primarily stemming from a finite number of patterns in extensive feature spaces. This issue results in a loss of feature information and can potentially reduce accuracy, particularly in datasets where the number of features ($|F|$) or labels ($|L|$) is substantial.

Conversely, algorithm adaptation methods directly exploit multi-label data for feature evaluation, affording greater control over individual labels and their inter-label correlations. Consequently, this approach has garnered increased attention in contrast to the problem transformation methods.

Within the domain of wrapper methods, Zhang et al. \cite{zhang2009feature} introduce MLNB, which employs a dual-stage approach to feature subset selection. Initially, it utilizes principal component analysis to eliminate irrelevant and redundant features. Subsequently, it harnesses a genetic algorithm coupled with a multi-label naive Bayes classifier to ascertain the optimal feature subset.
Shao et al. \cite{shao2013symptom} formulate MLFS as a hybrid optimization problem, addressing it through a blend of simulated annealing, genetic algorithm, and hill-climbing optimization techniques.
Gharroudi et al. \cite{gharroudi2014comparison} explore the utilization of three variants of the Random Forest classifier: Binary Relevance Random Forest (BRRF), Random Forest Label Power-set (RFLP), and Random Forest of Predictive Clustering Trees (RFPCT), to assess feature importance in MLFS tasks.
Despite the capability of wrapper-based MLFS methods to identify relatively accurate feature subsets, they often grapple with execution time limitations. The runtime of these algorithms is dictated by the time needed for model training on each feature subset. As the number of potential subsets escalates exponentially with feature space dimensions, these wrapper-based approaches become viable only for scenarios featuring a small feature space.
Furthermore, the feature subset selected by a wrapper method is significantly influenced by the specific learning model employed. Put differently, a subset identified based on one model may not inherently be accurate or optimal for other learning models. This lack of generalizability curtails the applicability of wrapper-based methods across diverse learning models.




Embedded methods in MLFS also offer effective approaches. The MLSI algorithm \cite{yu2005multi} employs Latent Semantic Indexing (LSI) \cite{dumais2004latent} to discover low-dimensional semantics of both features and labels simultaneously. Similarly, the Multi-label Informed Feature Selection (MIFS) algorithm by Jian et al. \cite{jian2016multi} combines $L_{2,1}$-norm and LSI within the loss function for regression problems. LSI identifies low-dimensional latent label correlations, while the $L_{2,1}$-norm discourages large feature subsets.
Zhang and Zhou \cite{zhang2010multilabel} introduce the MDDM method, which aims to reduce the feature space dimensionality through a closed-form maximization problem based on the Hilbert-Schmidt Independence Criterion \cite{gretton2005measuring}. However, these embedded methods, while effective, share a similar limitation with wrapper methods: they tend to favor the specific learning model used during selection. As a result, a feature subset chosen via an embedded approach might not generalize well across various learning models. Moreover, they may lose valuable label information \cite{zhang2019distinguishing}. In contrast, this paper employs a filter-based MLFS method, as filter methods offer simplicity, efficiency, and model-agnosticism. This design ensures that the proposed method can operate as an independent data pre-processing module.



Several prominent filter-based MLFS methods in the literature employ information-theoretical measures to assess potential features. Li et al. \cite{li2014multi} introduce IGMF, which gauges the significance of a candidate feature using normalized information gain as follows:
\begin{equation}\label{IGMF}
J(f) = \frac{2*\left(H(f) + H(L) - H(f,L)\right)}{H(f) + H(L)}
\end{equation}
The computation of $H(L)$ in Eq.\ref{IGMF} entails a single $|L|$-dimensional joint probability estimation, rendering IGMF computationally efficient. Nonetheless, this approach also exposes the challenge of high-dimensional joint probability estimation. Consequently, IGMF may encounter difficulties in accurately estimating joint probabilities, particularly when confronted with an extensive array of labels.
Moreover, IGMF does not account for feature redundancy within its evaluation process. This characteristic implies that the method might overlook redundant features when striving to identify the optimal feature subset.

Lee and Kim \cite{lee2013feature} demonstrate that the information-theoretical-based MLFS problem can be expressed as a sum of multiple multivariate mutual information terms. They propose the Pairwise Multi-label Utility (PMU) heuristic function, defined as:

\begin{equation}\label{PMU}
J(f) =\sum_{l_i\in L}I(f;l_i) -\sum_{f_i\in S}\sum_{l_j\in L}I(f;f_i;l_j)-\sum_{l_i\in L}\sum_{l_j\in L}I(f;l_i;l_j)
\end{equation}

Subsequently, Lee et al. \cite{lee2015mutual} introduced modifications to the PMU method due to its computational inefficiency when dealing with large $|L|$ values. Specifically, the original method's redundancy measure faced challenges in scalability under such circumstances. To enhance scalability, the revised method, known as D2f, omits pairwise label correlations when evaluating redundancy. Consequently, D2f employs the following specific heuristic function:
\begin{equation}\label{D2f}
J(f) = \sum_{l_i\in L}I(f;l_i) -\sum_{f_i\in S}\sum_{l_j\in L}I(f;f_i;l_j)
\end{equation}
A comparable approach is presented in MDMR \cite{Lin2015}, which integrates mutual information with max-dependency and min-redundancy metrics. Max-dependency gauges the extent of a feature's relation to the target variable, with high dependency indicating high relevance. Min-redundancy quantifies the similarity between a feature and others in the set. Thus, MDMR adopts the following heuristic function:
\begin{equation}\label{MDMR}
J(f) = \sum_{f_j \in S} \sum_{l_i \in L} I(f;l_i) - I(f,l_i,f_j)
\end{equation}
The PMU, D2f, and MDMR algorithms exhibit two significant shortcomings: 1) They employ second-order correlations for redundancy assessment, resulting in increased time complexity proportional to the subset size ($|S|$). Consequently, these algorithms struggle to effectively rank the feature space or select relatively large subsets. 
2) The number of correlation terms considered for relevance evaluation remains fixed at $|L|$ for all features. However, in redundancy evaluation, the number of terms grows as more features are selected. For instance, in a multi-label dataset with $|L|=6$, using D2f, evaluating the second feature involves 6 relevance and 6 redundancy terms, while evaluating the tenth feature requires considering 6 relevance and 60 redundancy terms. As the selected subset size expands, the influence of relevance on the assessment of subsequent features diminishes.

LRFS, proposed by Zhang et al. \cite{zhang2019distinguishing}, introduces conditional mutual information to account for pairwise label redundancy in the assessment of feature relevance. The LRFS criterion is expressed as:

\begin{equation}\label{LRFS}
J(f) = \sum_{l_i\in L}\sum_{l_j\in L, l_i\neq l_j}I(f;l_j|l_i) - \frac{1}{|S|} \sum_{f_i\in S}I(f;f_i)
\end{equation}
Zhang et al. \cite{zhang2021multib} propose label supplementation to capture dynamic changes in label relationships when evaluating candidate features. Using this new measure, they introduce two algorithms, LSMFS and MLSMFS. The LSMFS heuristic function is defined as:

\begin{multline}\label{eq:LSMFS}
J(f)= \sum_{l_i \in L}  I(f,l_i) + \sum_{l_j \in L, l_j \neq l_i} \max\{ 0, I(f,l_i, l_j) \} 
- \sum_{f_i\in S} I(f,f_i)
\end{multline}

MLSMFS, a similar algorithm to LSMFS, employs maximum label supplementation to address label relationships:

\begin{equation}\label{eq:MLSMFS}
J(f) = \sum_{l_i \in L}  I(f,l_i) + \max_{l_j \in L, l_j \neq l_i} \left\{ 0, I(f,l_i|l_j) \right\}  
- \sum_{f_i\in S} I(f,f_i)
\end{equation}

While the aforementioned algorithms incorporate efficient redundancy calculations, their inclusion of second-order correlations in relevance assessment limits their effectiveness in handling extensive label spaces. To address this limitation, Lee and Kim \cite{lee2017scls} introduce the Scalable Criterion for a Large Label Set (SCLS) method. SCLS employs first-order correlations to evaluate the relevance of candidate features, enabling more efficient processing of datasets with sizable label spaces. The heuristic function for SCLS is denoted as follows:

\begin{equation}\label{SCLS}
J(f) = \sum_{l_i \in L}I(f,l_i) - \sum_{f_i\in S}\frac{I(f,f_i)}{H(f)}\sum_{l_i \in L}I(f,l_i)
\end{equation}

However, despite its computational efficiency, SCLS overlooks a significant portion of valuable information inherent in label subsets.




We present a summary of information-theoretical filter-based multi-label feature selection methods in Table \ref{Table: MLFS-Algorithms}. Among these methods, PPT-MI and IGMF adopt an abstract perspective on the label space, utilizing correlations of order $|L|$ for relevance analysis. However, they do not incorporate redundancy analysis. In contrast, PMU, D2F, and MDMR employ first-order correlations for relevance assessment and second-order correlations for redundancy analysis. On the other hand, LRFS, LSMFS, and MLSMFS utilize second-order correlations for relevance evaluation and first-order correlations for redundancy. The computational complexity of each method for ranking the entire feature space is presented in the fifth column of the table.
Upon examination of the complexity column, we can categorize these methods into five distinct computational complexity classes: 1) IGMF, 2) PPT-MI, 3) SCLS (with a complexity attributed to $|F|^3$), 4) LRFS, LSMFS, and MLSMFS (with a complexity attributed to $|F|^3 + |L|^2|F|^2$), and 5) PMU D2F, and MDMR (with a complexity attributed to $|L||F|^3$).

In this paper, we introduce a novel approach that combines first and $|L|$th order correlations to estimate the relevance of features. In terms of computational complexity, the proposed algorithm falls within the same complexity class as SCLS, rendering it well-suited for addressing large-scale problems. However, unlike SCLS, the proposed algorithm takes into account the valuable information present in label subsets, contributing to its enhanced performance.

\begin{table}[H]
	\renewcommand{\arraystretch}{1.3}
	\caption{The summary of multi-label feature selection methods based on information theory}
	\label{Table: MLFS-Algorithms}
	\centering
	\footnotesize
	\begin{adjustbox}{width=1.0\textwidth}
		\begin{tabular}{lcccc}
			
			\hline
			\textbf{Acronym}& \textbf{Type} & \textbf{ Order of correlations for relevance} &   \textbf{Order of correlations for redundancy} & \textbf{Computational Complexity} \\
			\hline
			PPT-MI \cite{doquire2011feature}      & Problem Transformation  & $|L|$  			& -   	 	& $O(|F|^2)$\\
			IGMF   \cite{li2014multi}             & Algorithm Adaptation    & $|L|$  			& -    		&  $O(|F||L|)$\\
			PMU    \cite{lee2013feature}          & Algorithm Adaptation    & first  			& second	&  $O\left(  |L||F|^2 + |L||F|^3 + |L|^2|F|^2 \right)$\\
			D2F    \cite{lee2015mutual}           & Algorithm Adaptation    & first  			& second	& $O\left(  |L||F|^2 + |L||F|^3 \right)$\\
   			MDMR    \cite{Lin2015}           & Algorithm Adaptation    & first  			& second	& $O\left(  |L||F|^2 + |L||F|^3 \right)$\\
			LRFS   \cite{zhang2019distinguishing} & Algorithm Adaptation    & second 			& first  	&$O\left(  |F|^3 + |L|^2|F|^2 \right)$\\
			LSMFS  \cite{zhang2021multib}         & Algorithm Adaptation    & second 			& first  	&$O\left(  |F|^3 + |L||F|^2 + |L|^2|F|^2  \right)$\\
			MLSMFS \cite{zhang2021multib}         & Algorithm Adaptation    & second 			& first  	&$O\left(  |F|^3 + |L||F|^2 + |L|^2|F|^2  \right)$\\
			SCLS   \cite{lee2017scls}             & Algorithm Adaptation    & first  			& first  	&$O\left( |F|^3 + |L||F|^2 \right)$\\
			ATR (Proposed)						  & Algorithm Adaptation    & first + $|L|$ 	& first 	& $O\left( |F|^3 + |L||F|^2 \right)$ \\
			\hline
		\end{tabular}
	\end{adjustbox}
\end{table}  
\section{The Proposed Method}
\label{section:The Proposed Method}
In this section we propose a novel multi-label feature selection method that benefits both algorithm \textbf{A}daptation and problem \textbf{T}ransformation approaches to assess the \textbf{R}elevance of a candidate feature (ATR). We begin by formally presenting the idea behind the information theoretic-based multi-label feature selection problem. Then, we introduce the hybrid heuristic utilized in our algorithm. Finally, we discuss the complexity of the proposed algorithm.

In the following, we present a theorem regarding the calculation of $I(S;L)$ using combinations of multivariate mutual information with varying cardinalities. In the equations below, $S'$ and $S'_k$ represent the power set of $S$, and subsets of size $k$ from $S'$, respectively.

\begin{equation}\label{mutual-information-using-antropy}
	I\left(S; L\right) = H(S)+H(L)-H(S,L)
\end{equation}

\begin{theorem}
	Let $S$ and $T$ be two sets of random variables. Then $I(S;T)$ can be calculated as \cite{lee2013feature}: 
	\begin{equation}
		I(S;T) = \sum_{k=2}^{|S|+|T|}\sum_{p=1}^{k-1} \left(-1\right)^k V_k\left( S'_{k-p}\times T'_p \right)
	\end{equation}
	where $\times$ is the Cartesian product of two sets and $V_k\left( S \right) = \sum_{X\in S_k}I(\{X\})$.
\end{theorem}
For example, suppose $S=\{a,b\}$ and $T=\{x,y\}$, then

\begin{equation*}\label{Joint Entropy}
	\begin{split}
		I(S,T) &= V_2\left( S'_1,T'_1 \right)-V_3\left( S'_2,T'_1 \right) -V_3\left( S'_1,T'_2 \right)+V_4\left( S'_2,T'_2 \right) \\ 
		&= \left[I(\{a,x\})+I(\{a,y\})+I(\{b,x\})+I(\{b,y\})\right] \\
		&- \left[I(\{a,b,x\})+I(\{a,b,y\})\right] - \left[I(\{a,x,y \})+I(\{b,x,y\})\right] \\
		&+ \left[I(\{a,b,x,y\})\right] \\
	\end{split}
\end{equation*}
 
\subsection{The Redundancy Analysis in ATR}

In the context of feature selection, the term redundant features refers to variables that carry similar information or provide little additional value when already considering other features in the dataset. Therefore,  the redundancy of a feature $f$, denoted as $Red\left(f\right)$,  quantifies the extent to which $f$ can be predicted or explained by a combination of other features. Suppose $S \subset F$ is a set of already evaluated features by the MLFS algorithm.  In the incremental selection strategy, the redundancy of a candidate feature $f \in F-S$  can be evaluated using the mutual information between $S$ and $f$:

\begin{equation}\label{redundancy-ideal}
	Red(f) = I(S;\{f\}) = \sum_{k=2}^{|S|+1} \left(-1\right)^k V_k\left( S'_{k-1}\times \{f\} \right)
\end{equation}
 
Calculating Eq.\ref{redundancy-ideal} is a formidable task, particularly when $|S|$ increases over time. This is due to the exponential growth in the number of subsets of $S$ as its size expands. Thus, for computational efficiency, we consider an approximated solution of Eq.\ref{redundancy-ideal} by constraining the calculations of $V_k(.)$ functions with less than two cardinality ($k=2$). Using such a constraint, we obtain: 

\begin{equation}\label{redundancy-proposed}
	\hat{Red}(f) = V_2(S'_1, \{f\})= \sum_{f_j \in S}I({f_j, f})
\end{equation}

It's worth noting that higher cardinalities in Eq.\ref{redundancy-ideal} could potentially capture more complex feature interactions, but doing so comes at a considerable cost of exponentially growing running time. Given that our primary goal is to efficiently rank features, we disregard higher-order correlations to make a balance between computational efficiency and preserving essential feature relationships.

\subsection{The Relevance Analysis in ATR}
The relevance of a feature, denoted as  ($Rel(f)$), in the context of the heuristic function is a measure that quantifies its contribution to determining $L$.  A prevailing heuristic for evaluating feature $f$ in the incremental selection approach is to assess its information gain with respect to the target variable $L$ upon its inclusion in set $S$. This heuristic aims to measure the extent to which incorporating feature $f$ into $S$ would increase the information about $L$. This heuristic can be formulated as the following function: 

\begin{equation}\label{relevance-ideal}
    \begin{split}
        Rel(f) &= I\left(S\cup \{ f \}; L\right) - I\left(S; L\right)\\
            &=\sum_{k=2}^{|S|+|L|+1}\sum_{p=1}^{k-1} \left(-1\right)^k V_k\left( \left(S\cup \{f\} \right)'_{k-1}\times L'_p \right) - \sum_{k=2}^{|S|+|L|}\sum_{p=1}^{k-1} \left(-1\right)^k V_k\left( S'_{k-1}\times L'_p \right)\\
    \end{split}
\end{equation}

Various information-theoretical filter-based MLFS methods consider distinct order of label correlations within the heuristic function. Some MLFS methods assume complete independence among all labels, leading them to employ $V_k(\cdot)$ functions with cardinality less than two ($k=2$) \cite{lee2013feature,lee2015mutual,lee2017scls}. As a result, they only utilize first-order label correlations when calculating feature relevance.  Although these approaches are relatively fast, they often suffer from reduced accuracy due to their oversight of significant information  concealed in label subsets.
The second group comprises methods that incorporate second-order label correlations by considering mutual information between pairwise labels \cite{zhang2019distinguishing,zhang2021multib}. As a part of their relevance analysis heuristic, these methods utilize $V_3(S'_1, L'_2)$. Generally, these methods offer a more sensitive and accurate evaluation of feature relevance. However, they require calculating pairwise mutual information for all label combinations, resulting in a computational inefficiency of $O(|L|^2)$ when the label space is large.

To benefit both computational efficiency and within-label significant information, we propose to use both first-order and $L$th order correlations to estimate feature relevance. In this regard, we utilize $V_2(\dot)$ functions as well as $V_{|L|+1}\left( S'_1 \times L'_{|L|} \right)$, obtaining the following heuristic for relevance analysis:  

\begin{equation}\label{relevance-proposed-initial}
    \begin{split}
        \hat{Rel}(f) &=  V_2\left(\left(S\cup \{ f \} \right)_1'\times L'_1 \right) - V_2\left(S _1'\times L'_1 \right)\\
        &+ \left( -1 \right)^{|L|+1} \left( V_{|L|+1}\left( \left(S\cup \{ f \} \right)_1' \times L'_{|L|} \right) -  V_{|L|+1}\left( S'_1 \times L'_{|L|} \right) \right)\\
        &= \sum_{l \in L}I({f, l}) + \left( -1 \right)^{|L|+1} I\left( \{f,l_1,l_2,\dots,l_{|L|} \} \right)
    \end{split}
\end{equation}
The term $I\left( {f,l_1,l_2,\dots,l_{|L|} } \right)$ offers a comprehensive and abstract perspective on the label space, capturing correlations of order $|L|$ to extract information carried by the entire set of labels. The computational efficiency of calculating this term is remarkably high, as it requires only a single multi-variate information calculation. However, it faces the challenge of high-dimensional joint probability estimation. This issue arises primarily due to the large label space and the limited number of available patterns. As a result, accurately estimating the joint probabilities becomes increasingly difficult, potentially affecting the precision. To tackle this problem, we adopt the pruned problem transformation (PPT) method proposed in \cite{read2008pruned}. PPT effectively addresses the high-dimensional joint probability estimation challenge by removing samples with label combinations that occur less frequently than a predefined minimum threshold $\tau$. Therefore, we rewrite Eq.\ref{relevance-proposed-initial} as follows: 

\begin{equation}\label{relevance-proposed-initial-enhanced}
\hat{Rel}(f) =   \sum_{l \in L}I({f, l}) + \left( -1 \right)^{|L|+1} I\left( \{f,PPT(L,\tau) \} \right)    
\end{equation}

\subsection{Multi-Label Feature Selection Using ATR}
Combining  Eq. \ref{redundancy-proposed} and Eq.\ref{relevance-proposed-initial-enhanced}, we propose the following information-theoretical-based heuristic function to evaluate candidate features in multi-label scenarios:

\begin{equation}\label{proposed-heuristic-function}
J(f) =   \sum_{l \in L}I({f, l}) + \left( -1 \right)^{|L|+1} I\left( \{f,PPT(L,\tau) \} \right)  - \sum_{f_j \in S}I({f_j, f})  
\end{equation}
Algorithm 1 represents the proposed multi-label feature selection method that adopts Eq. (\ref{proposed-heuristic-function}) to rank a set of features. The algorithm initiates with an empty selected subset $S$ and iteratively identifies the feature $f^*$ that maximizes the value of the heuristic function, incorporating it into $S$. After each iteration, the selected feature $f^*$ is excluded from the candidate feature set $F$ to prevent redundancy. This iterative process continues until the desired number of features $N$ is attained.

\begin{algorithm}
    \caption{Multi-Label Feature Selection Using Adaptive and Transformed Relevance (ATR)}
    \label{algo:atr}
    \begin{algorithmic}[1]
        \State \textbf{Input:}
        \State $F$: Set of candidate features,
        \State $N$: The number of to be selected features,
        \State $\tau$: The pruning parameter in PPT
        \State \textbf{Output:}
        \State $S$: Set of top $N$ ranked features
        \State
        \State Initialize $S= \emptyset$, $F = \{f_1, f_2, \dots, f_{|F|}\}$
        \For{$i=1:N$}
            \State $f^* = \arg\max_{f \in F} \text{Eq. (\ref{proposed-heuristic-function})}$
            \State $S = S\cup\{f^*\}$
            \State $F = F-\{f^*\}$
        \EndFor
        \State \textbf{return} $S$
    \end{algorithmic}
\end{algorithm}

\subsection{The Time Complexity of ATR}
 
The time complexity of ATR is determined by the number of mutual information calculations involved in evaluating features using Eq. (\ref{proposed-heuristic-function}). Specifically, for a single feature $f$, our algorithm conducts $|S|$ calculations for redundancy and $|L|+1$ evaluations for relevance analysis. Consequently, the worst-case time complexity for evaluating a single feature is $O\left( |F| + |L| \right)$, considering that $S = F - {f}$ in the worst-case scenario.
To identify the best feature $f^*$ for inclusion in $S$ at each iteration, our algorithm must evaluate all the features in $F-S$. As a result,  finding $f^*$ results in a time complexity of $O\left( |F|\left( |F| + |L| \right) \right) = O\left(  |F|^2 + |F||L|  \right) $. Furthermore, when ranking all the features in $F$, our algorithm performs the aforementioned process $|F|$ times. Therefore, the overall time complexity for ranking all the features is $O(|F|^3 + |F|^2|L|)$.

It is important to note two considerations for the time complexity analysis presented above. Firstly, the analysis considers ranking all the features ($N=|F|$). In practical applications, the goal is often to select only a small number of features ($N \ll |F|$). Consequently, the overall time complexity will be significantly reduced when a smaller subset of features is targeted.
Secondly, during the implementation phase, we have the advantage of being able to calculate mutual information once and reuse it multiple times. This practice eliminates redundant evaluations, significantly speeding up the algorithm's execution. By efficiently managing mutual information calculations, the algorithm's actual performance can be considerably faster than suggested by the worst-case time complexity analysis.

\section{Experimental Results}
\label{section:Introduction}

In this section, we present the results of our experimental evaluation, where we compare our proposed method, against eight existing models, PPT-MI \cite{doquire2011feature}, IGMF \cite{li2014multi}, PMU \cite{lee2013feature}, D2F \cite{lee2015mutual}, LRFS \cite{zhang2019distinguishing}, LSMFS \cite{zhang2021multib}, MLSMFS \cite{zhang2021multib}, and SCLS \cite{lee2017scls}. For both ATR and PPT-MI, we have set the parameter $\tau$ to 6, which aligns with the value suggested in the original paper of PPT \cite{read2008pruned}. Table \ref{table:benchmark} provides a brief overview and statistics of the benchmarks used in our experiments. Whenever possible, we obtained the datasets from the Mulan library \cite{tsoumakas2011mulan}, preserving the default train-test split provided by the library. For datasets not available in the Mulan repository, we performed a random train-test split, allocating 40 percent of each dataset as the test set. We utilized the MLKNN \cite{Zhang2005} classifier, a widely adopted algorithm for multi-label classification tasks. Furthermore, to implement the experiments, we leveraged our previously published library, PyIT-MLFS \cite{Eskandari2022}, which offers robust and efficient functionalities for multi-label classification tasks.
\begin{table*}[t]
\caption{Summary of the benchmarks high dimensional data sets: M: data set size, $|F|$: number of features, $|L|$: number of labels, Card: label cardinality which represents the average number of labels for each instance, Dens: label density which represents the label cardinality divided by the total number of labels}
\centering
\footnotesize
\setlength{\extrarowheight}{1pt}
\begin{tabular}{lccccccr}
\toprule
\textbf{No.}&\textbf{Dataset} & \textbf{Domain} & \textbf{M} & \textbf{$|F|$} & \textbf{$|L|$} & \textbf{Card} & \textbf{Dens} \\
\midrule
1. & Emotions         & Music    & 593   & 72   & 6    & 1.868 & 0.31  \\
2. & Birds  & Audio    & 645   & 260   & 19    & 1.014 & 0.053  \\
3. &Enron            & Text     & 1702  & 1001 & 53   & 3.378 & 0.064 \\
4.& Medical          & Text     & 978   & 1449 & 45   & 1.245 & 0.028 \\
5. &Scene            & Image    & 2407  & 294  & 6    & 1.074 & 0.179 \\
6. &Yeast            & Biology  & 2417  & 103  & 14   & 4.237 & 0.303 \\
7. &Genbase          & Biology  & 662   & 1186 & 27   & 1.252 & 0.046 \\
8. &Tmc2007-500      & Text     & 28600 & 500  & 22   & 2.22  & 0.101 \\
9. &Bibtex           & Text     & 7395  & 1836 & 159  & 2.402 & 0.015 \\
10.& GnegativePseAAC  & Biology  & 1392  & 440  & 8    & 1.046 & 0.131 \\
11.& PlantPseAAC      & Biology  & 978   & 440  & 12   & 1.079 & 0.090 \\
12. & Ng20             & Text     & 19300 & 1006 & 20   & 1.029 & 0.051 \\
\bottomrule
\end{tabular}
\label{table:benchmark}
\end{table*}

\subsection{Evaluation Metrics}
The comparison is carried out using six evaluation metrics, including Hamming Loss, Label Ranking Loss, Coverage Error, F1 Score, Jaccard Score, and Accuracy Score \cite{spolaor2016systematic}. Consider a test set represented as $\mathbb{X}_T^{(F)} = \{(\mathbf{x}_i, \mathbf{y}_i) \mid 1 \leq i \leq M'\}$. Let $\mathbf{\hat{y}_i}$ denote the predicted label set corresponding to $\mathbf{x_i}$.
The Hamming loss measures the fraction of labels that are incorrectly predicted:
\begin{equation}
    HL(\mathbf{y},\mathbf{\hat{y}}) = \frac{1}{M'} \sum_{i=1}^{M'} \left( \frac{1}{|L|} (\mathbf{y_i} \oplus \mathbf{\hat{y}_i}) \right)
\end{equation}
where $\oplus$ represents the exclusive OR (XOR) operation. The label ranking loss evaluates the average number of label pairs that are incorrectly ordered:

\begin{equation}
    LRL(\mathbf{y},\mathbf{\hat{y}}) = \frac{1}{M'} \sum_{i=1}^{M'} \frac{1}{|\mathbf{y_i}| \cdot |\mathbf{\hat{y}_i}|} \left|  C \right|
\end{equation}
where
\begin{equation}
    C =   \{(\mathbf{y_j}, \mathbf{y_k}) | f(\mathbf{x_i}, \mathbf{y_i}) \leq f(\mathbf{x_i}, \mathbf{y_i}), (\mathbf{y_j}, \mathbf{y_k}) \in \mathbf{y_i} \times \mathbf{\hat{y}_i}\} 
\end{equation}
in which the $f(\mathbf{x_i}, \mathbf{y_j})$ represents the probability or chance that label $\mathbf{y_j}$ from set $\mathbf{y}$ is the correct label for instance $\mathbf{x_i}$. The complementary set $\mathbf{\hat{y}_i}$ consists of labels that are not included in $\mathbf{y_i}$. The coverage error measures the average number of additional labels that need to be included in the predicted label set to cover all true labels. This measure is defined as:
\begin{equation}
    CE(\mathbf{y},\mathbf{\hat{y}}) = \frac{1}{M'} \sum_{i=1}^{M'} \left( \max_{\mathbf{y} \in \mathbf{y_i}} \text{rank}(f(\mathbf{x_i}, \mathbf{y})) - 1 \right)
\end{equation}
The F1-score is a widely used measure for evaluating the accuracy of a classification tasks, providing a harmonic mean of precision and recall of the model: 
\begin{equation}
F1-Score(\mathbf{y},\mathbf{\hat{y}}) = 2 \cdot \frac{Precision(\mathbf{y},\mathbf{\hat{y}}) \cdot Recall(\mathbf{y},\mathbf{\hat{y}})}{Precision(\mathbf{y},\mathbf{\hat{y}}) + Recall(\mathbf{y},\mathbf{\hat{y}})}
\end{equation}
where
\begin{equation}
Precision(\mathbf{y},\mathbf{\hat{y}}) = \frac{|\mathbf{\hat{y}_i} \cap \mathbf{y_i}|}{|\mathbf{\hat{y}_i}|}
\end{equation}
and
\begin{equation}
Recall(\mathbf{y},\mathbf{\hat{y}}) = \frac{|\mathbf{\hat{y}_i} \cap \mathbf{y_i}|}{|\mathbf{y_i}|}
\end{equation}
 The Jaccard score evaluates the overlap between the predicted and true label sets by calculating the size of their intersection divided by the size of their union:
\begin{equation}
    JS(\mathbf{y},\mathbf{\hat{y}}) = \frac{|\mathbf{y_i} \cap \mathbf{\hat{y}_i}|}{|\mathbf{y_i} \cup \mathbf{\hat{y}_i}|}
\end{equation}
The accuracy score calculates the percentage of correctly predicted labels out of the total number of labels:
\begin{equation}
    AS(\mathbf{y},\mathbf{\hat{y}}) = \frac{1}{M'} \sum_{i=1}^{M'} \frac{|\mathbf{\hat{y}_i} \cap \mathbf{y_i}|}{|\mathbf{\hat{y}_i} \cup \mathbf{y_i}|}
\end{equation}

 Furthermore, it is important to note that Hamming Loss, Label Ranking Loss, and Coverage Error measure the  discrepancies between the predicted and true label sets. Therefore, achieving lower values for these metrics indicates a higher precision and accuracy in the multi-label classification task. Conversely, the F1 Score, Jaccard Score, and Accuracy Score capture the overlap between the predicted and true label sets. Consequently, higher scores in these metrics indicate superior performance in classification tasks.
\subsection{Classification Accuracy}
\begin{figure*}
	\subfloat{\includegraphics[width=2.2in,height = 1.6in ]{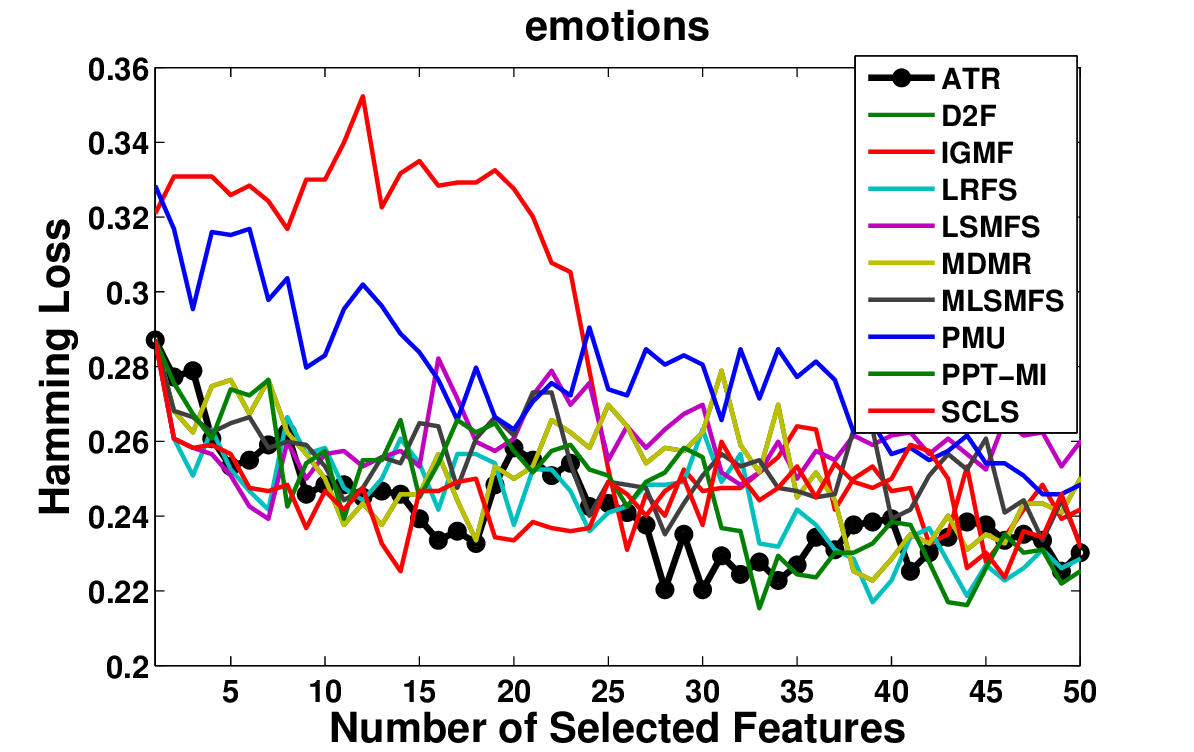}}
	\subfloat{\includegraphics[width=2.2in,height = 1.6in ]{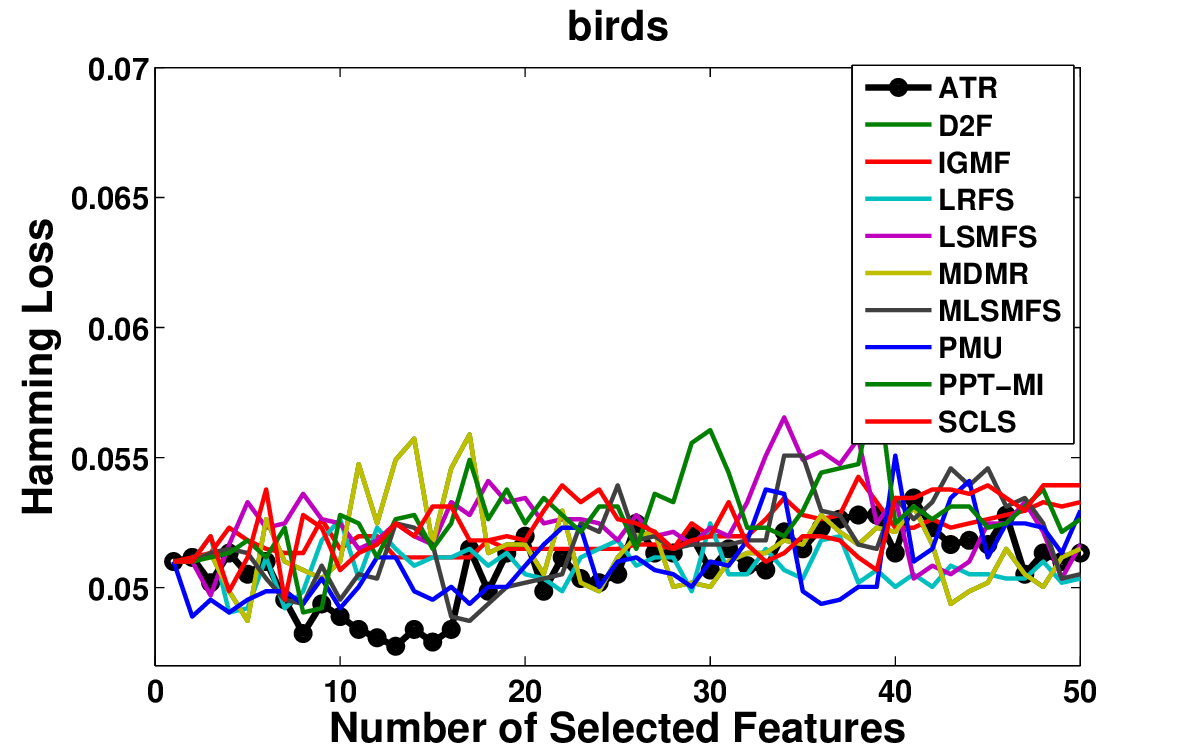}}
	\subfloat{\includegraphics[width=2.2in,height = 1.6in ]{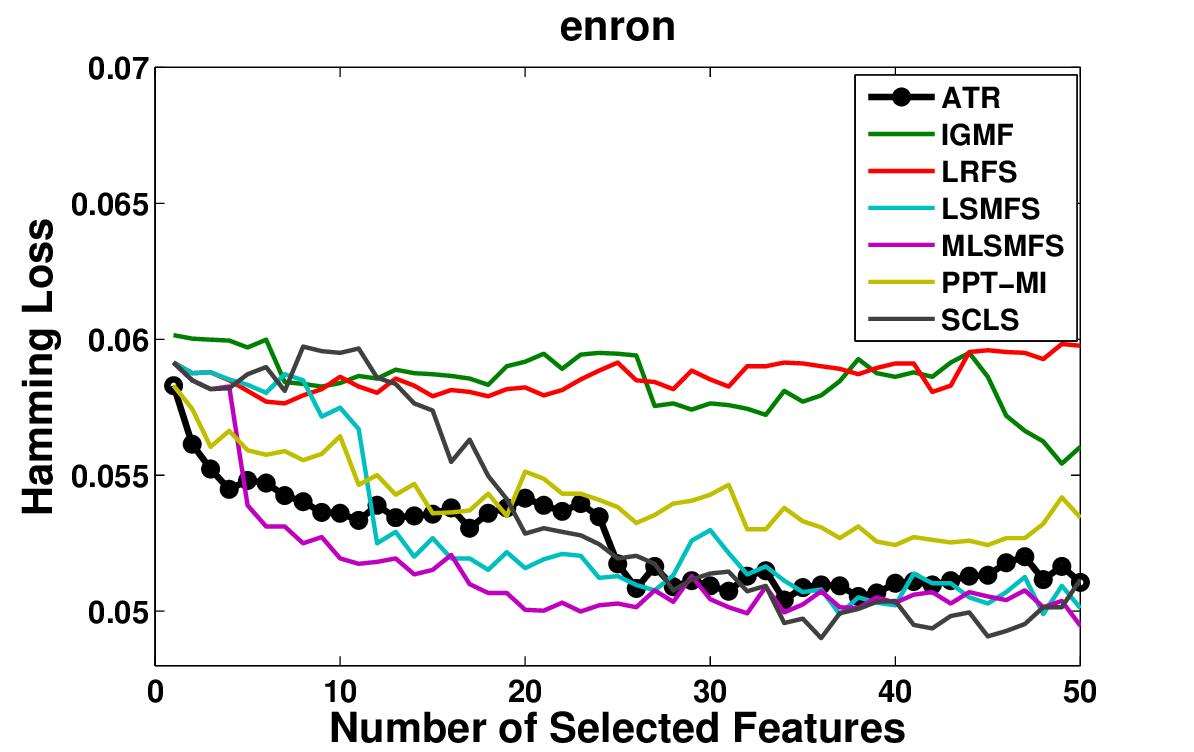}}\\
	\subfloat{\includegraphics[width=2.2in,height = 1.6in ]{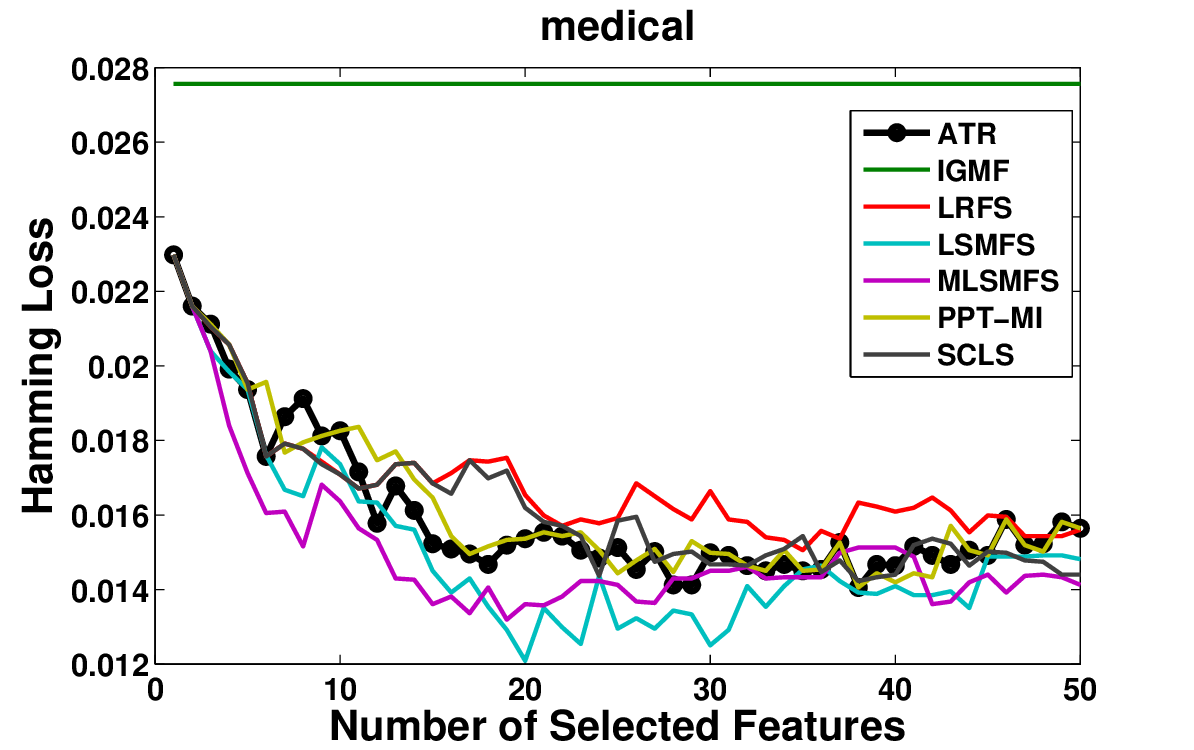}}
	\subfloat{\includegraphics[width=2.2in,height = 1.6in ]{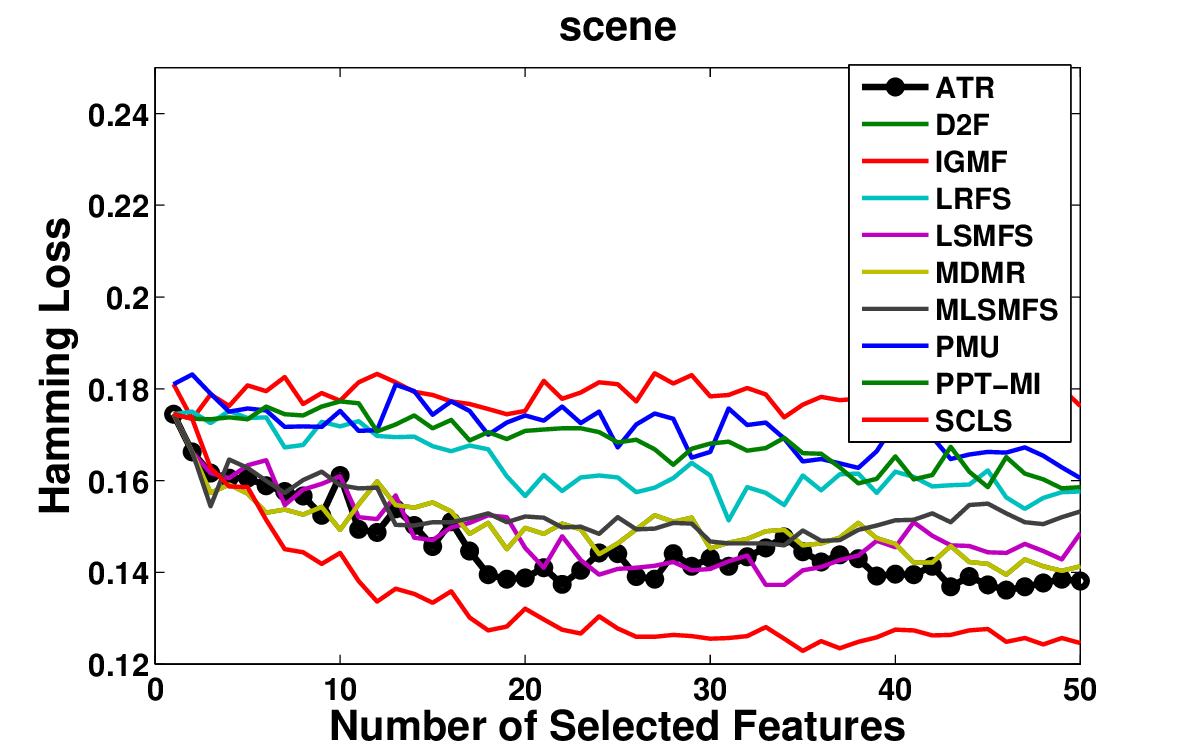}}
	\subfloat{\includegraphics[width=2.2in,height = 1.6in ]{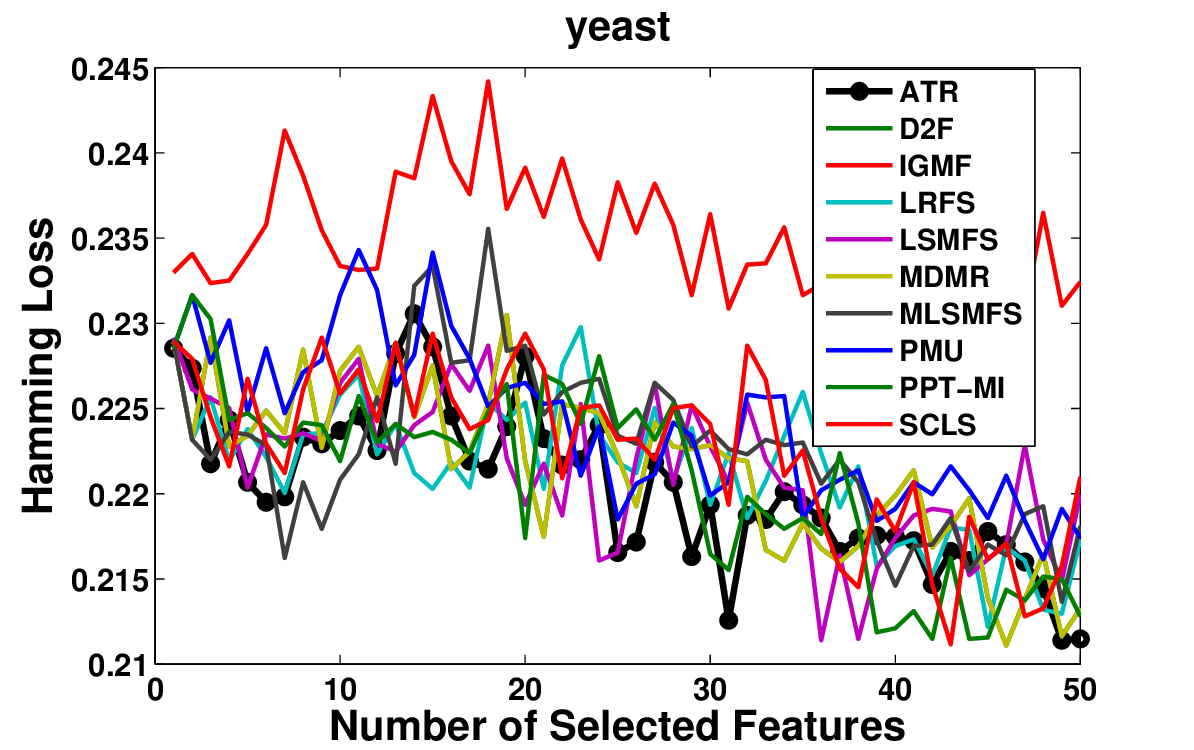}}\\
	\subfloat{\includegraphics[width=2.2in,height = 1.6in ]{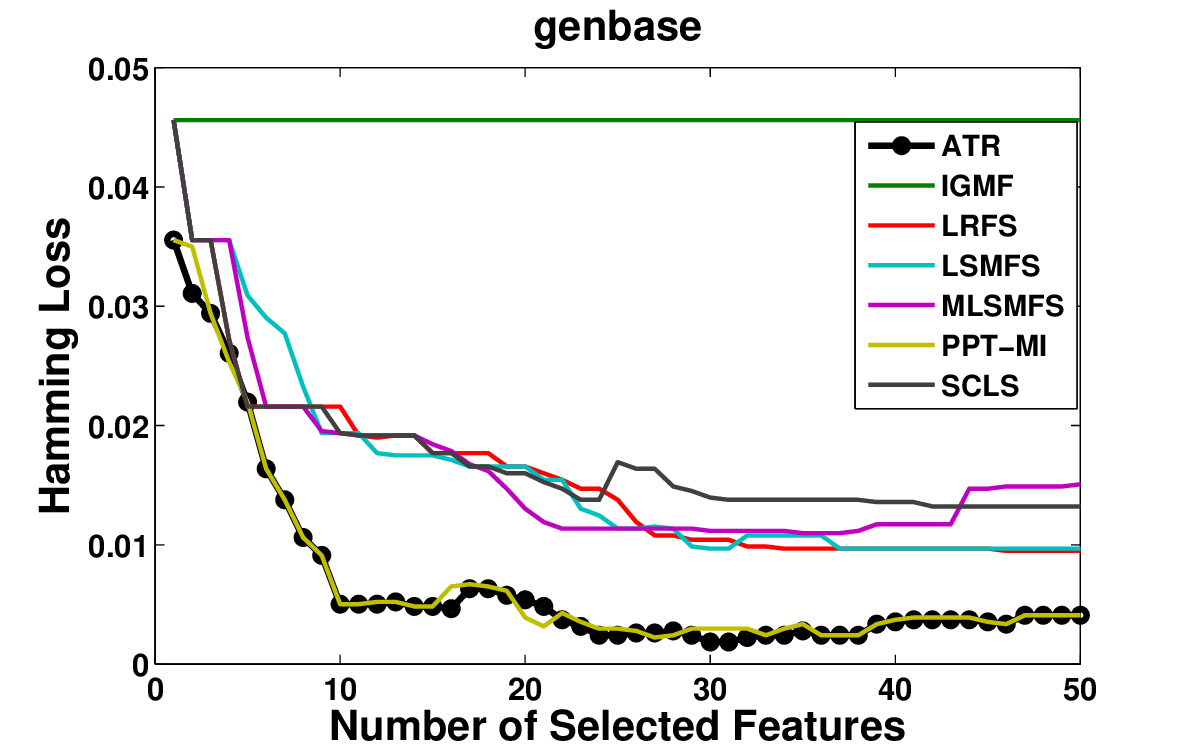}}
	\subfloat{\includegraphics[width=2.2in,width=2.2in,height = 1.6in ]{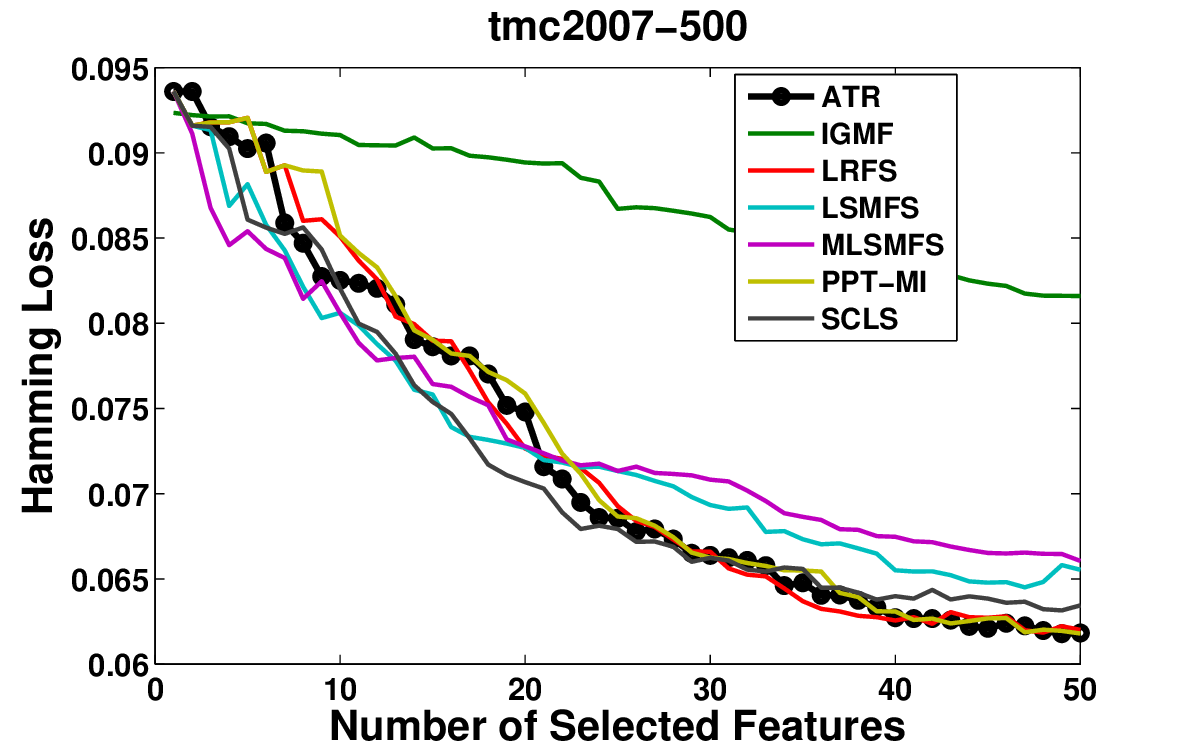}}
	\subfloat{\includegraphics[width=2.2in,height = 1.6in ]{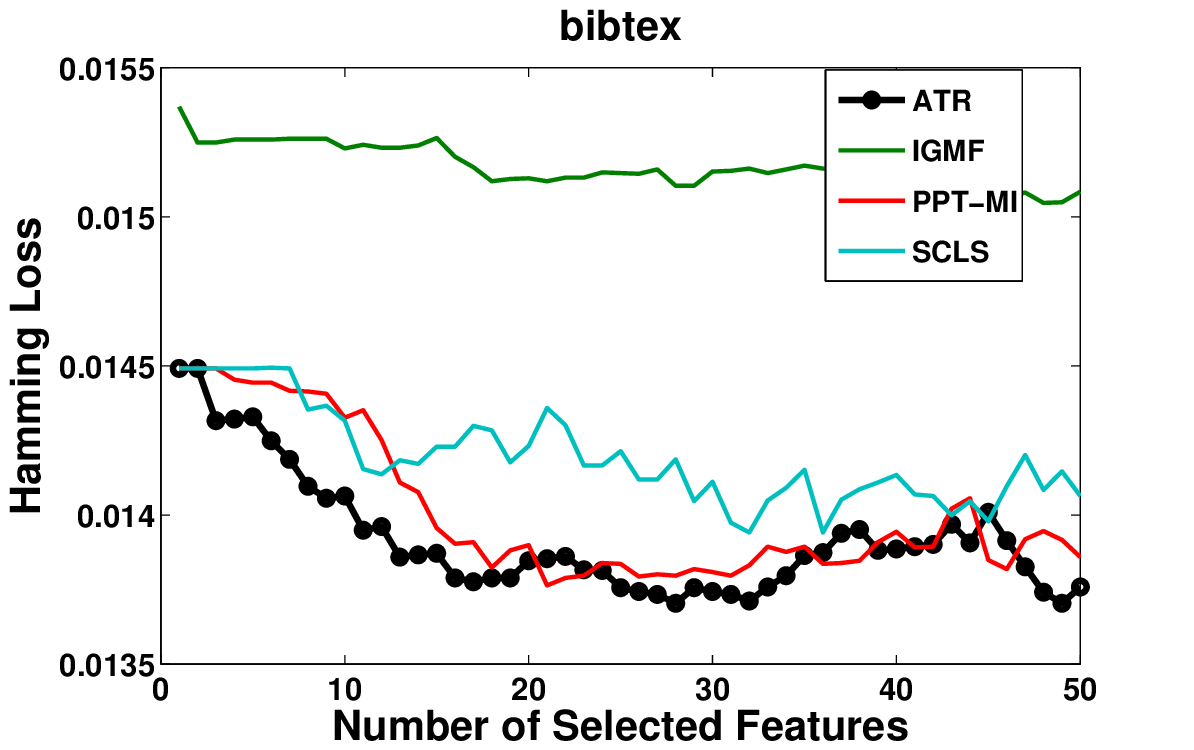}}\\
	\subfloat{\includegraphics[width=2.2in,height = 1.6in ]{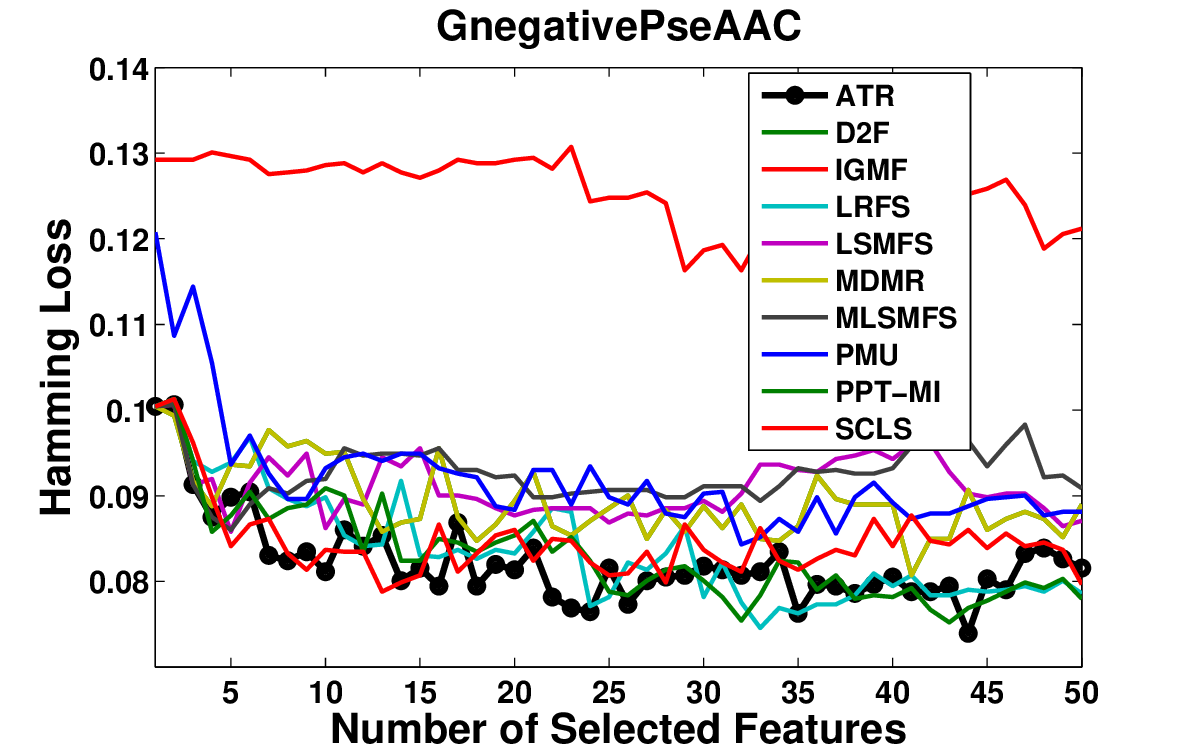}}
	\subfloat{\includegraphics[width=2.2in,height = 1.6in ]{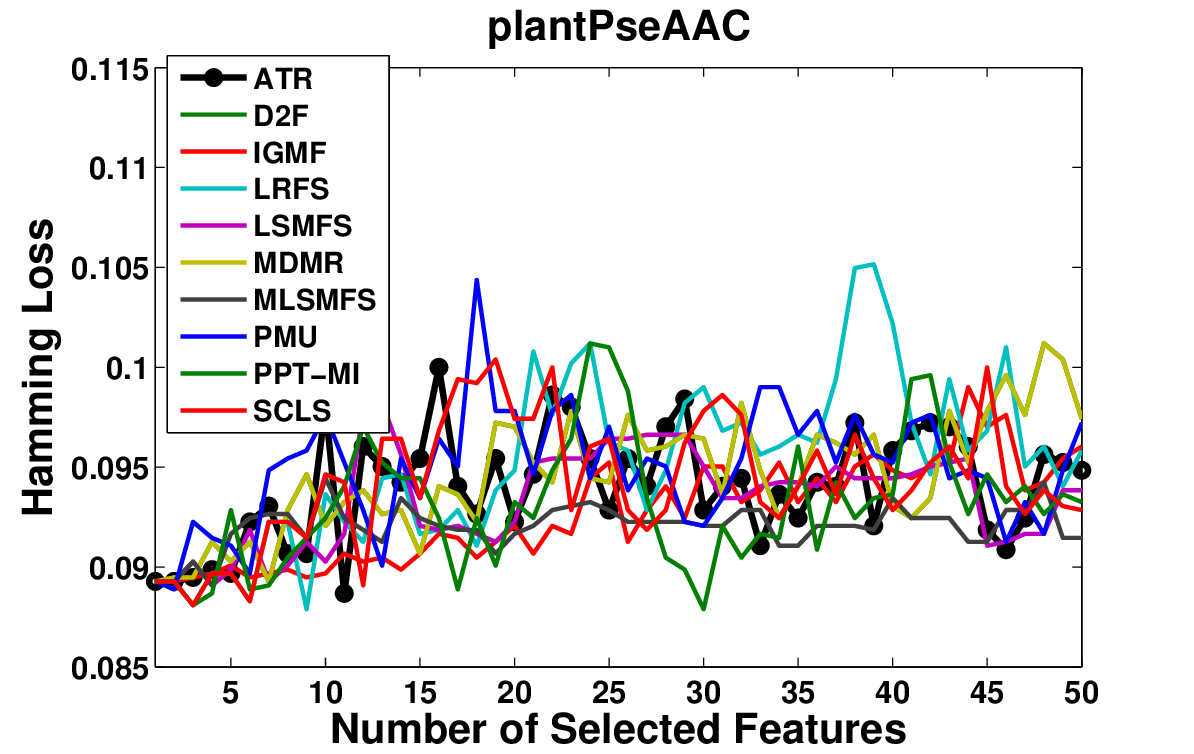}}
	\subfloat{\includegraphics[width=2.2in,height = 1.6in ]{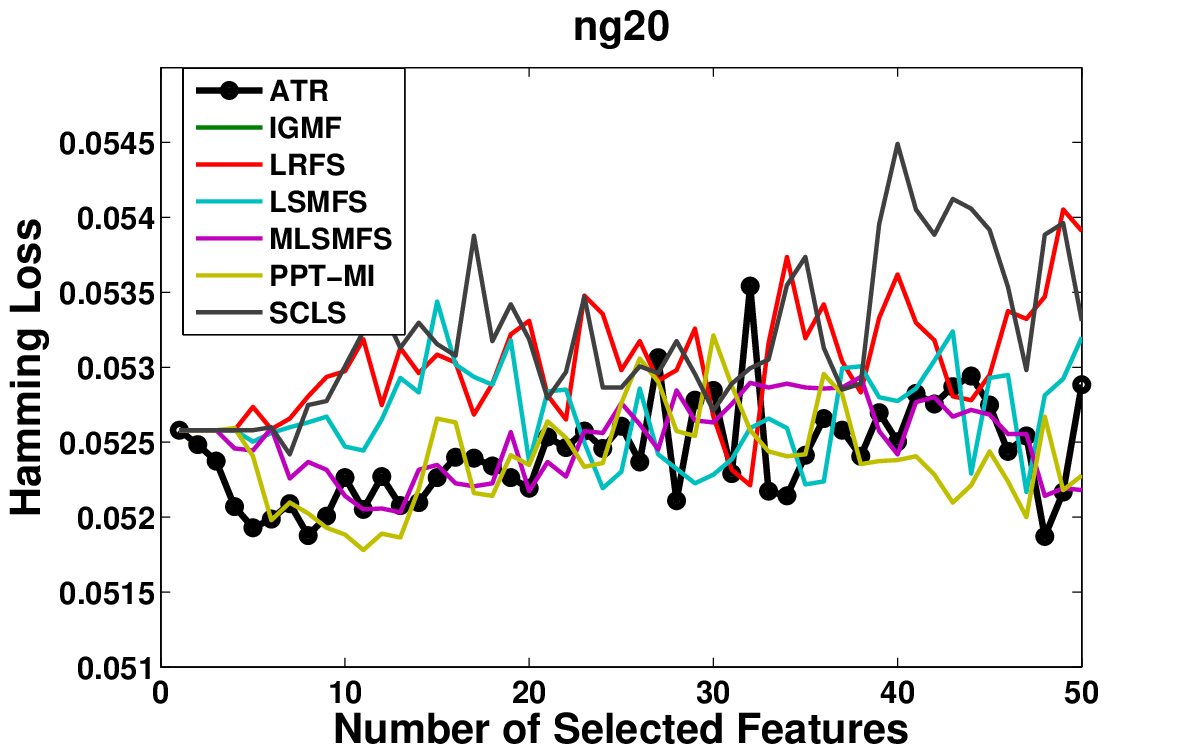}}
	
	\caption{Comparison of MLKNN Hamming Losses for $N=1,2, \dots 50$ using the ten MLFS algorithms }
	\label{fig:mlknn_HL}
\end{figure*}

\begin{figure*}
	
	\centering
	\subfloat{\includegraphics[width=2.2in,height = 1.6in ]{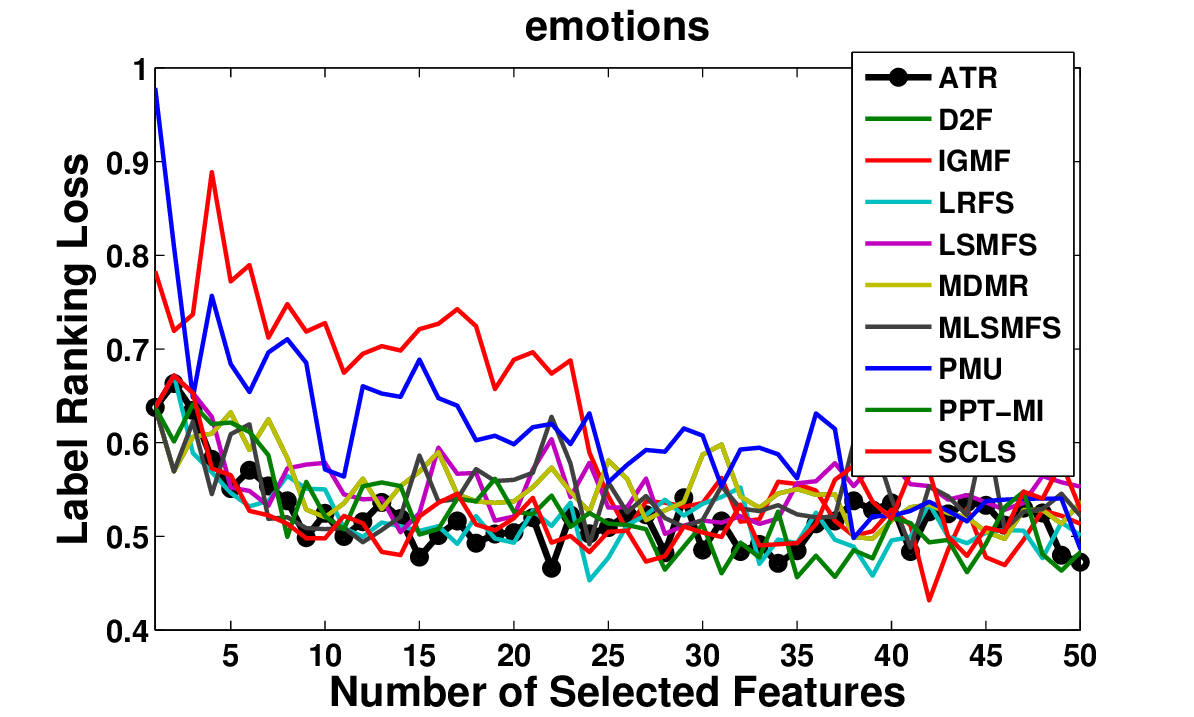}}
	\subfloat{\includegraphics[width=2.2in,height = 1.6in ]{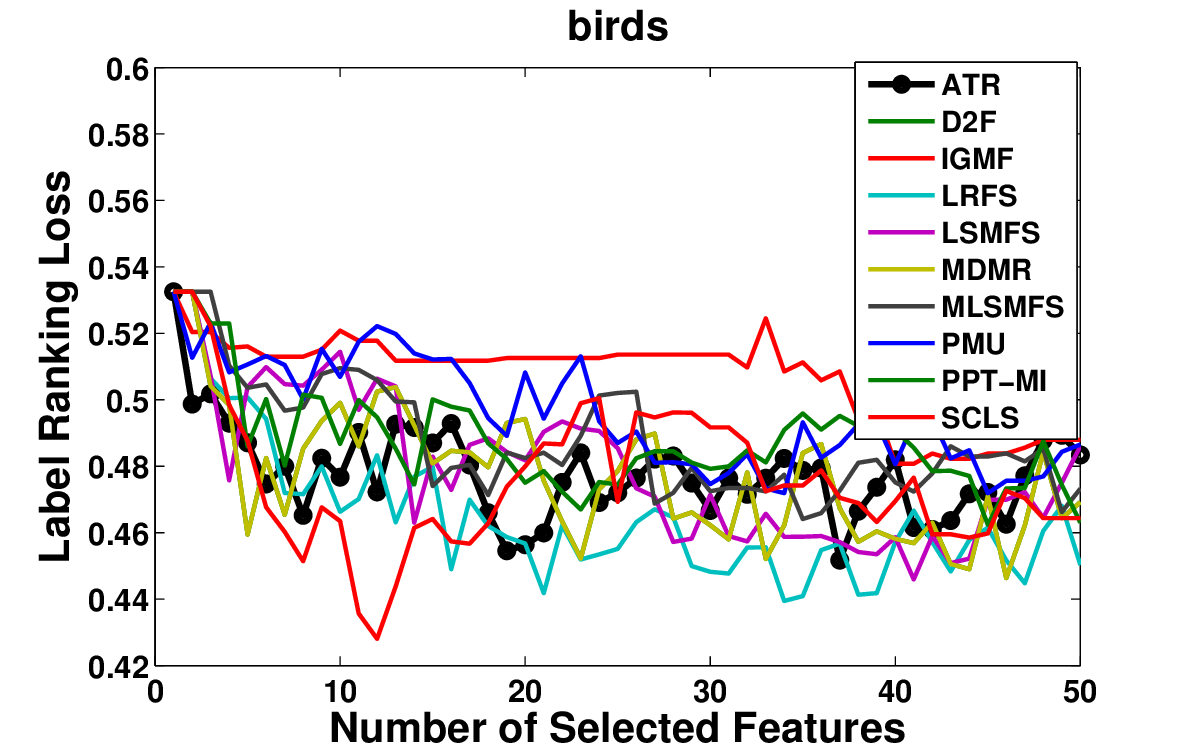}}
	\subfloat{\includegraphics[width=2.2in,height = 1.6in ]{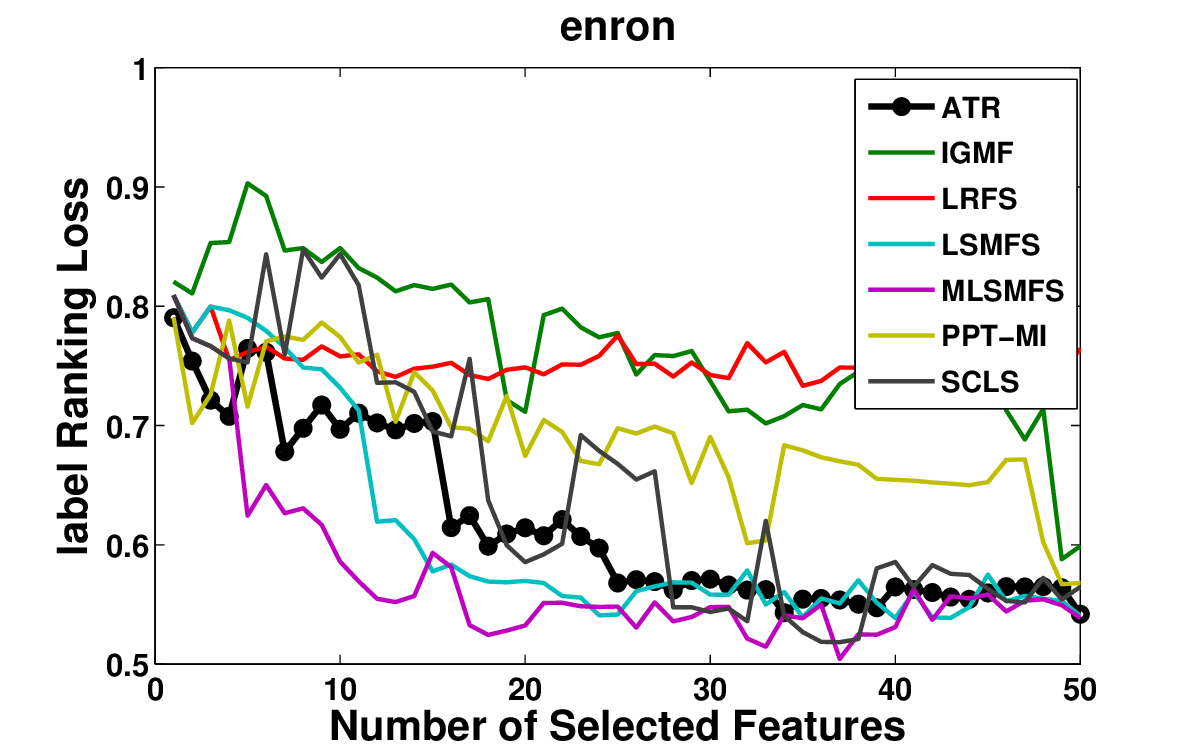}}\\
	\subfloat{\includegraphics[width=2.2in,height = 1.6in ]{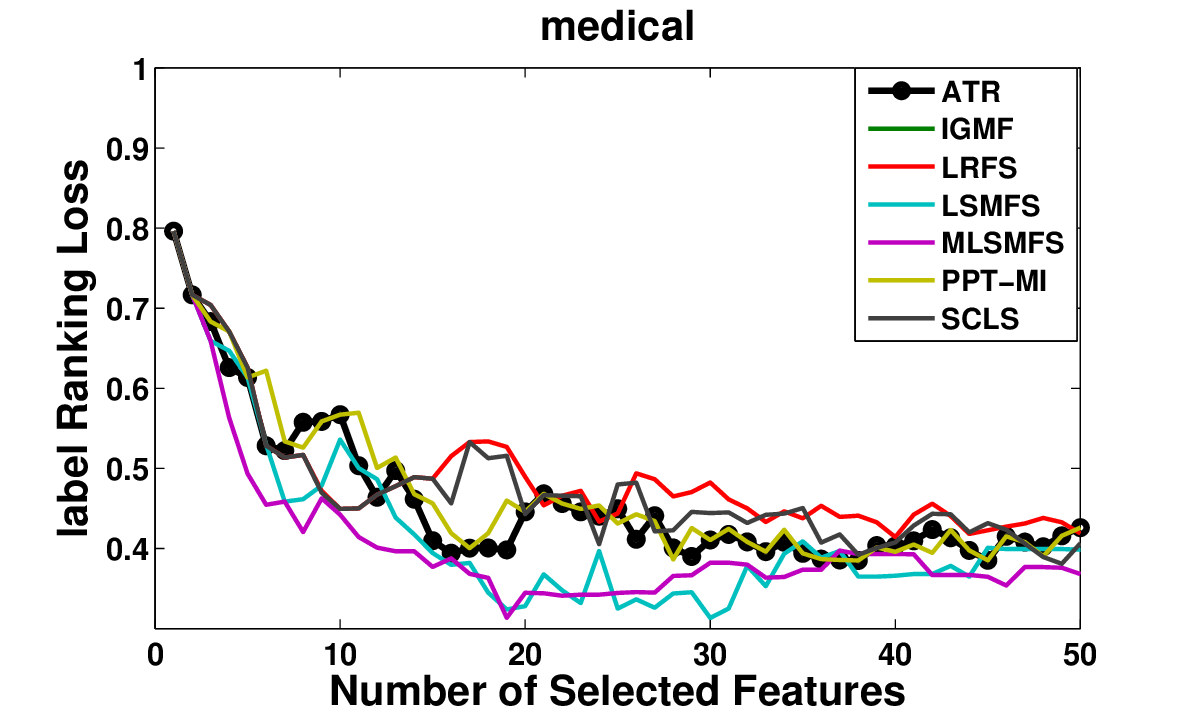}}
	\subfloat{\includegraphics[width=2.2in,height = 1.6in ]{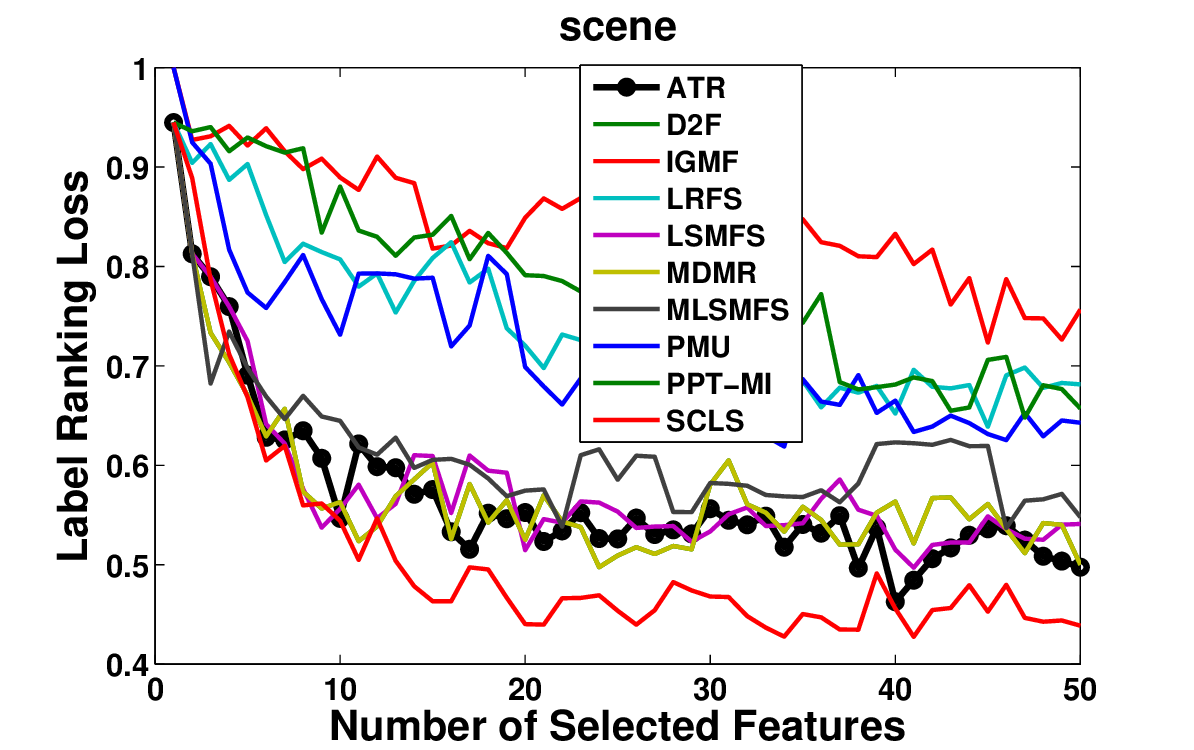}}
	\subfloat{\includegraphics[width=2.2in,height = 1.6in ]{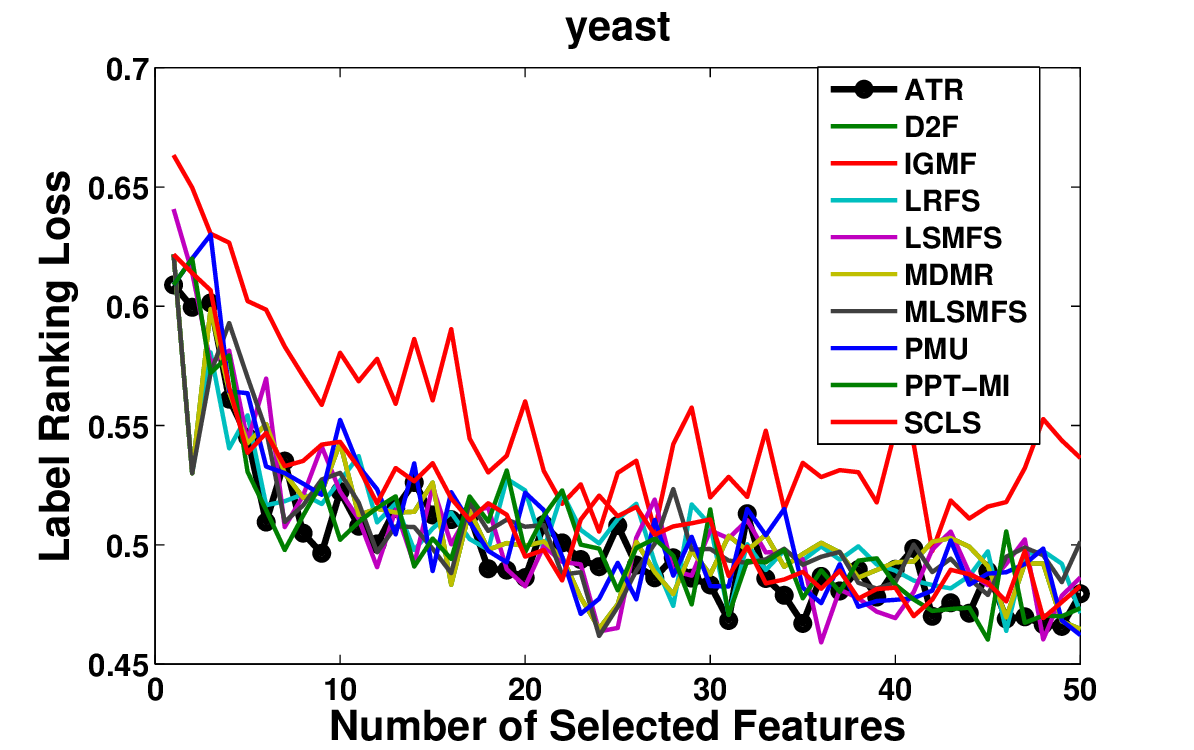}}\\
	\subfloat{\includegraphics[width=2.2in,height = 1.6in ]{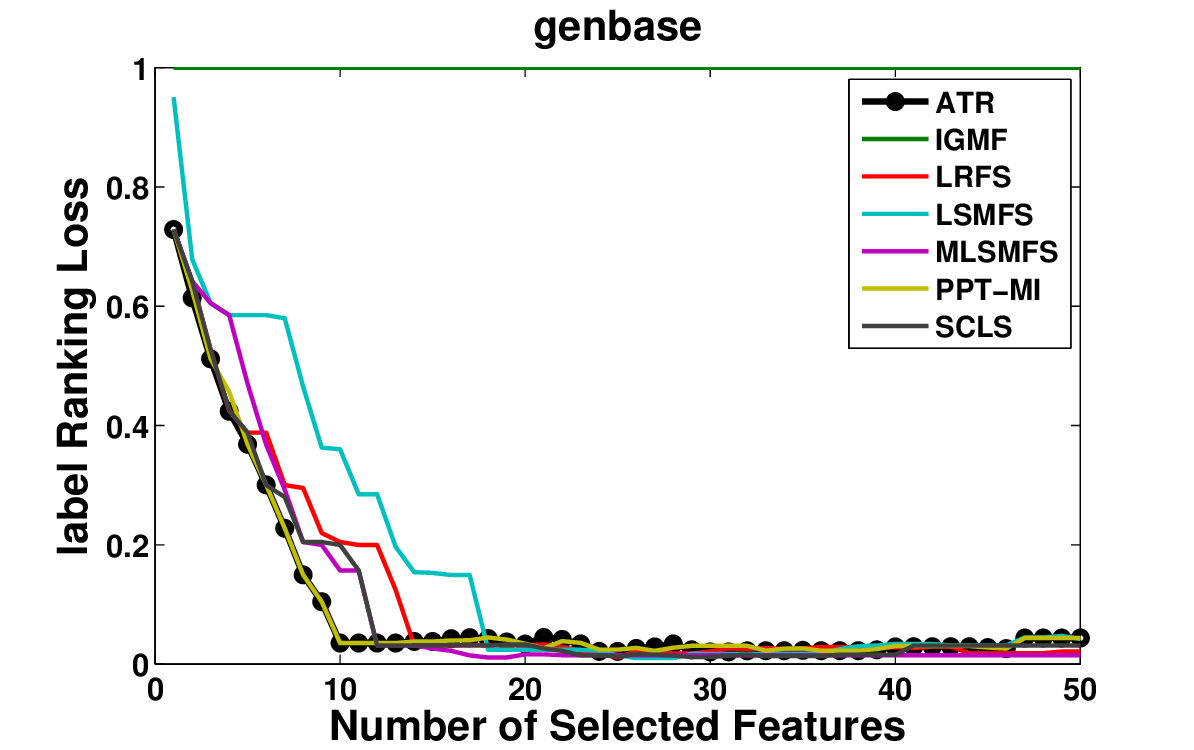}}
	\subfloat{\includegraphics[width=2.2in,height = 1.6in ]{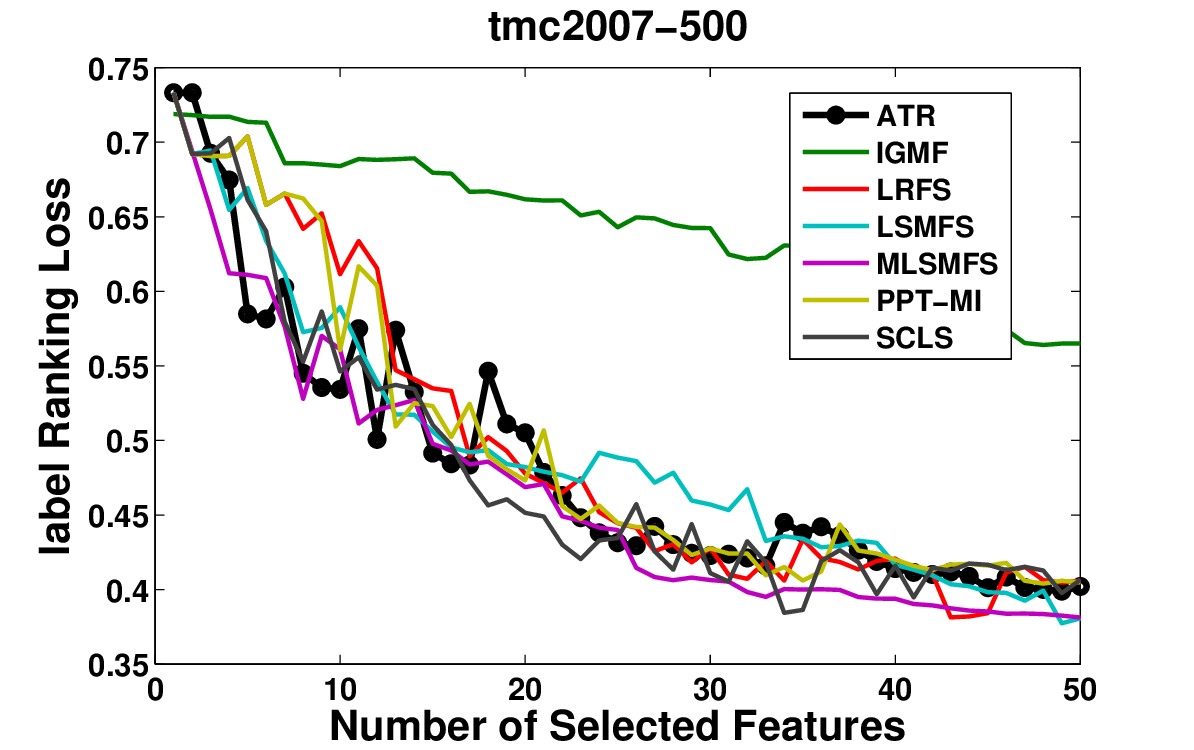}}
	\subfloat{\includegraphics[width=2.2in,height = 1.6in ]{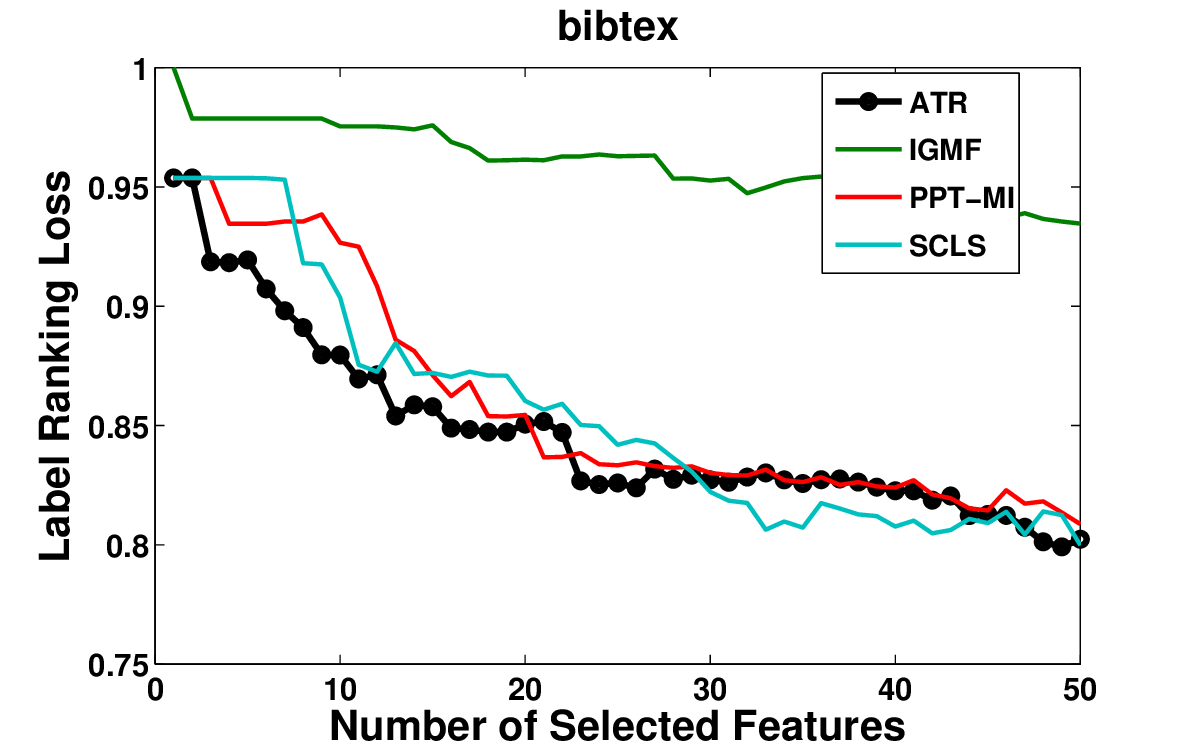}}\\
	\subfloat{\includegraphics[width=2.2in,height = 1.6in ]{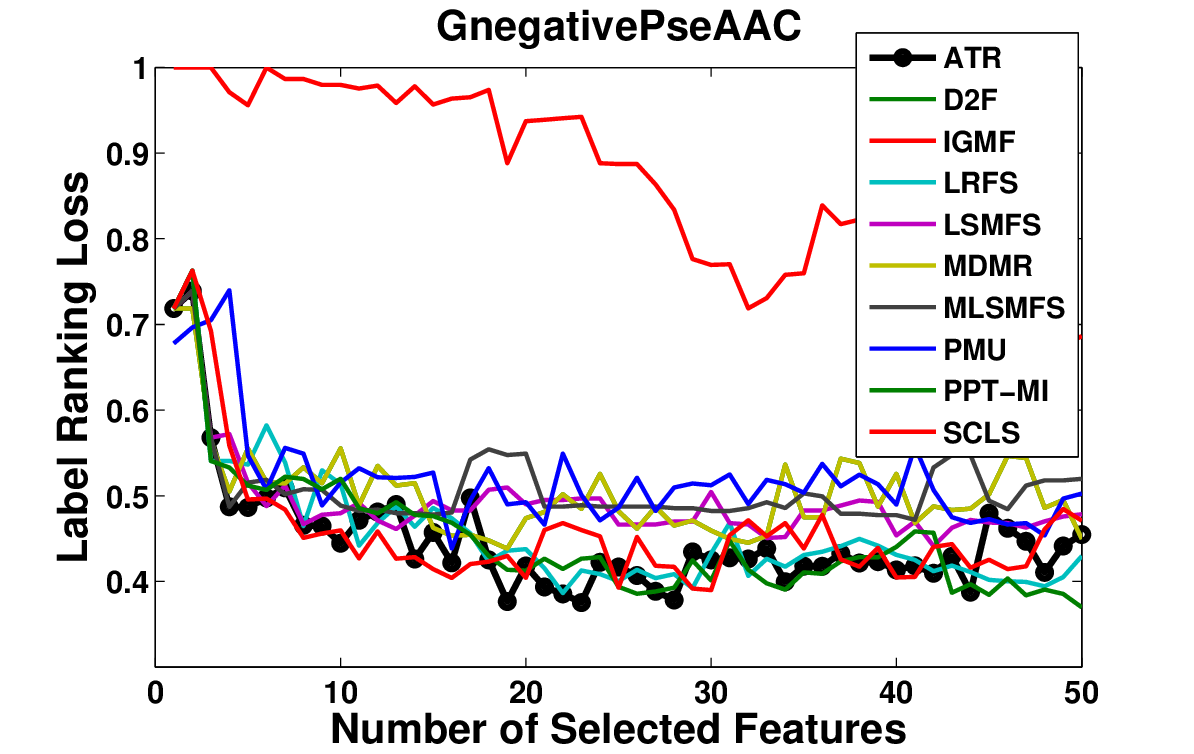}}
	\subfloat{\includegraphics[width=2.2in,height = 1.6in ]{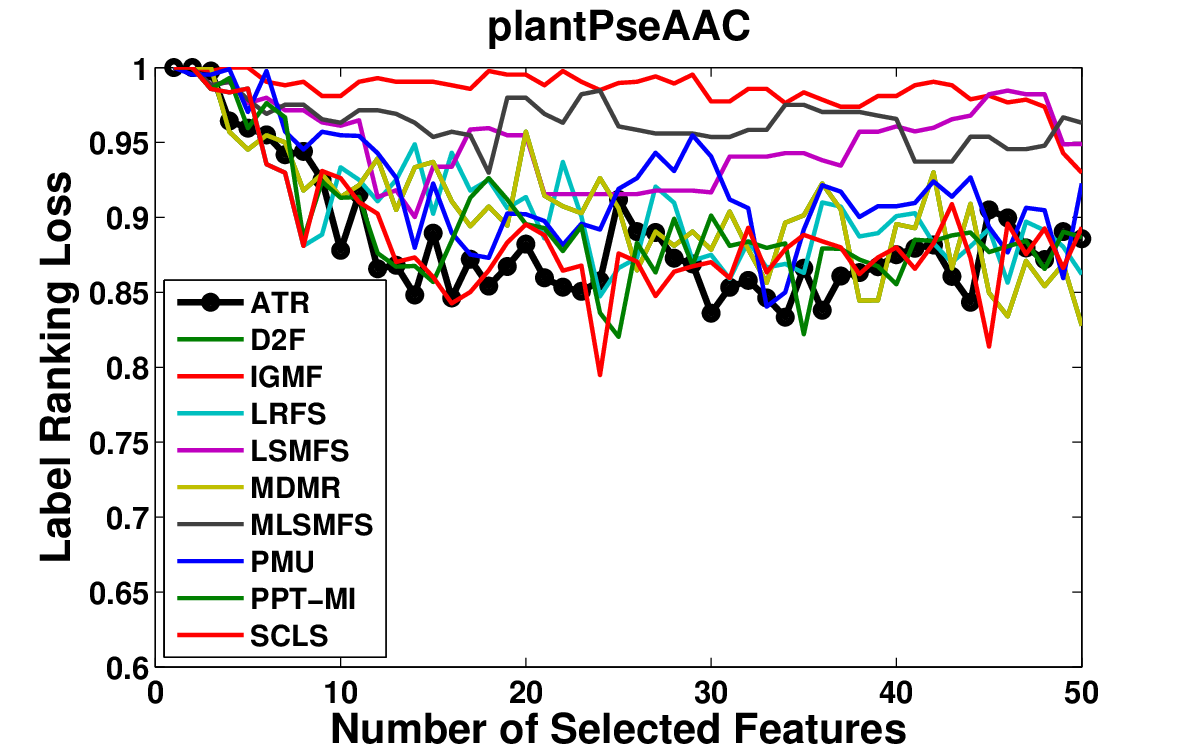}}	
	\subfloat{\includegraphics[width=2.2in,height = 1.6in ]{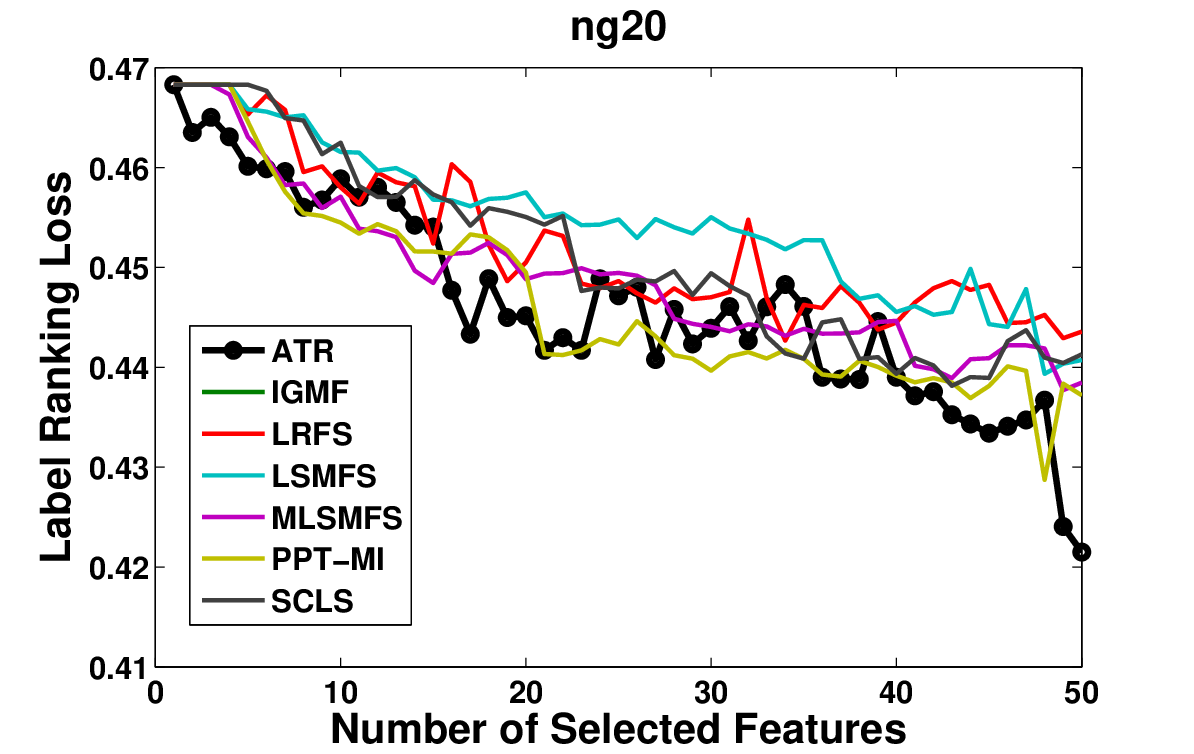}}
	
	\caption{Comparison of MLKNN Label Ranking Losses for $N=1,2, \dots 50$ using the ten MLFS algorithms}
	\label{fig:mlknn_LRL}
\end{figure*}

\begin{figure*}
	
	\centering
	\subfloat{\includegraphics[width=2.2in,height = 1.6in ]{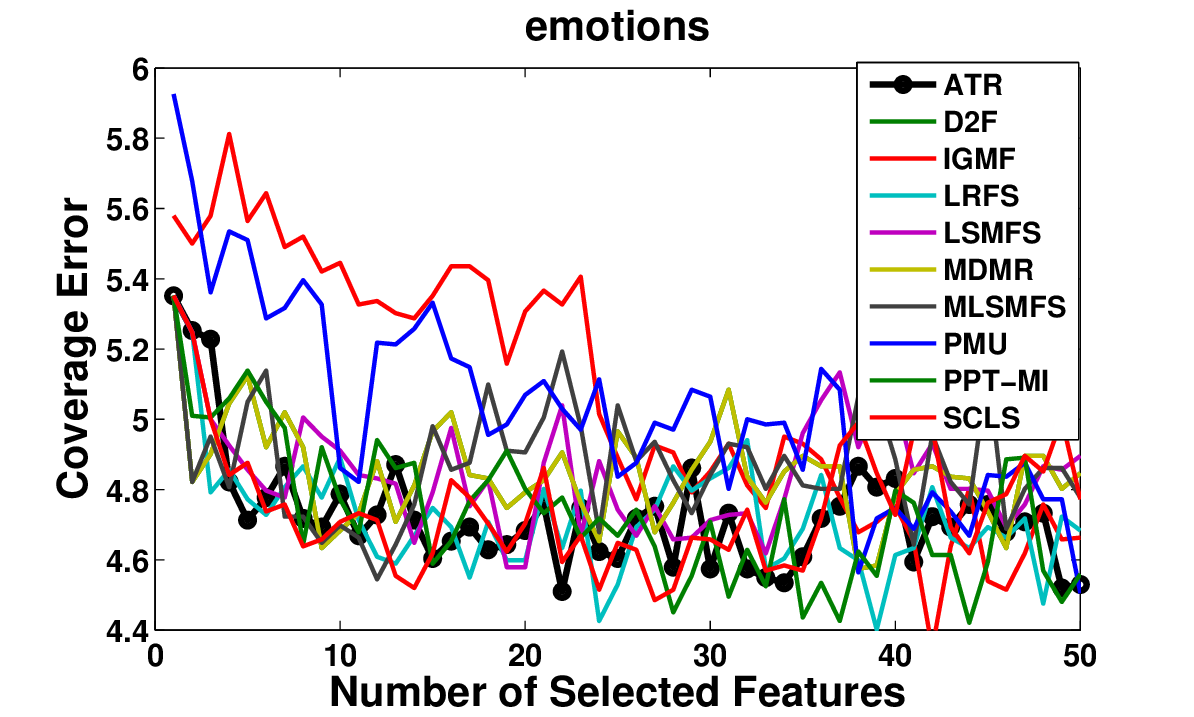}}
	\subfloat{\includegraphics[width=2.2in,height = 1.6in ]{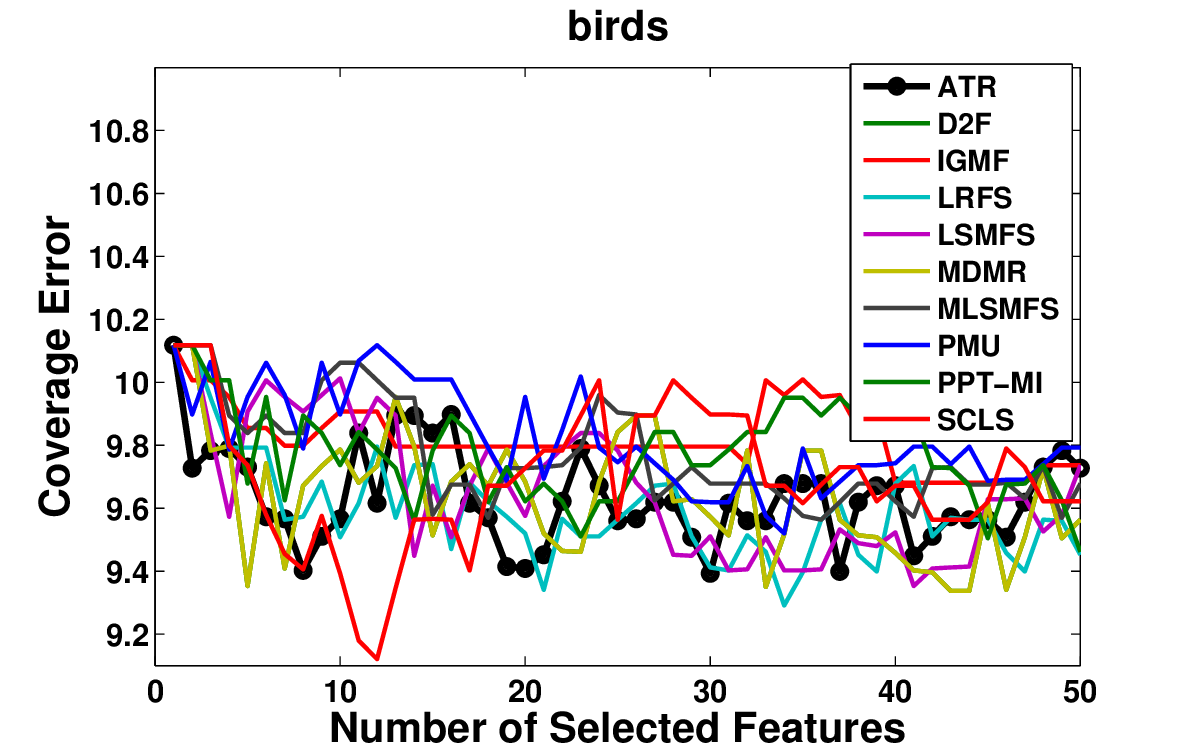}}
	\subfloat{\includegraphics[width=2.2in,height = 1.6in ]{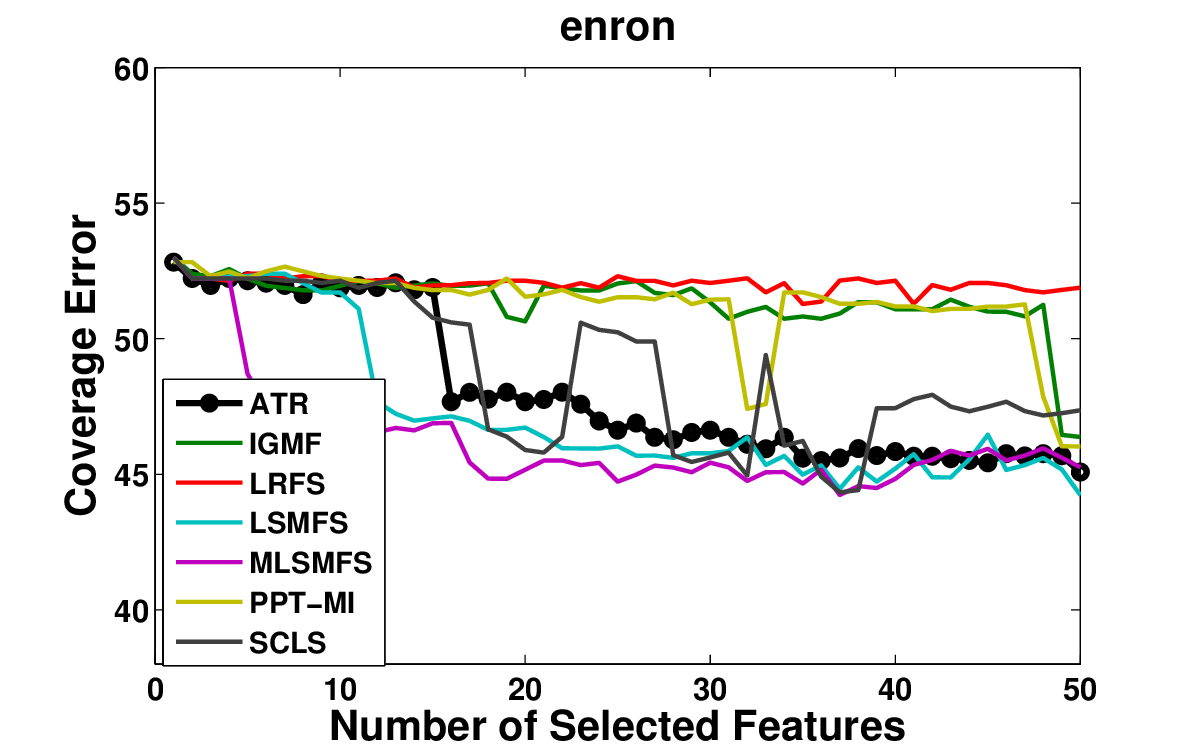}}\\
	\subfloat{\includegraphics[width=2.2in,height = 1.6in ]{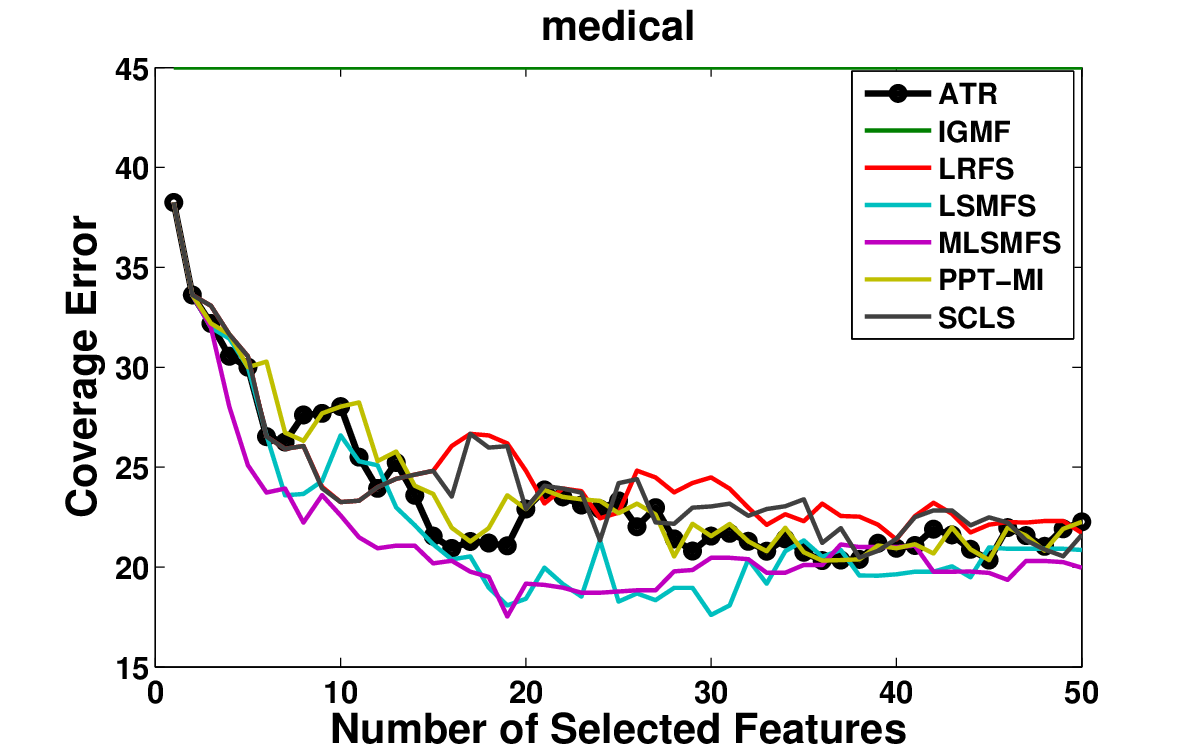}}
	\subfloat{\includegraphics[width=2.2in,height = 1.6in ]{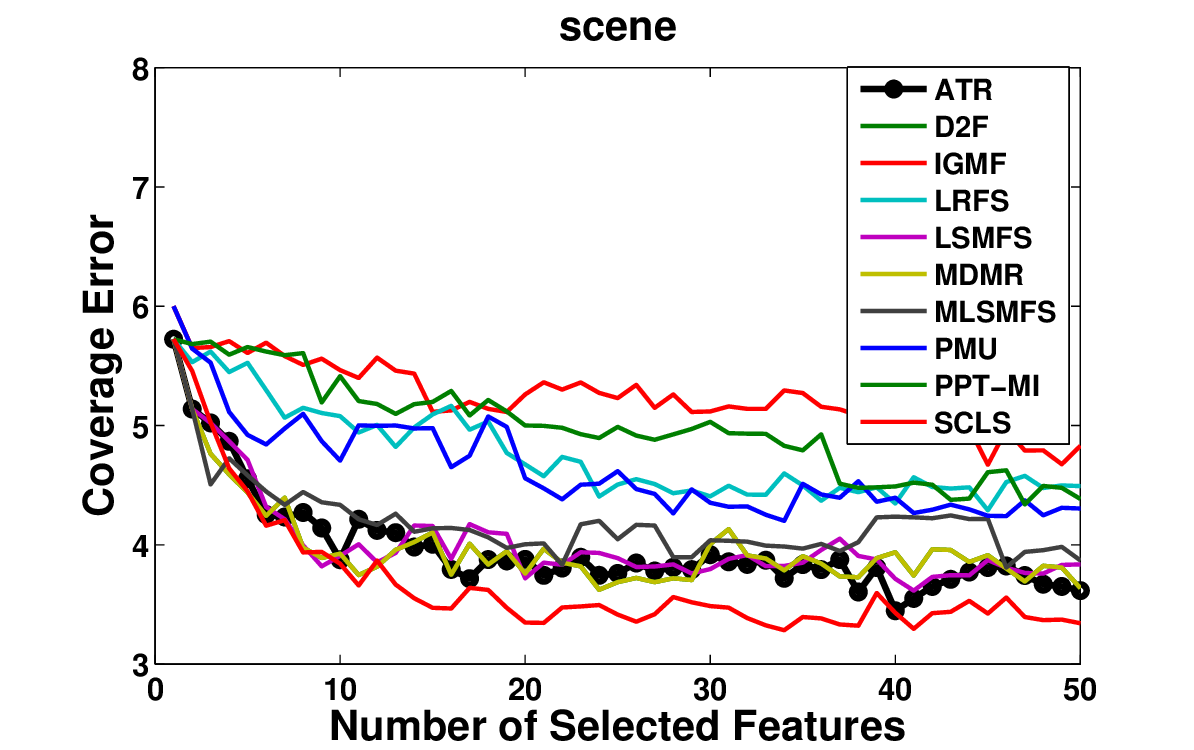}}
	\subfloat{\includegraphics[width=2.2in,height = 1.6in ]{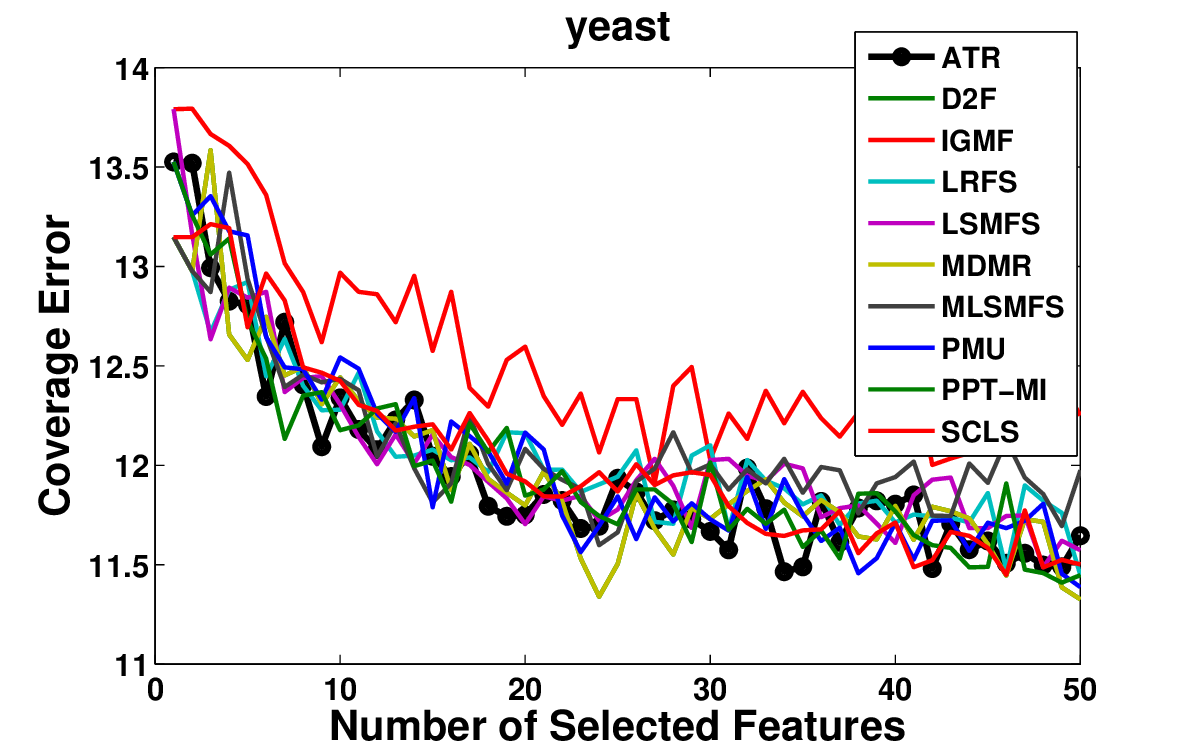}}\\
	\subfloat{\includegraphics[width=2.2in,width=2.2in,height = 1.6in ]{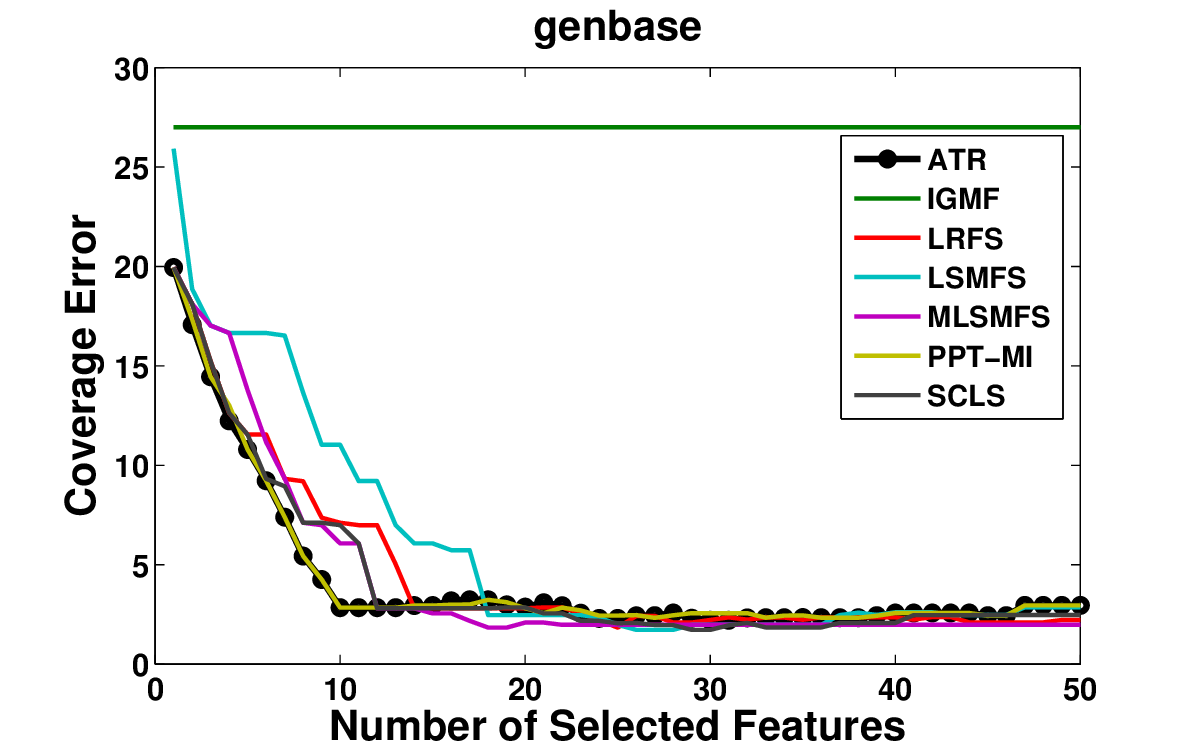}}
	\subfloat{\includegraphics[width=2.2in,height = 1.6in ]{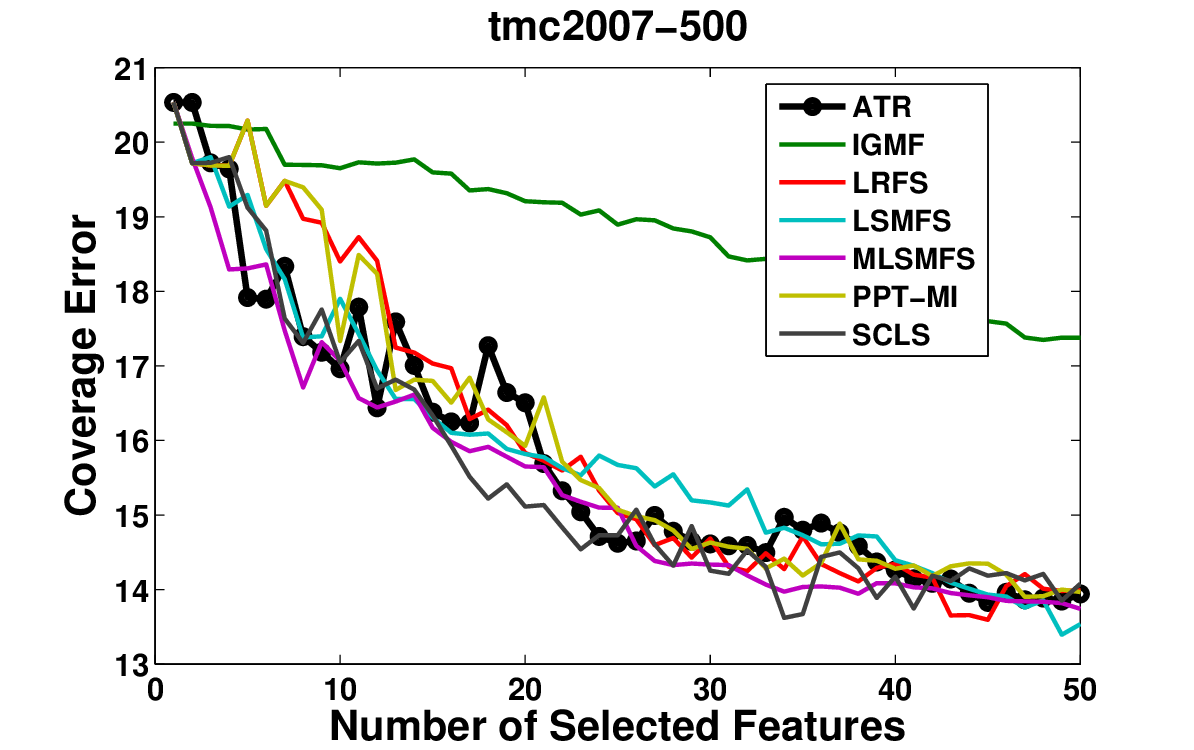}}
	\subfloat{\includegraphics[width=2.2in,height = 1.6in ]{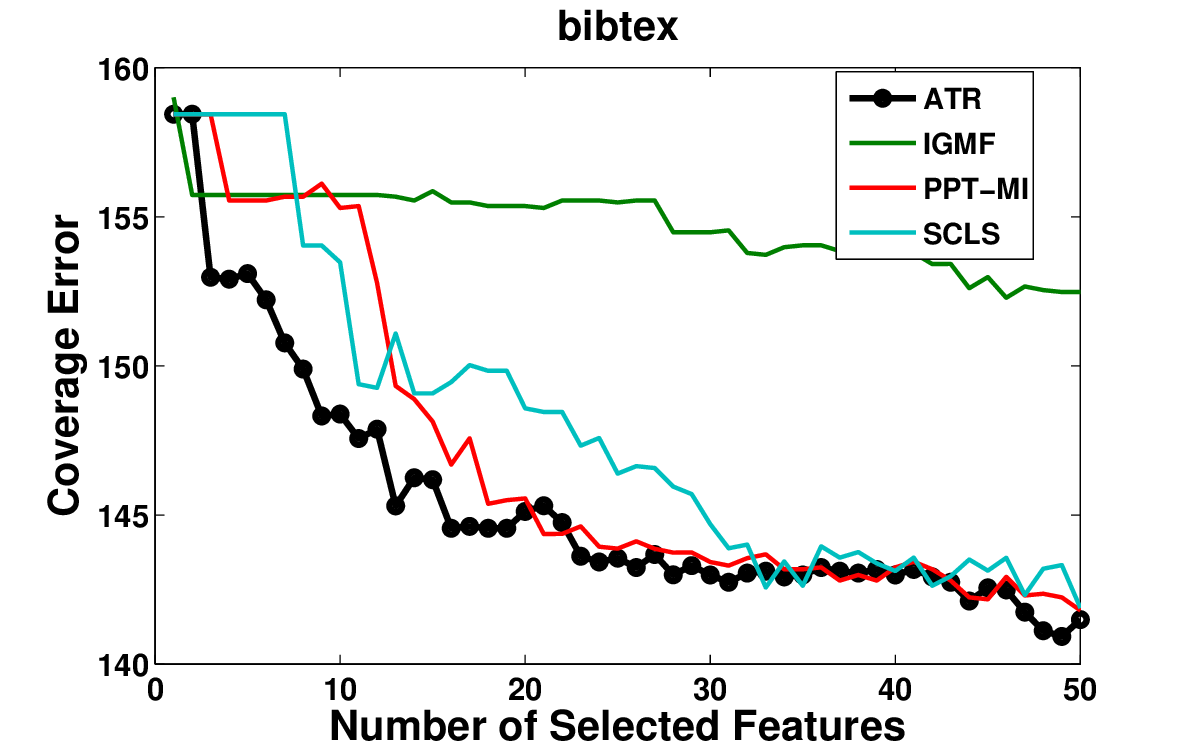}}\\
	\subfloat{\includegraphics[width=2.2in,height = 1.6in ]{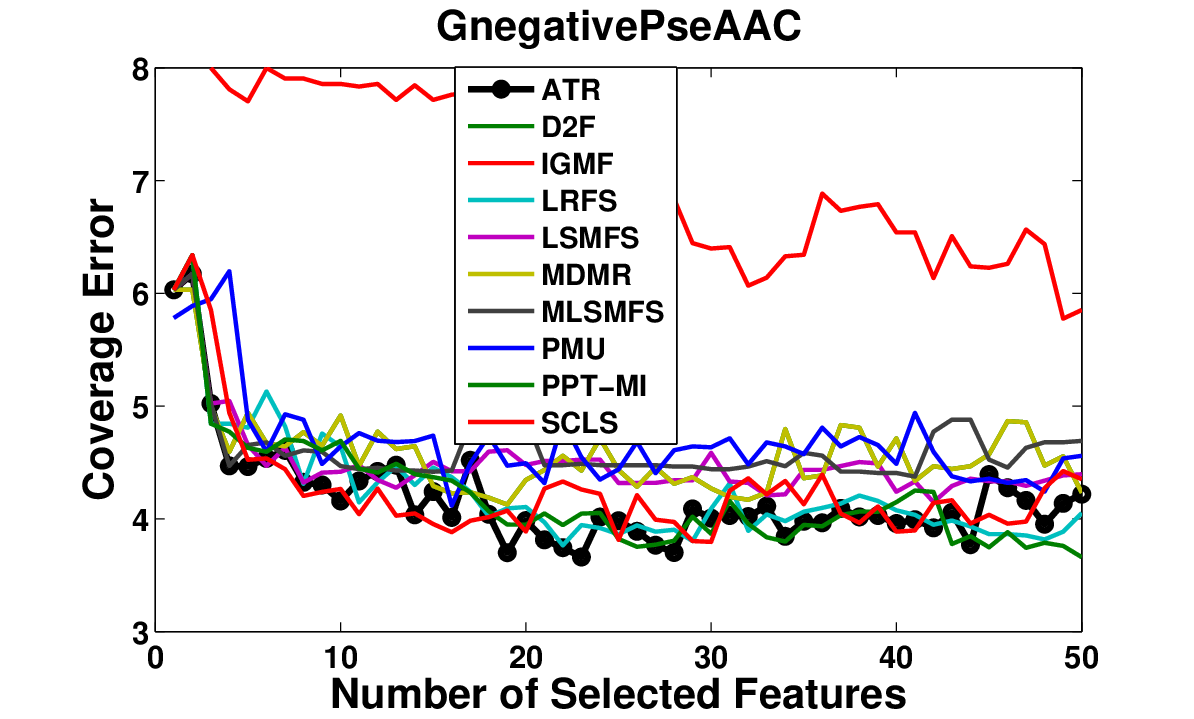}}
	\subfloat{\includegraphics[width=2.2in,height = 1.6in ]{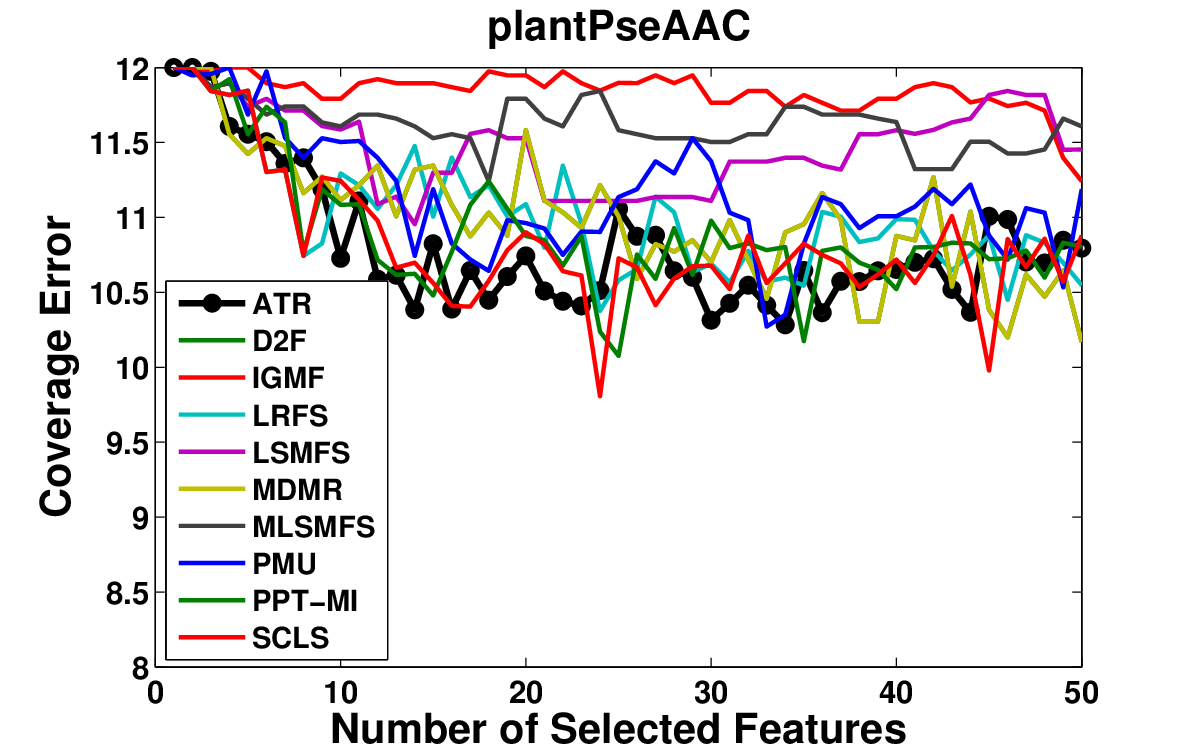}}	
	\subfloat{\includegraphics[width=2.2in,height = 1.6in ]{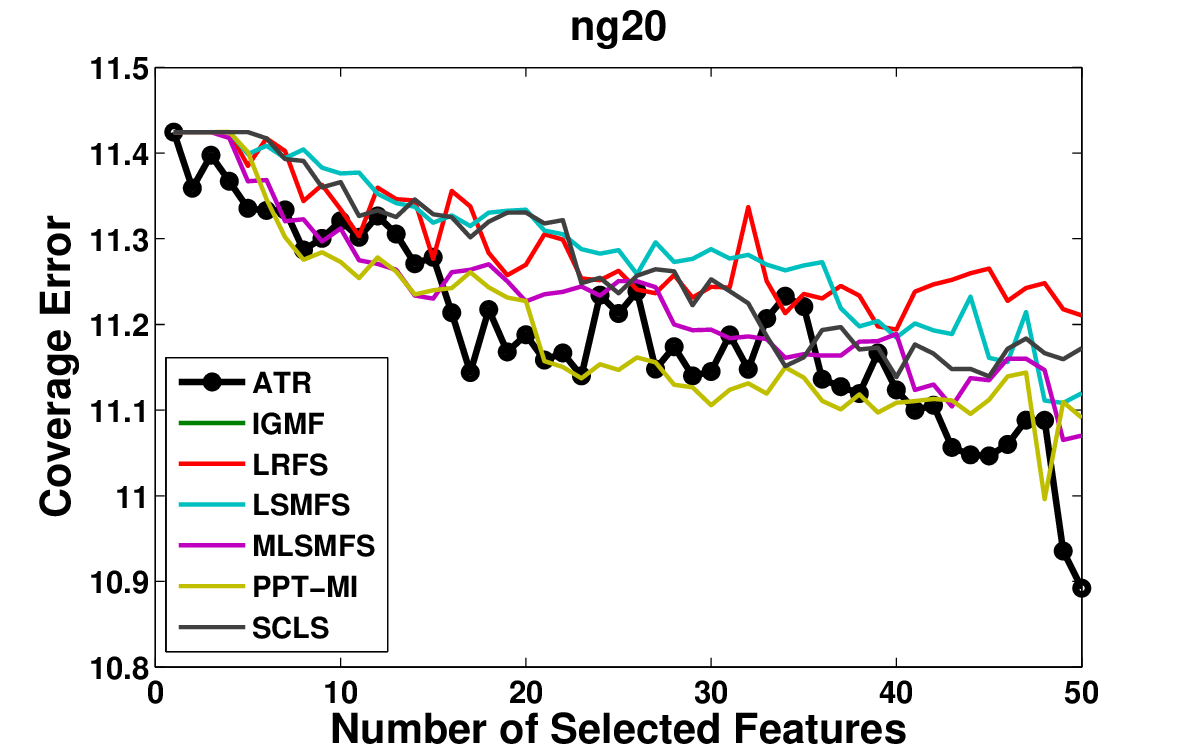}}
	\caption{Comparison of MLKNN Coverage Error for $N=1,2, \dots 50$ using the ten MLFS algorithms}
	\label{fig:mlknn_CE}
\end{figure*}

\begin{figure*}
	
	\centering
	\subfloat{\includegraphics[width=2.2in,height = 1.6in ]{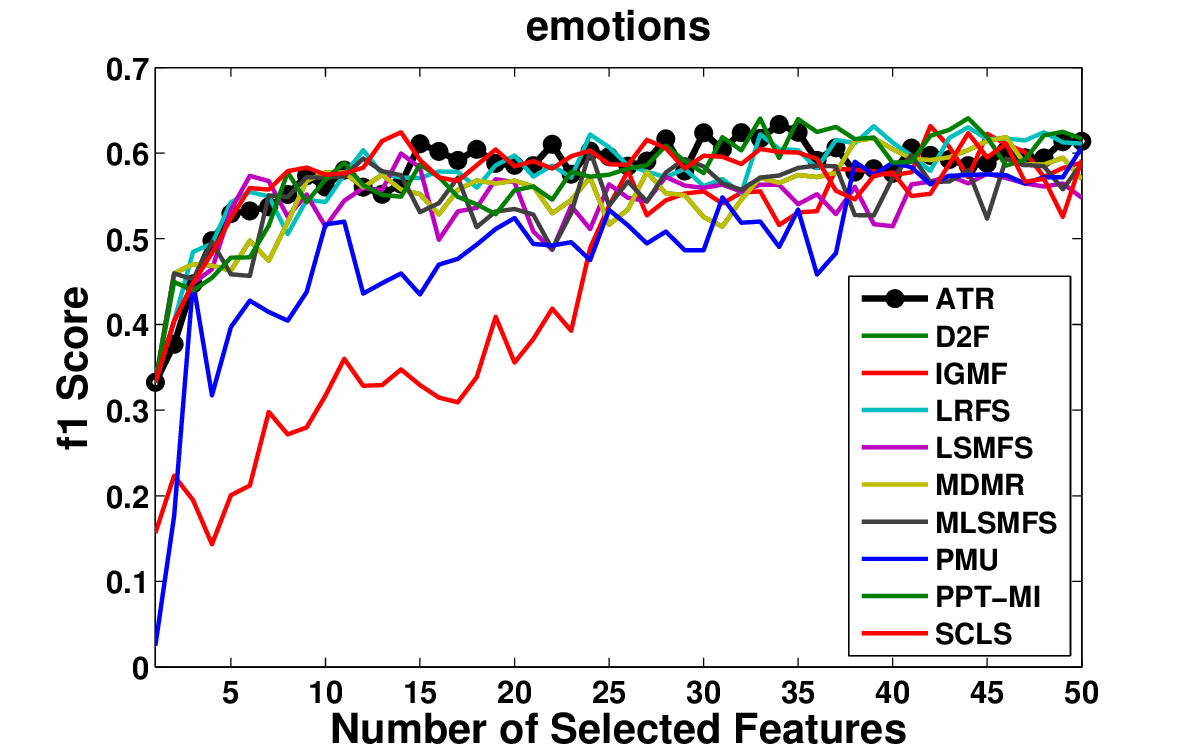}}
	\subfloat{\includegraphics[width=2.2in,height = 1.6in ]{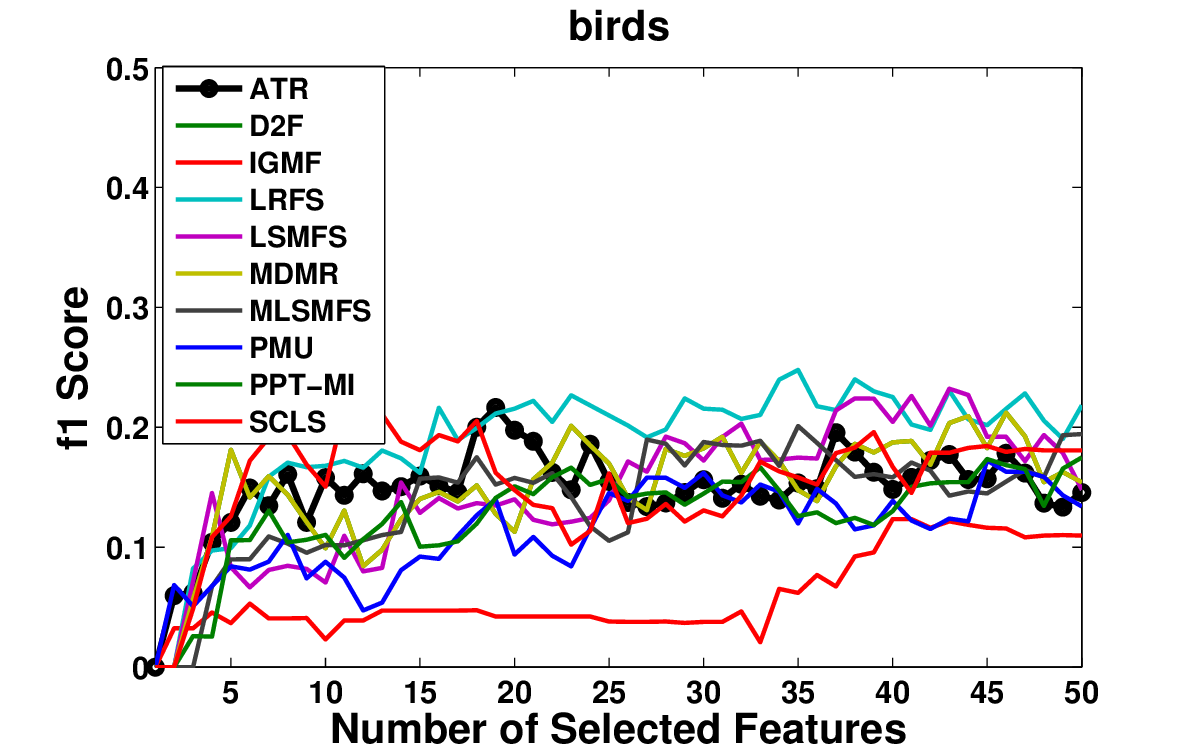}}
	\subfloat{\includegraphics[width=2.2in,height = 1.6in ]{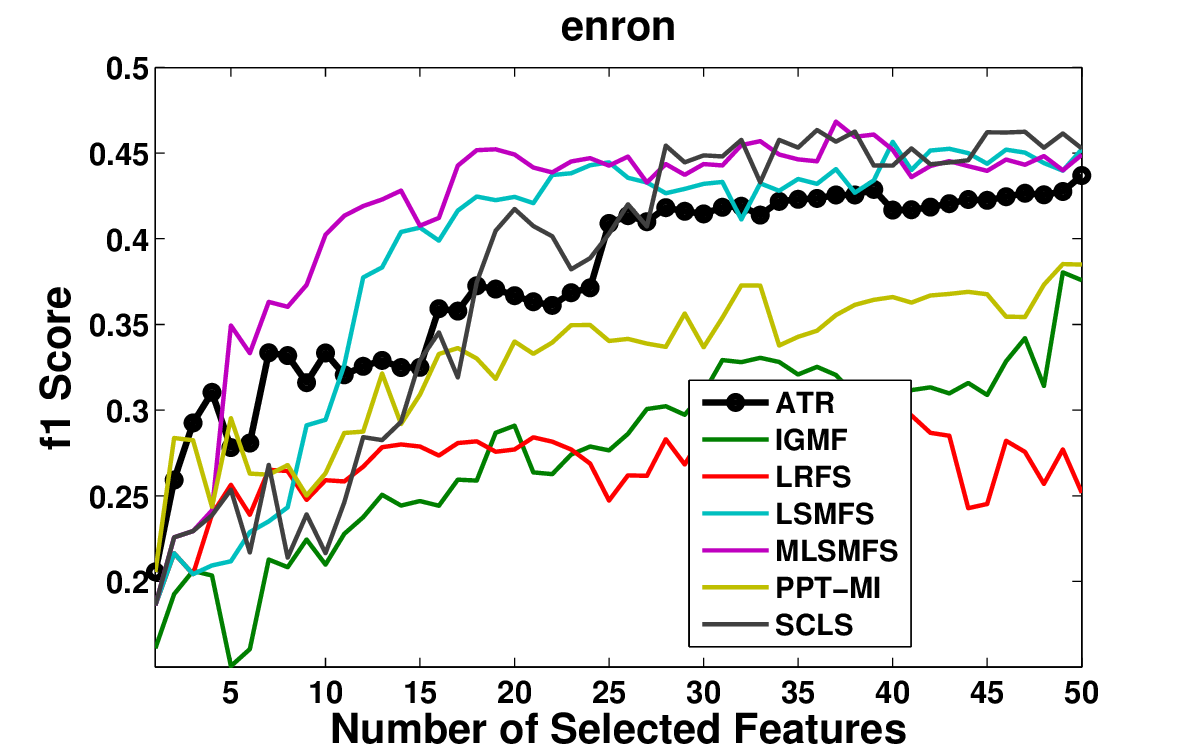}}\\
	\subfloat{\includegraphics[width=2.2in,height = 1.6in ]{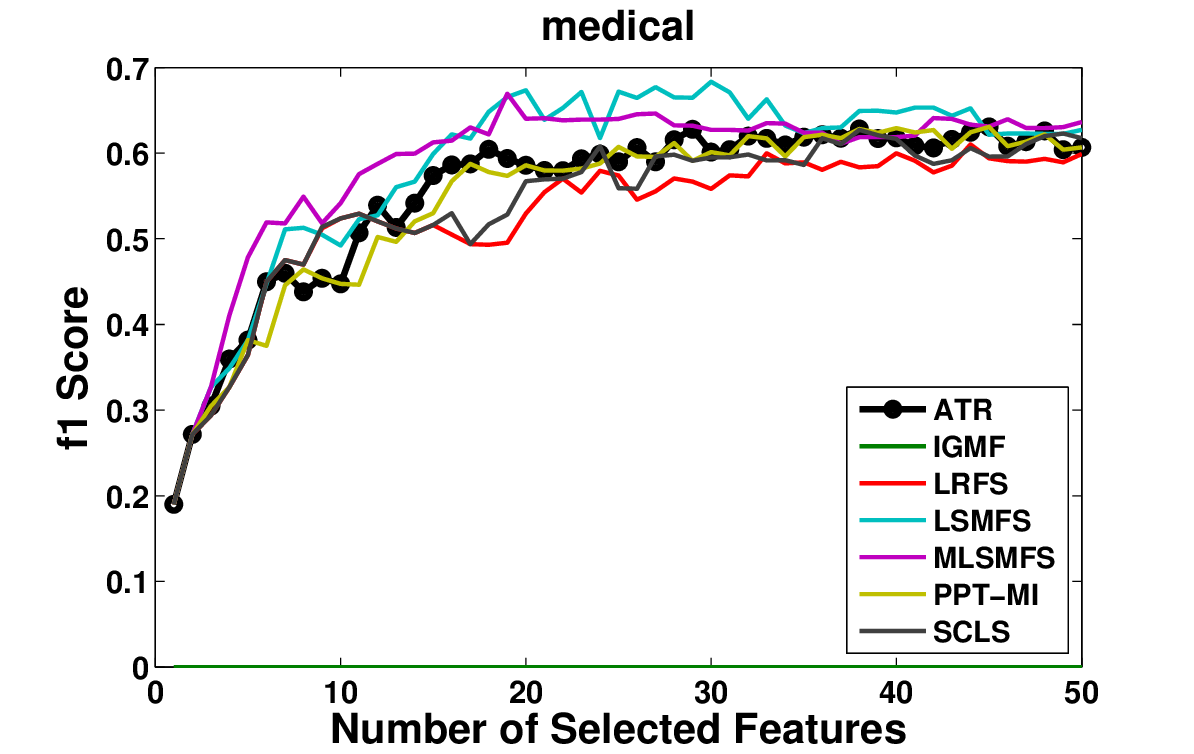}}
	\subfloat{\includegraphics[width=2.2in,height = 1.6in ]{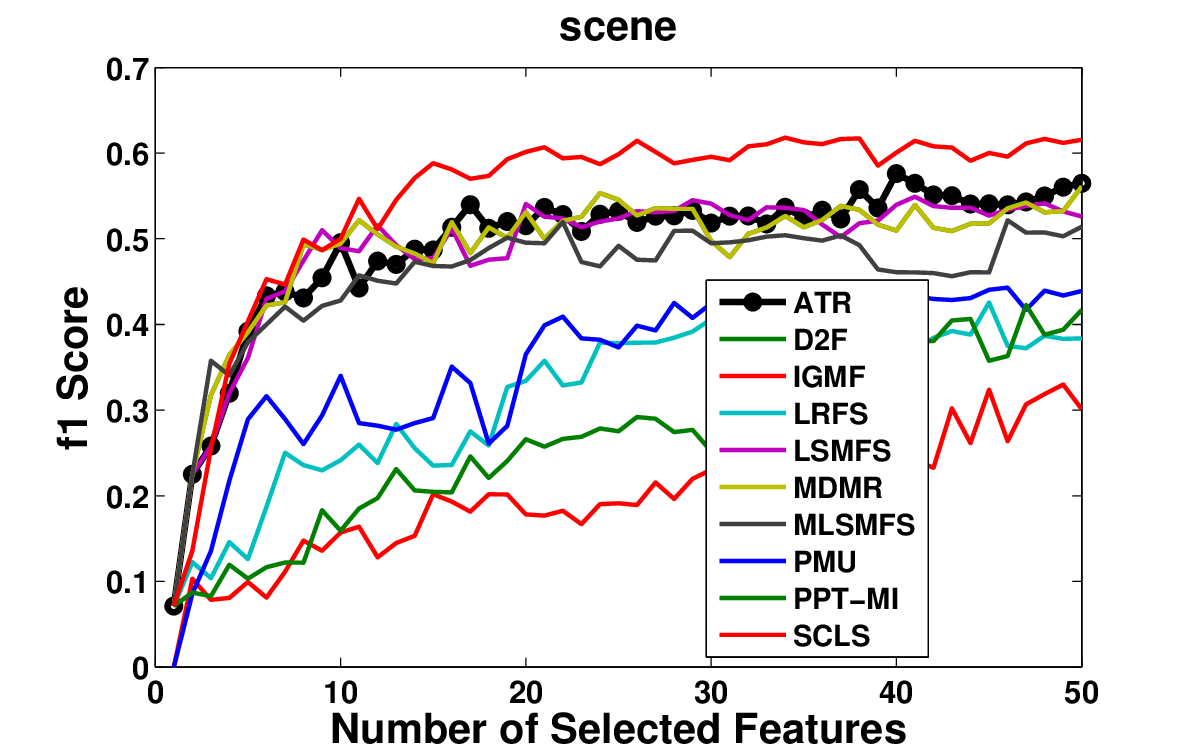}}
	\subfloat{\includegraphics[width=2.2in,height = 1.6in ]{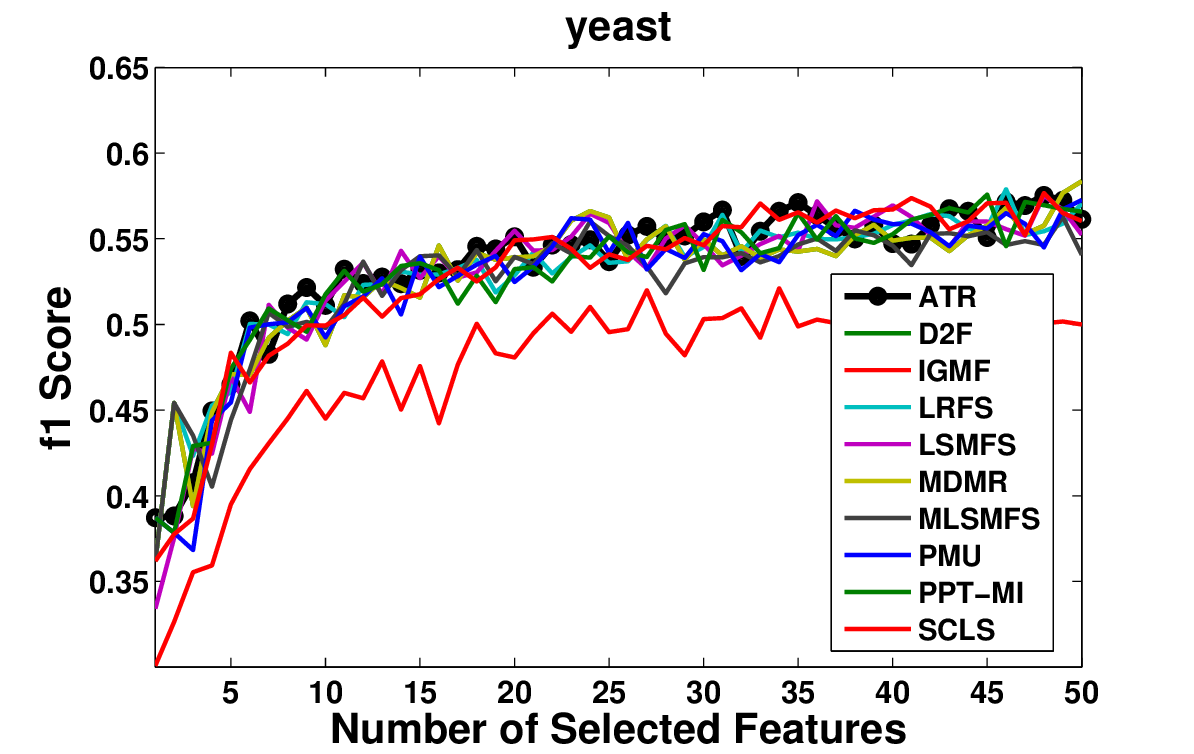}}\\
	\subfloat{\includegraphics[width=2.2in,height = 1.6in ]{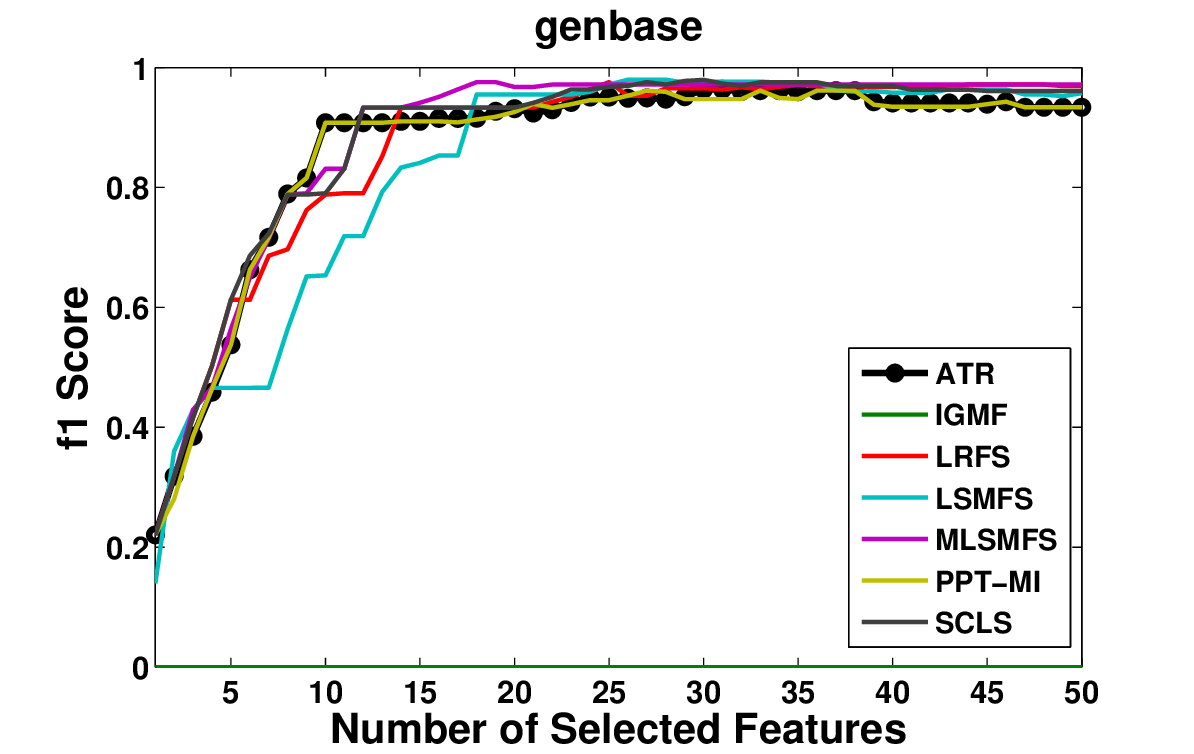}}
	\subfloat{\includegraphics[width=2.2in,height = 1.6in ]{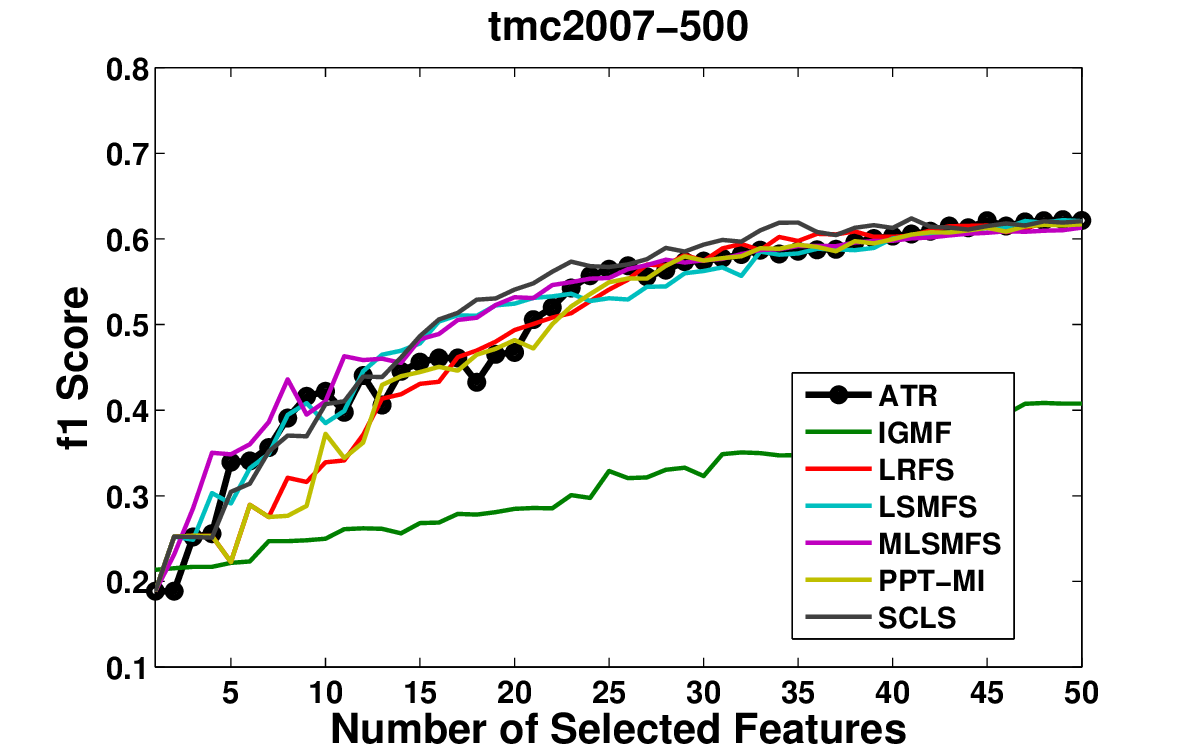}}
	\subfloat{\includegraphics[width=2.2in,height = 1.6in ]{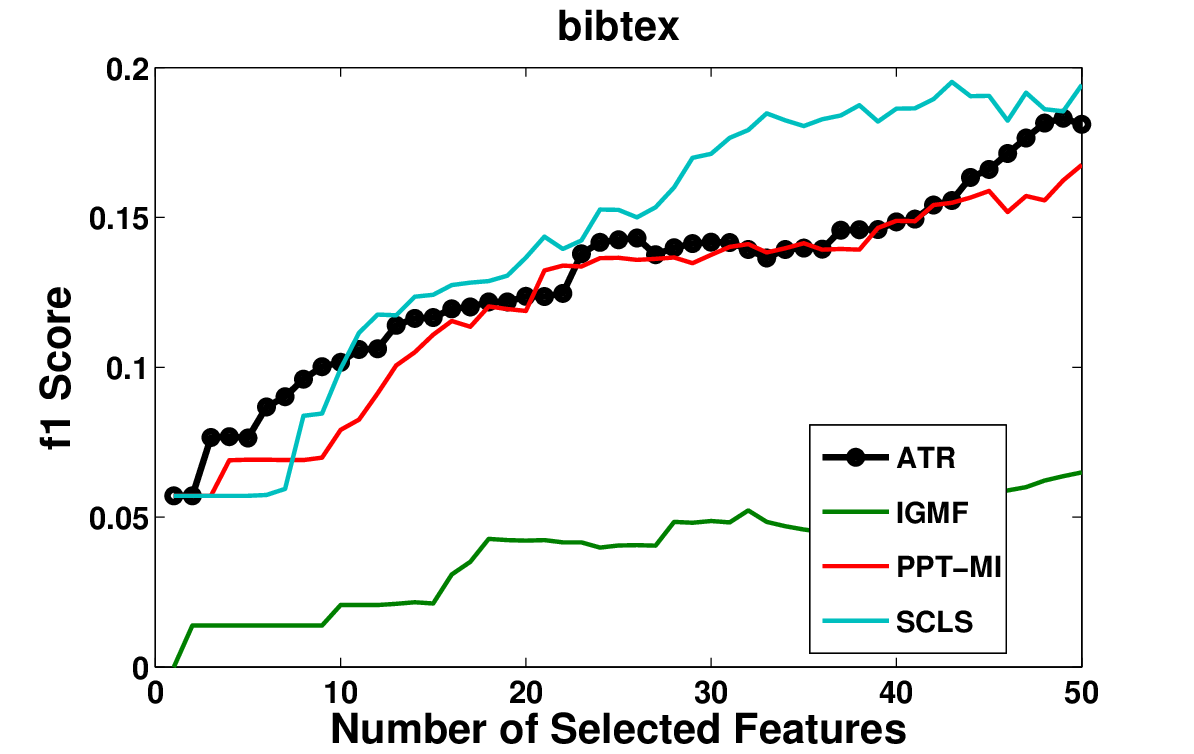}}\\
	\subfloat{\includegraphics[width=2.2in,height = 1.6in ]{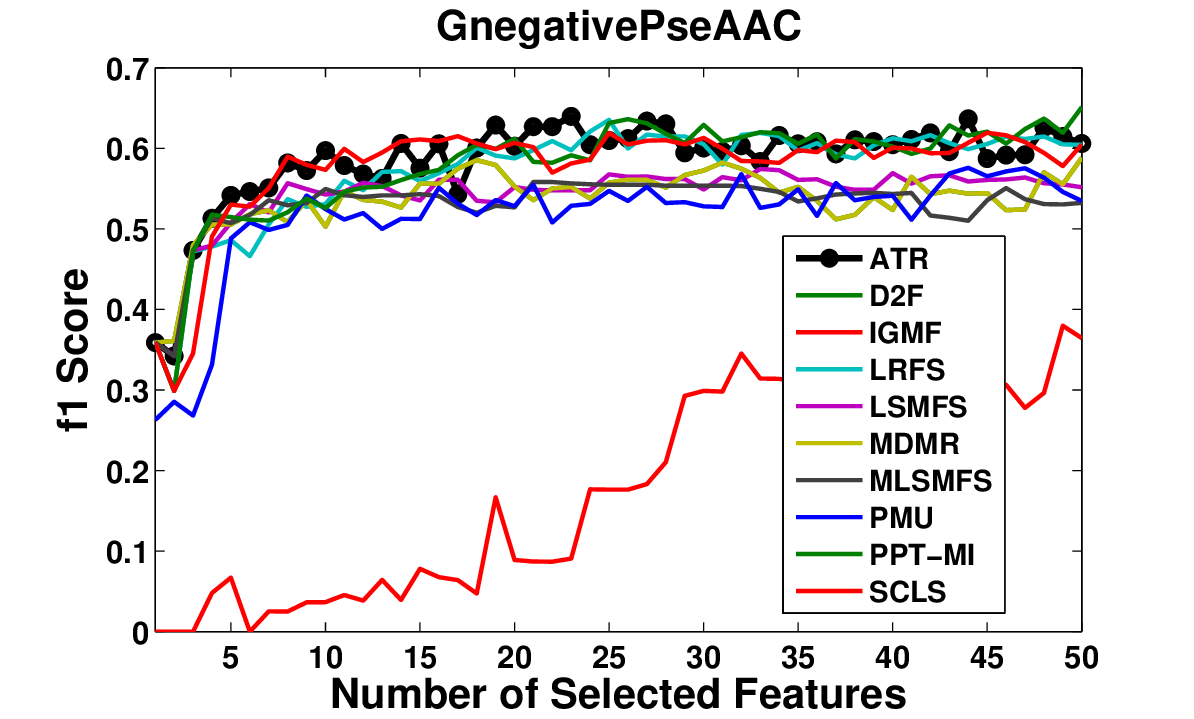}}
	\subfloat{\includegraphics[width=2.2in,height = 1.6in ]{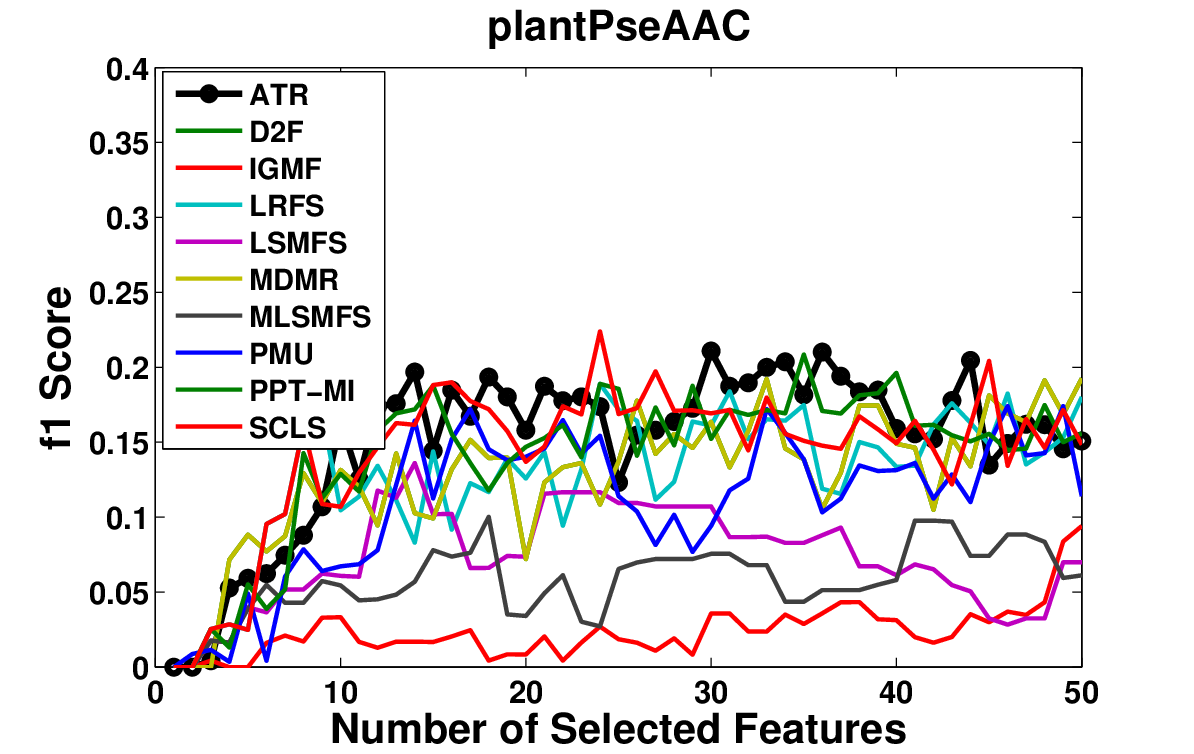}}	
	\subfloat{\includegraphics[width=2.2in,height = 1.6in ]{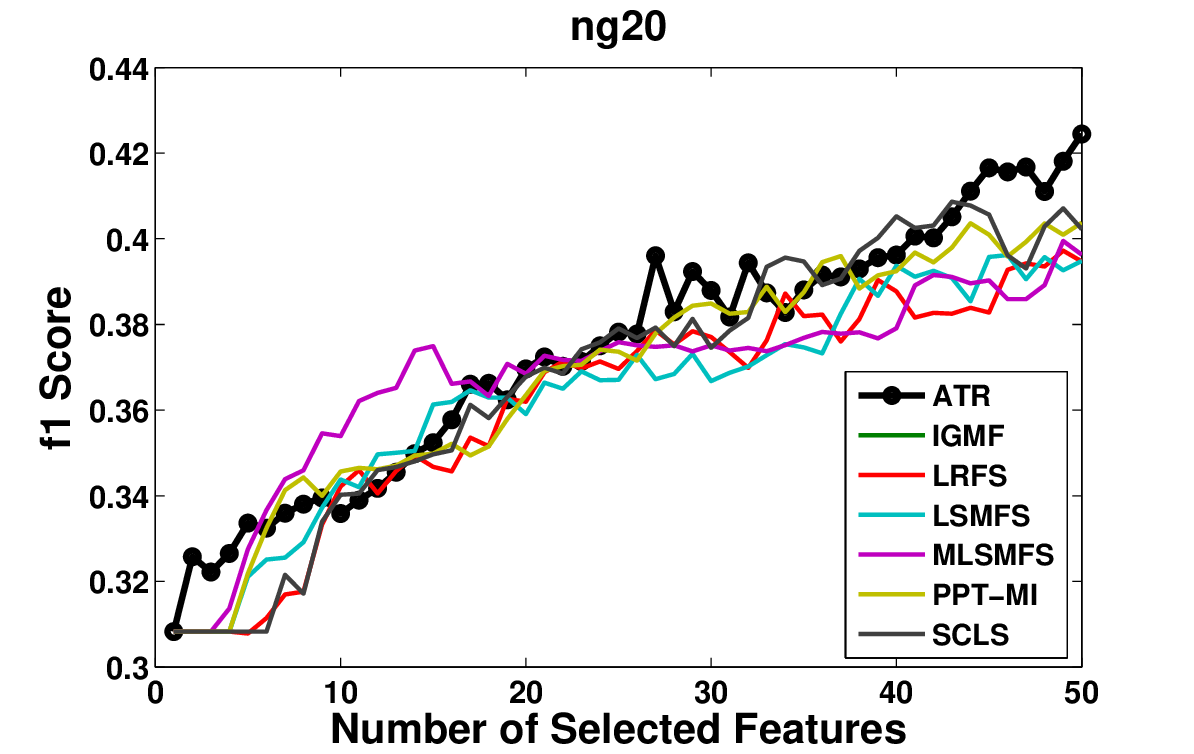}}
	\caption{Comparison of MLKNN F1-Scores for $N=1,2, \dots 50$ using the ten MLFS algorithms}
	\label{fig:mlknn_f1}
\end{figure*}

\begin{figure*}
	
	\centering
	\subfloat{\includegraphics[width=2.2in,height = 1.6in ]{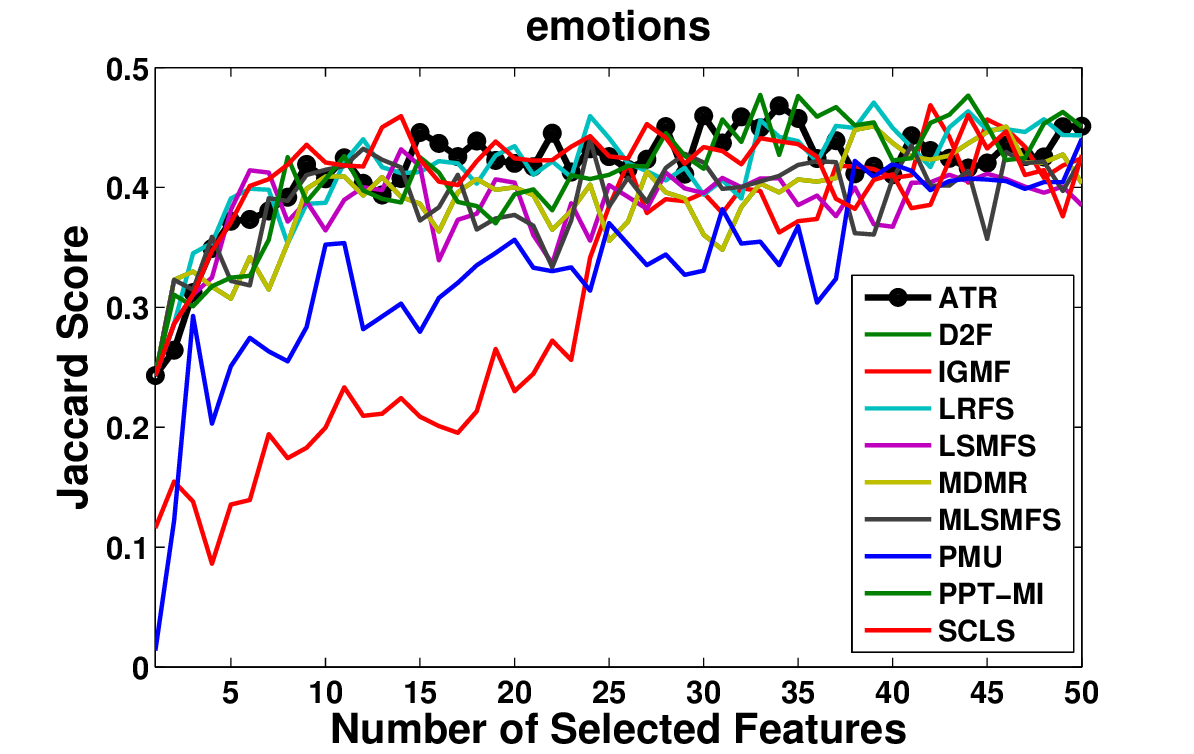}}
	\subfloat{\includegraphics[width=2.2in,height = 1.6in ]{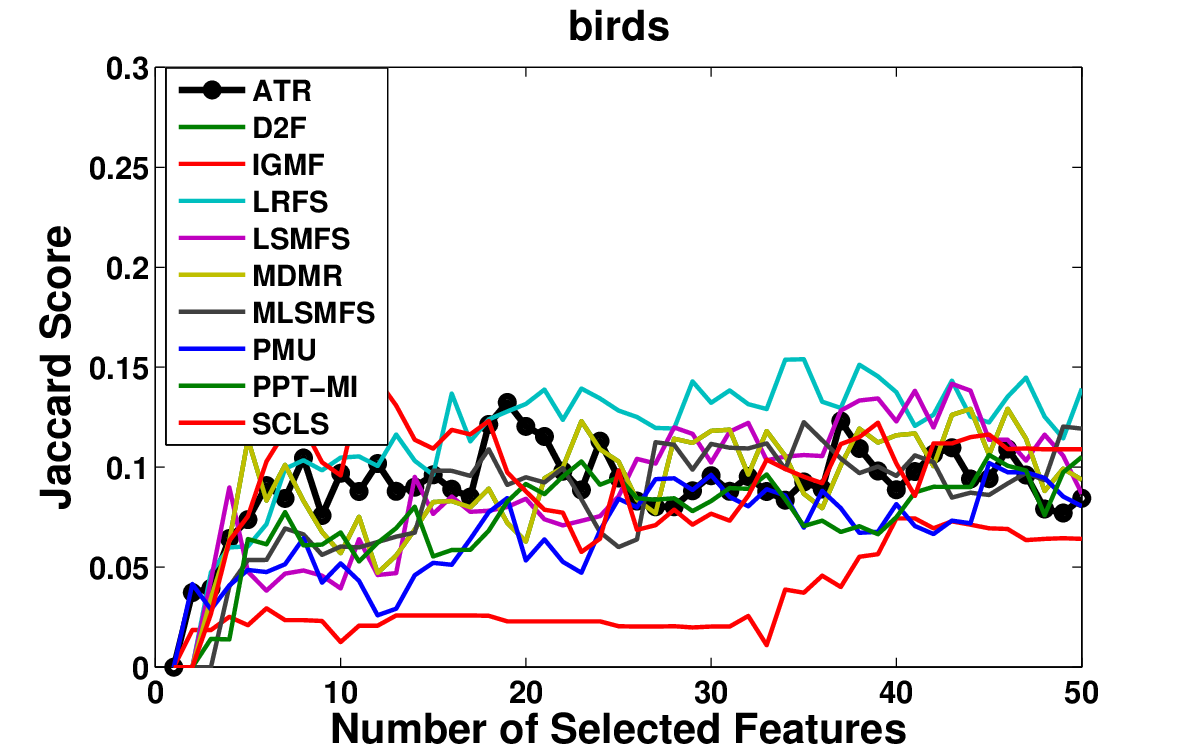}}
	\subfloat{\includegraphics[width=2.2in,height = 1.6in ]{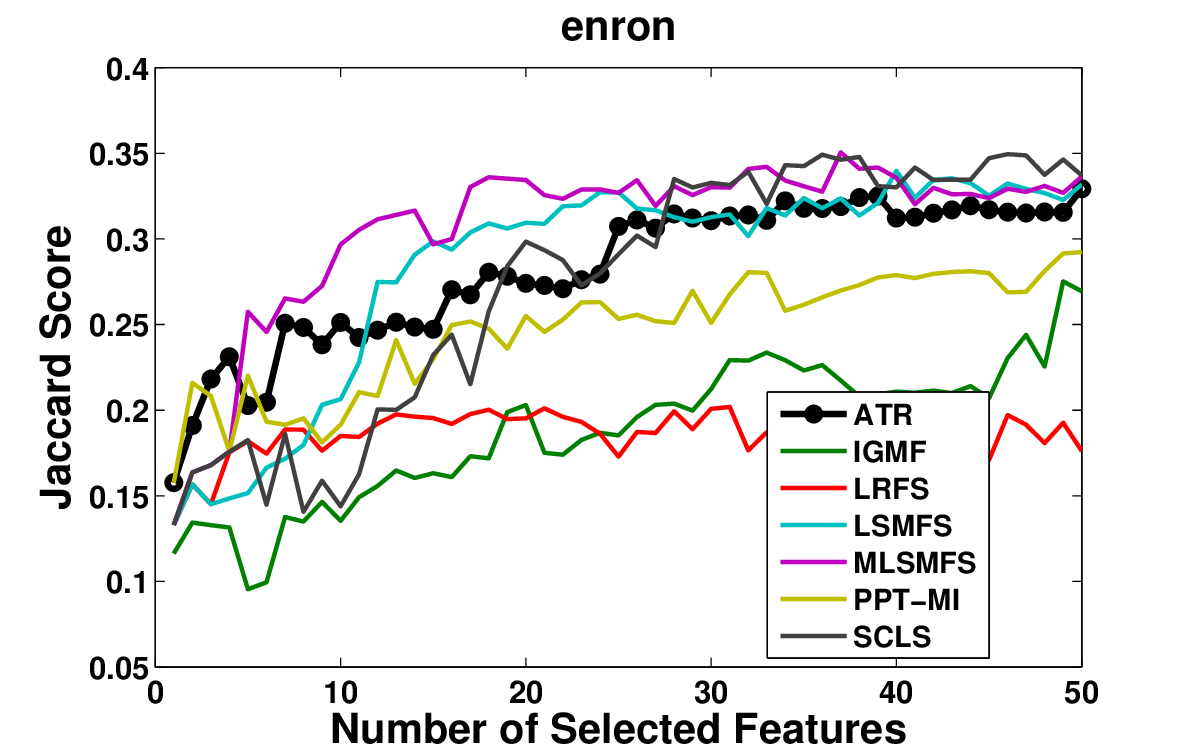}}\\
	\subfloat{\includegraphics[width=2.2in,height = 1.6in ]{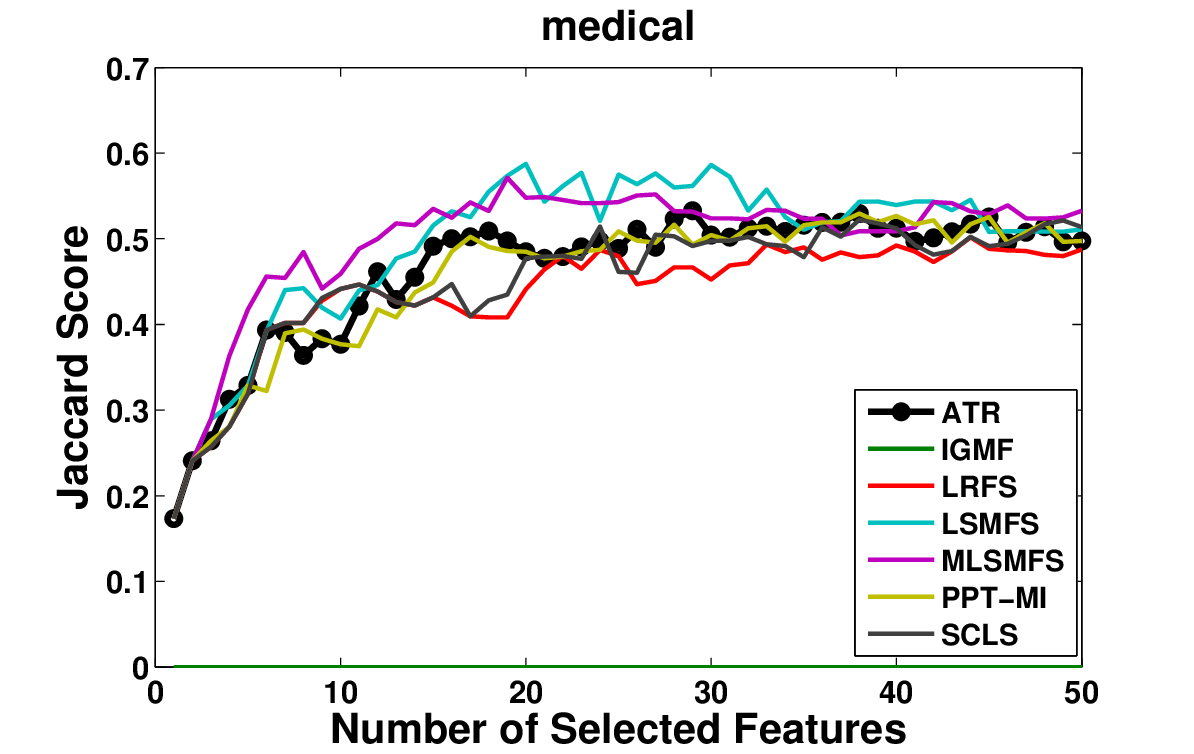}}
	\subfloat{\includegraphics[width=2.2in,height = 1.6in ]{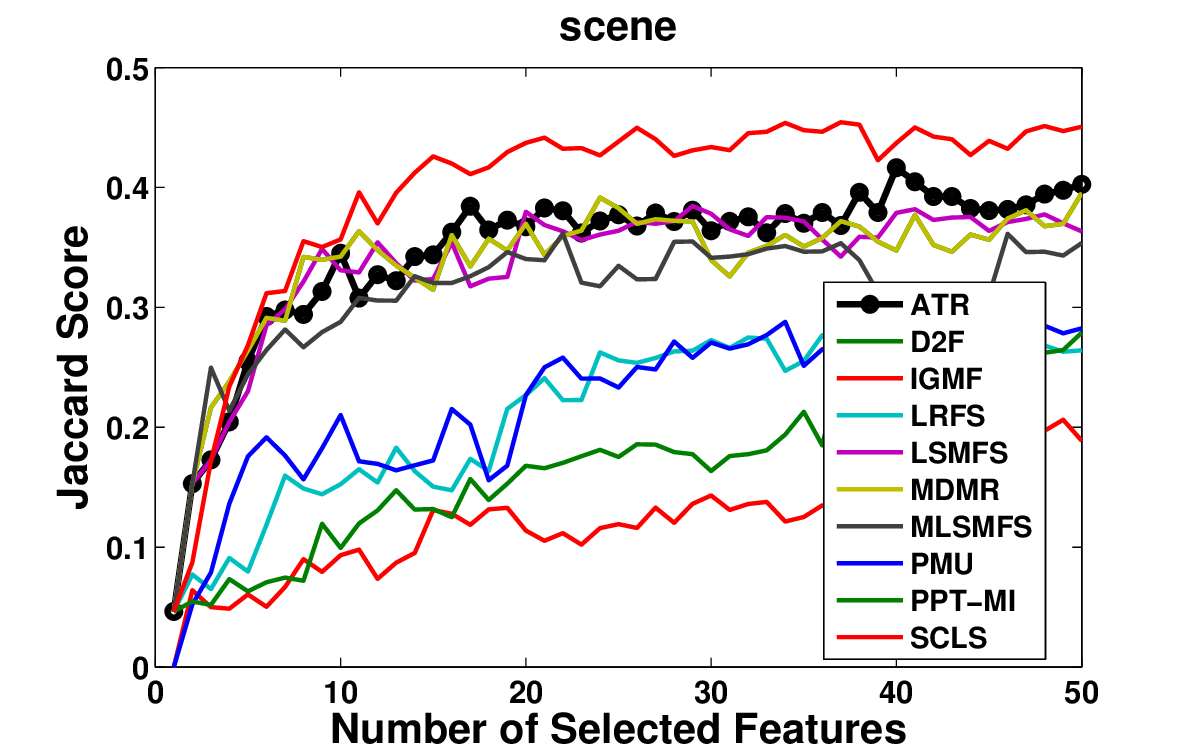}}	
	\subfloat{\includegraphics[width=2.2in,height = 1.6in ]{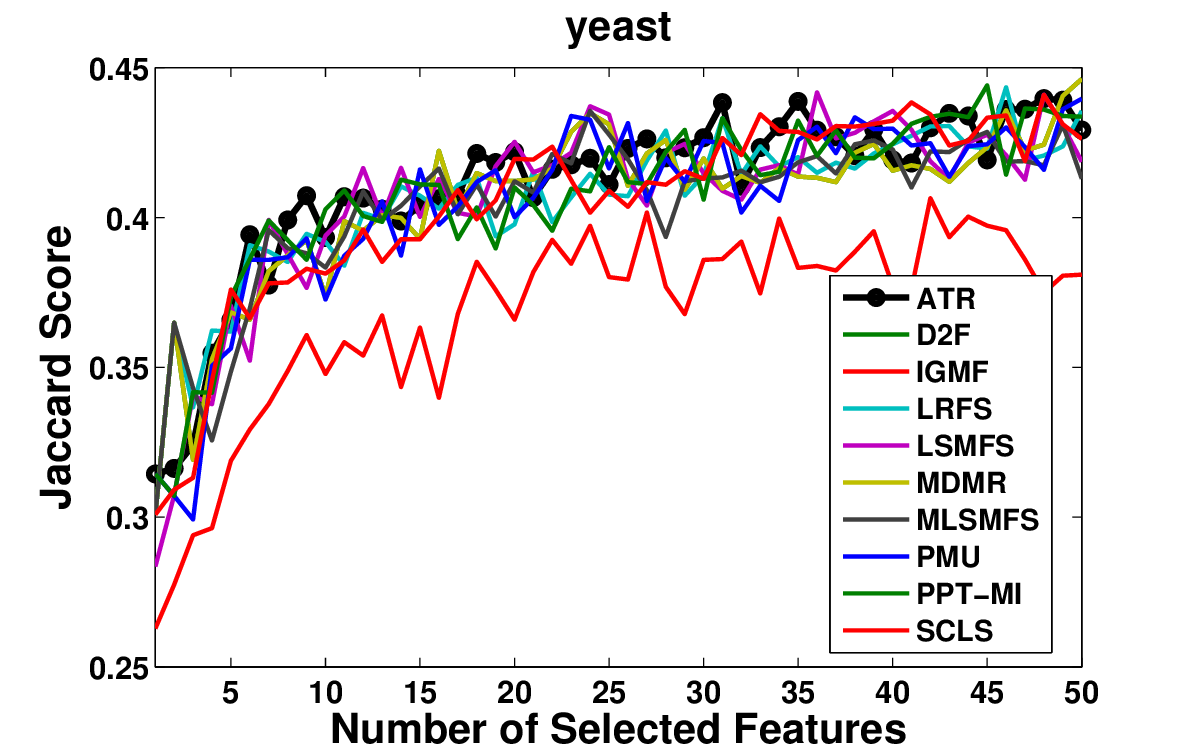}}\\
	\subfloat{\includegraphics[width=2.2in,height = 1.6in ]{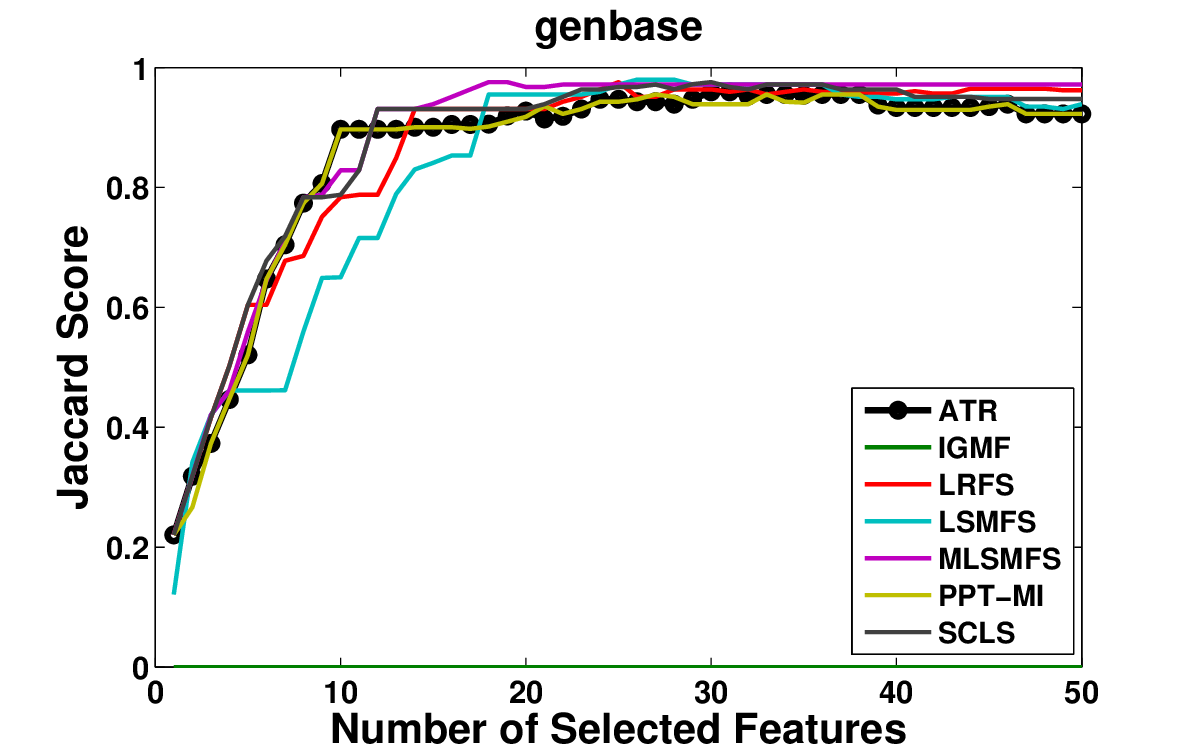}}
	\subfloat{\includegraphics[width=2.2in,height = 1.6in ]{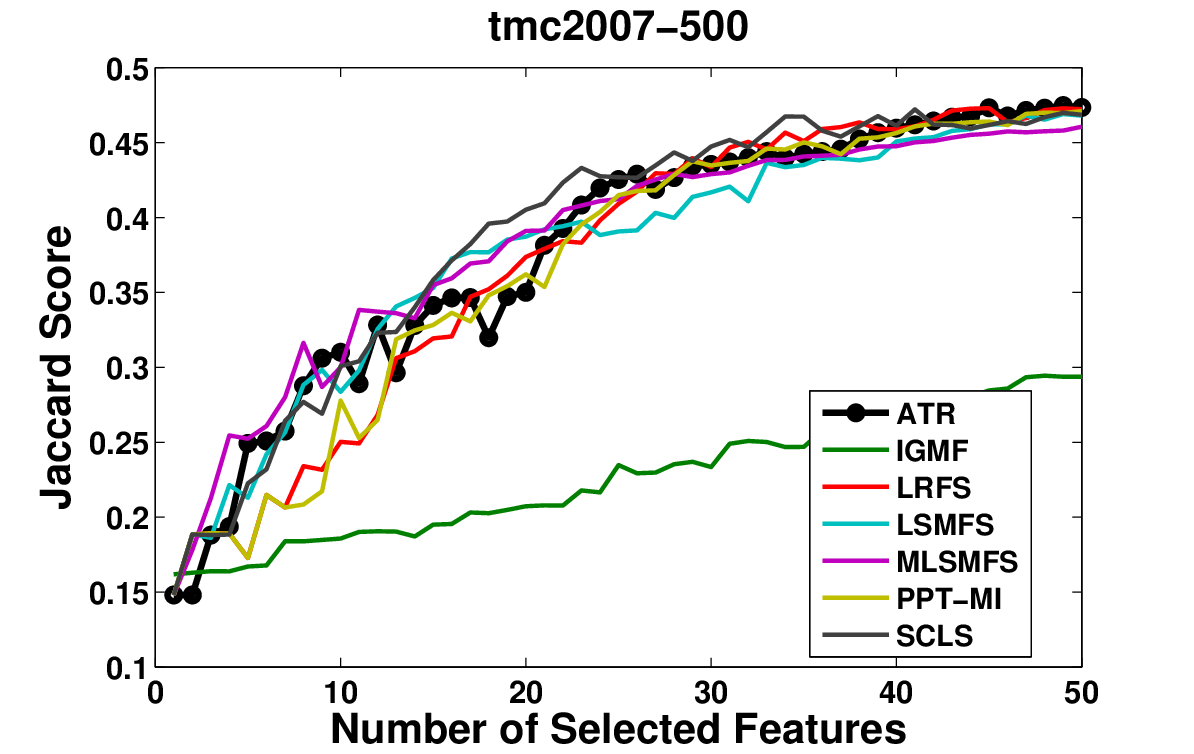}}
	\subfloat{\includegraphics[width=2.2in,height = 1.6in ]{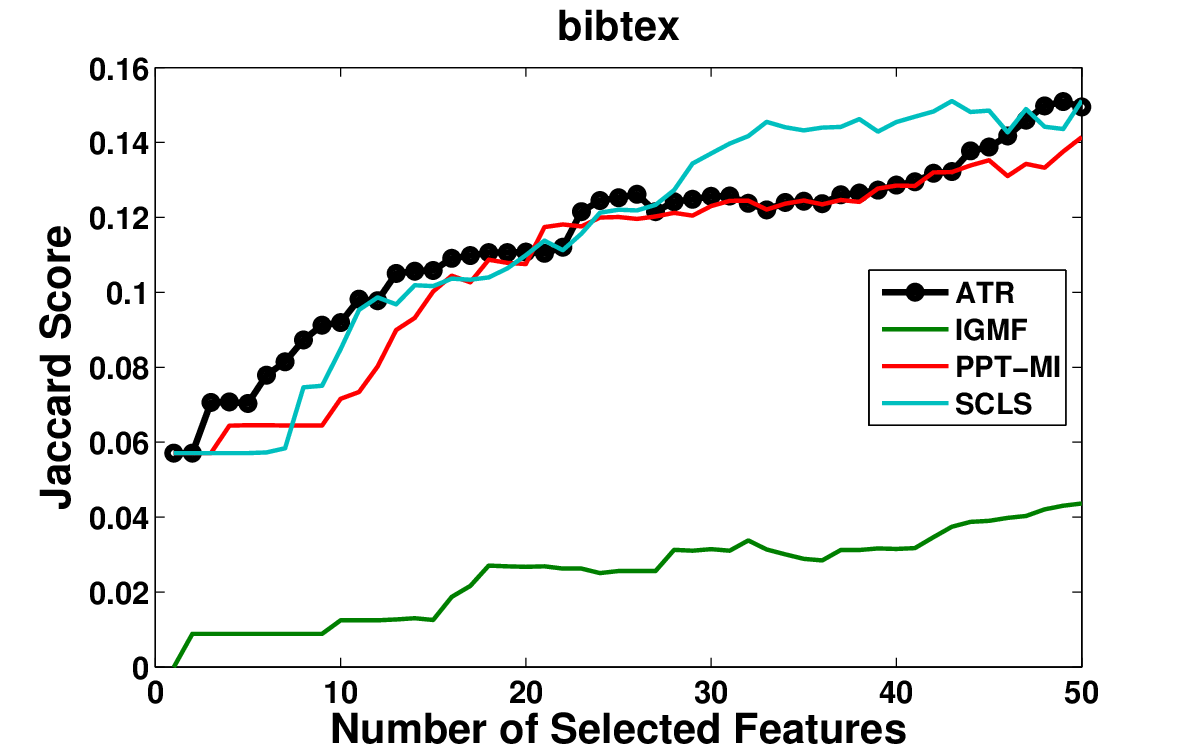}}\\
	\subfloat{\includegraphics[width=2.2in,height = 1.6in ]{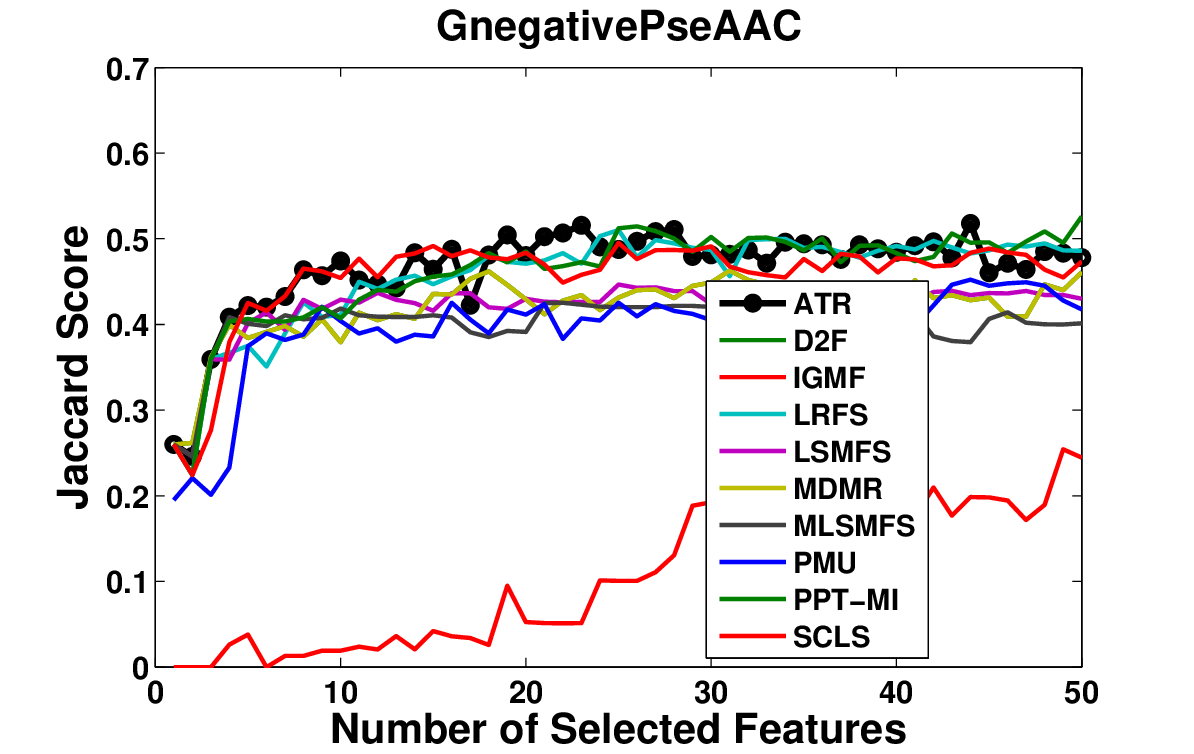}}
	\subfloat{\includegraphics[width=2.2in,height = 1.6in ]{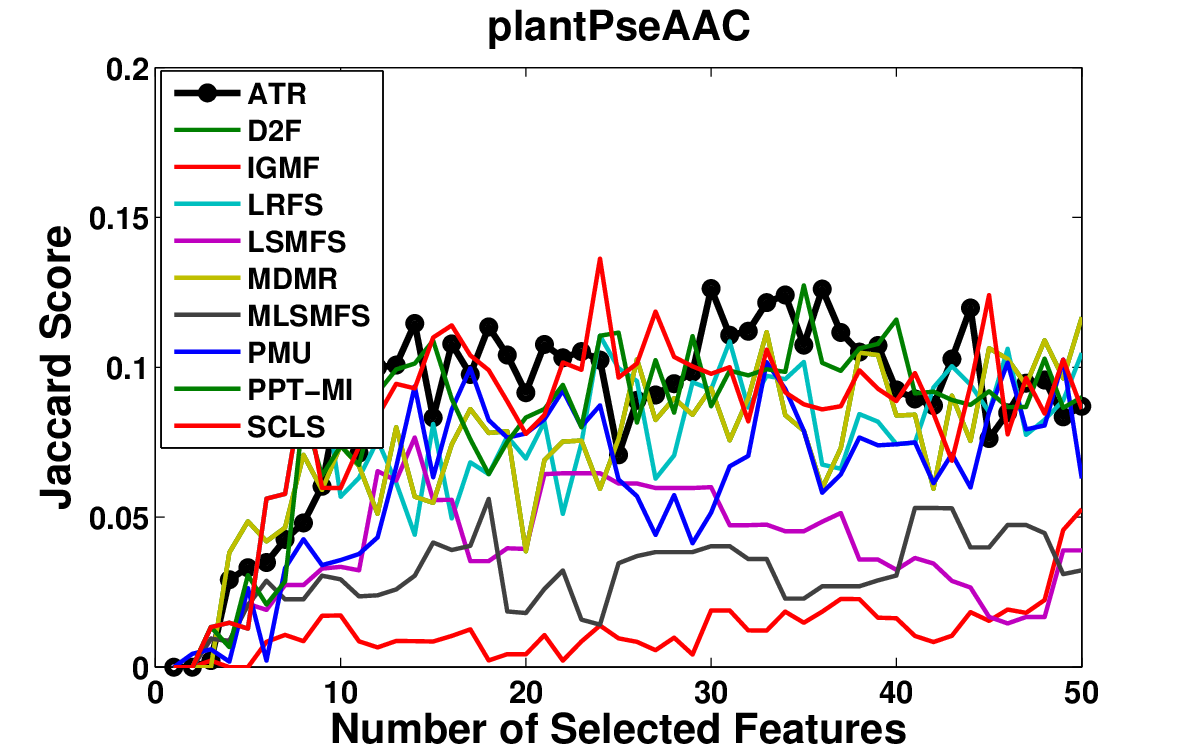}}	
	\subfloat{\includegraphics[width=2.2in,height = 1.6in ]{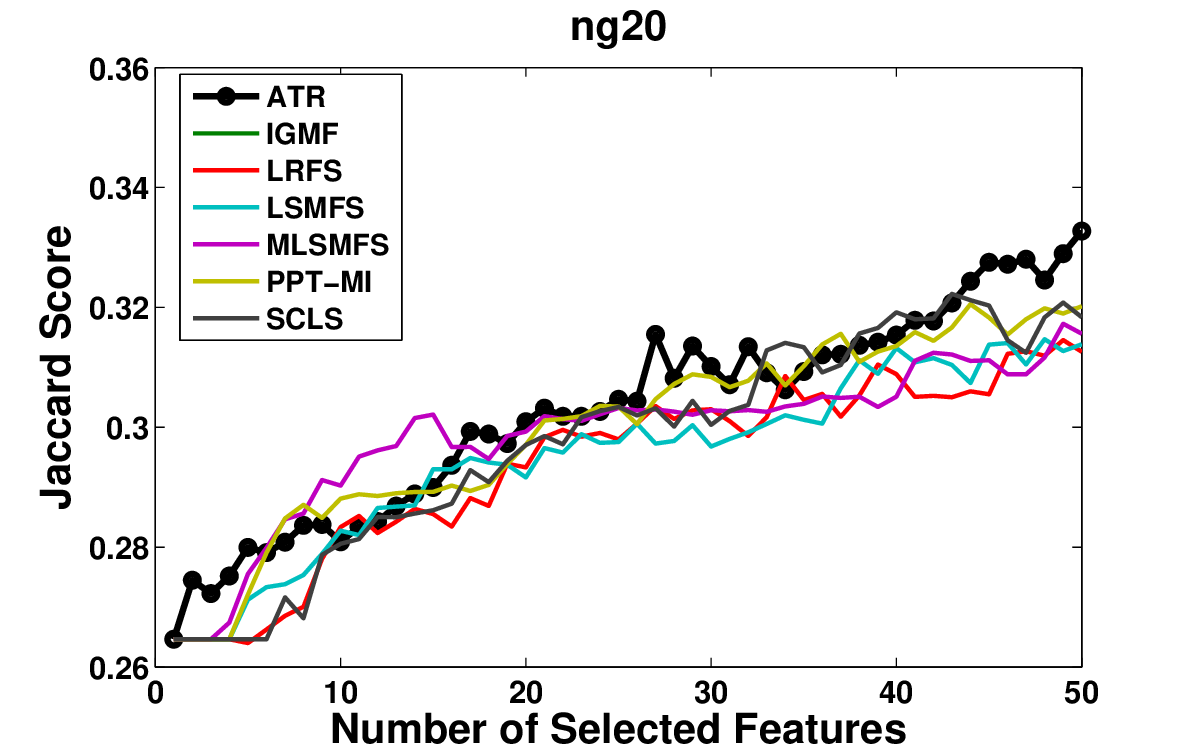}}
	
	\caption{Comparison of MLKNN Jaccard Scores for $N=1,2, \dots 50$ using the ten MLFS algorithms}
	\label{fig:mlknn_JS}
\end{figure*}

\begin{figure*}
	
	\centering
     
	\subfloat{\includegraphics[width=2.2in,height = 1.6in ]{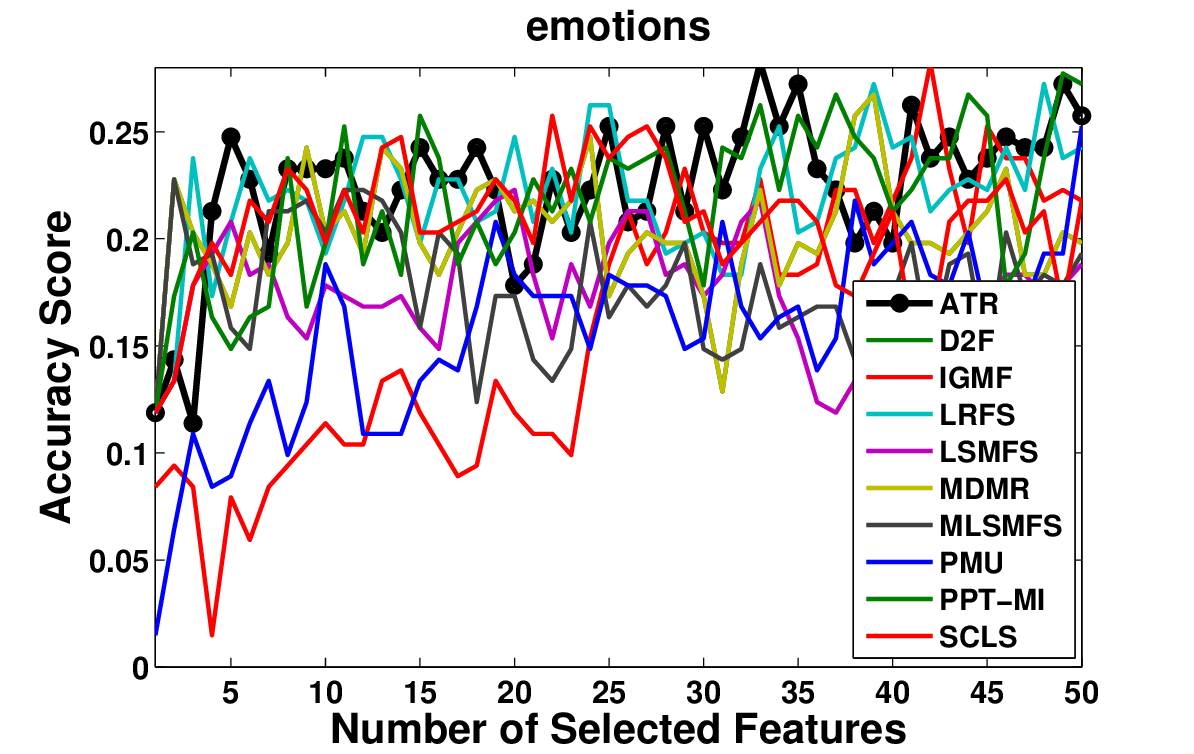}}
			\subfloat{\includegraphics[width=2.2in,height = 1.6in ]{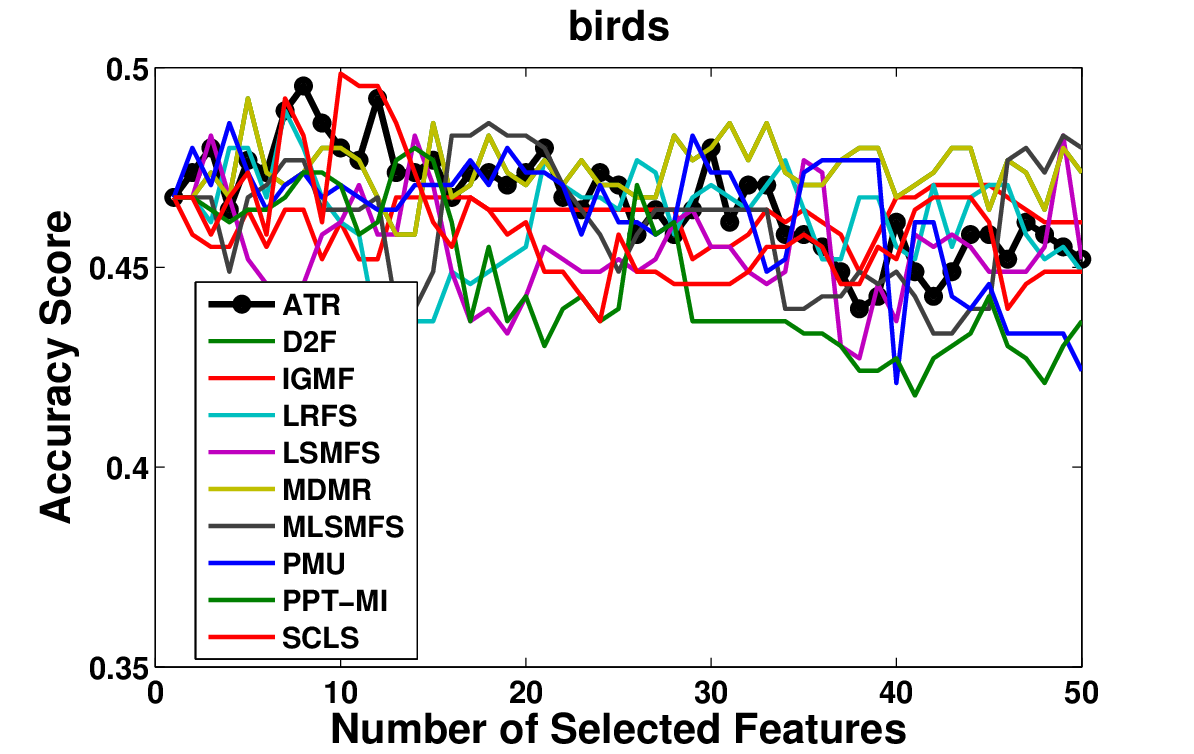}}
			\subfloat{\includegraphics[width=2.2in,height = 1.6in ]{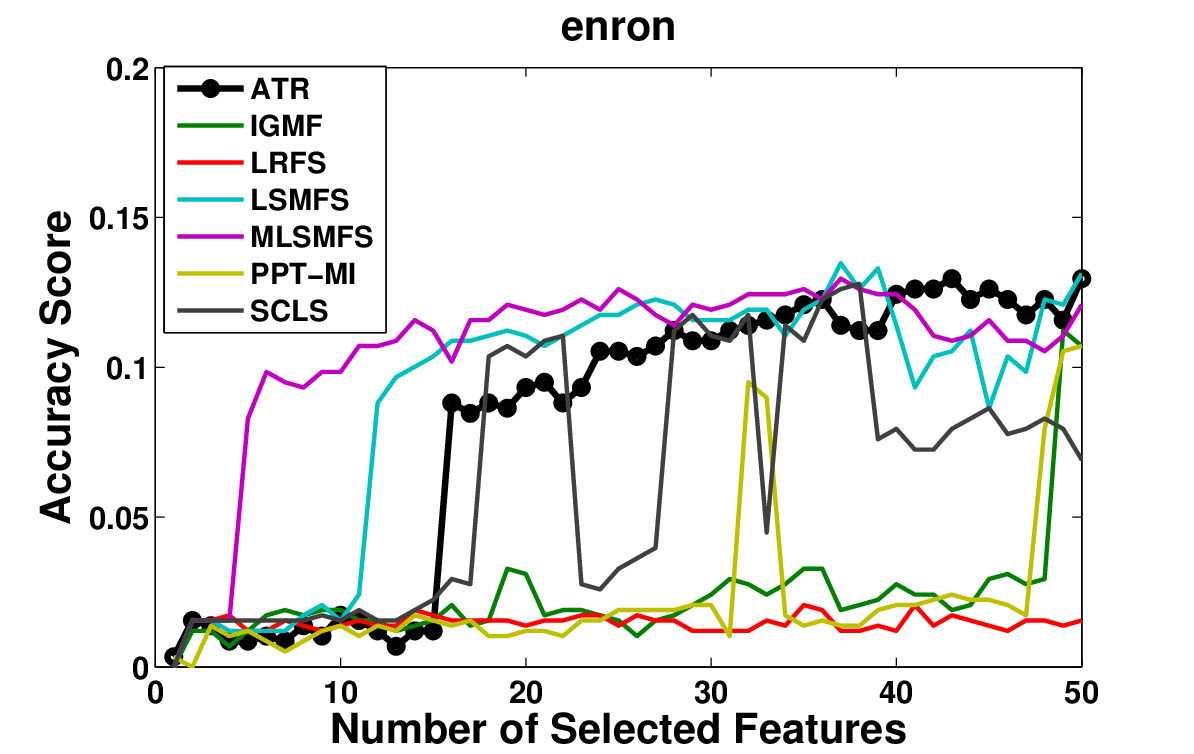}}\\
			
   \subfloat{\includegraphics[width=2.2in,height = 1.6in ]{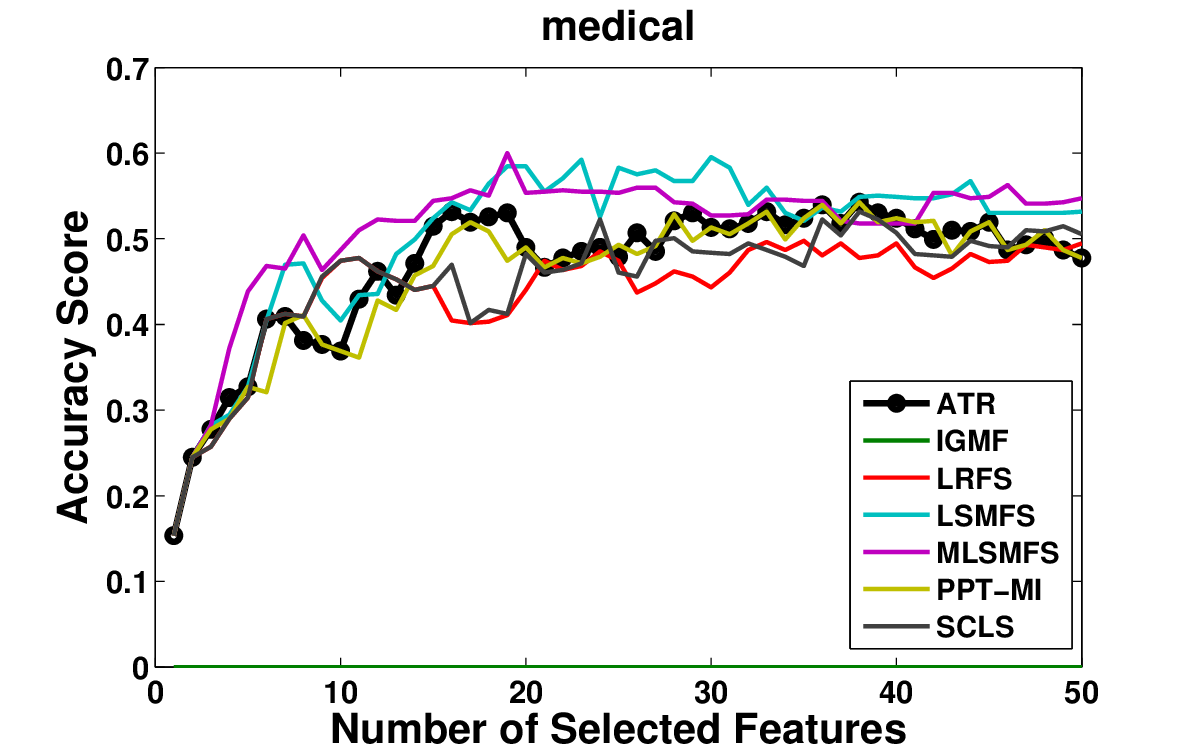}}
			\subfloat{\includegraphics[width=2.2in,height = 1.6in ]{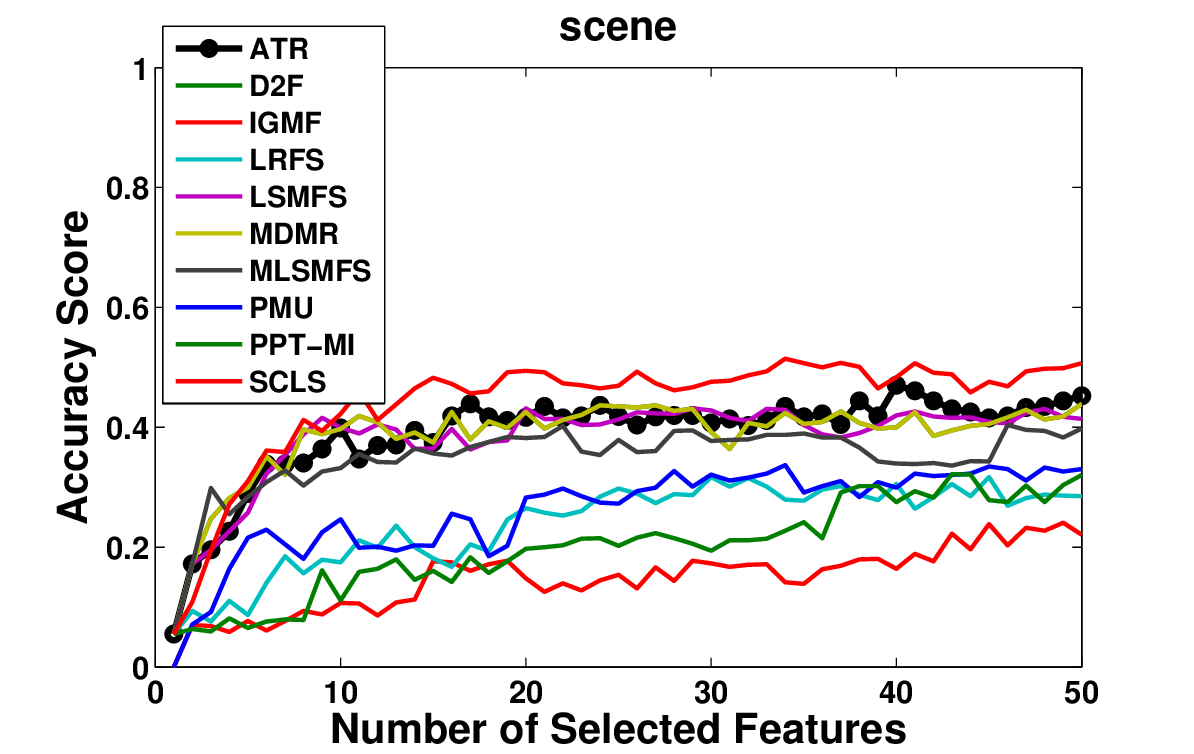}}			
			\subfloat{\includegraphics[width=2.2in,height = 1.6in ]{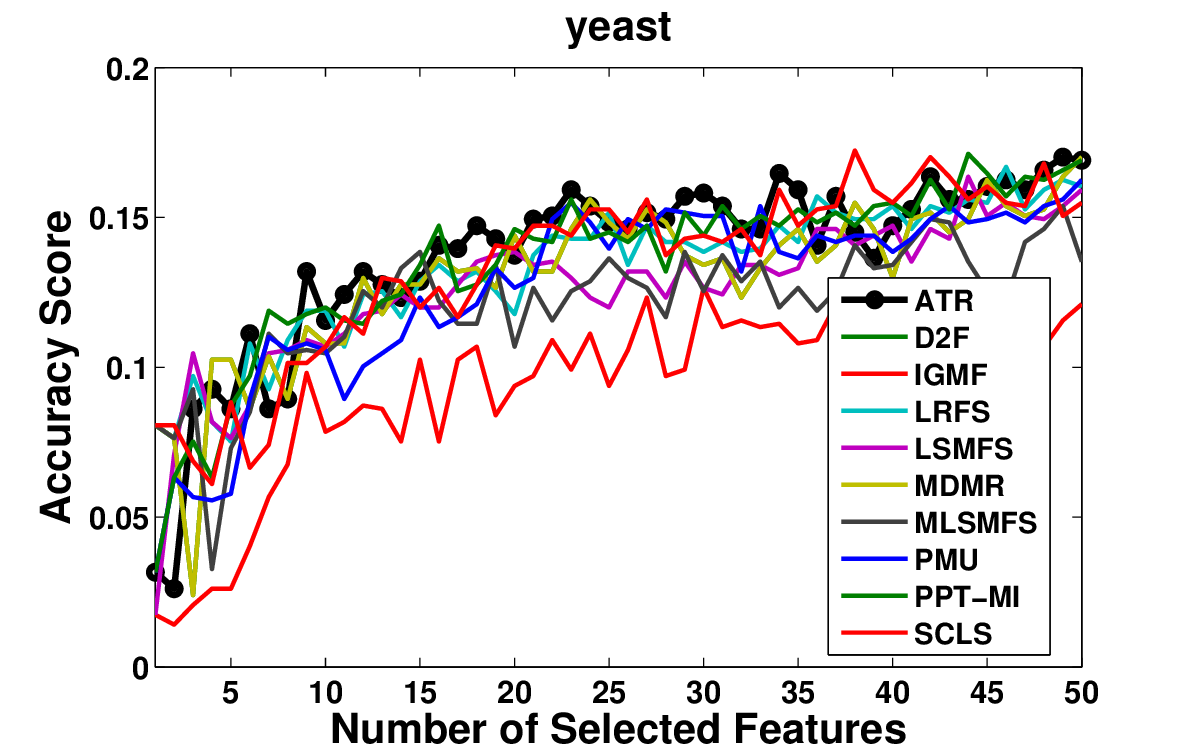}}\\
			\subfloat{\includegraphics[width=2.2in,height = 1.6in ]{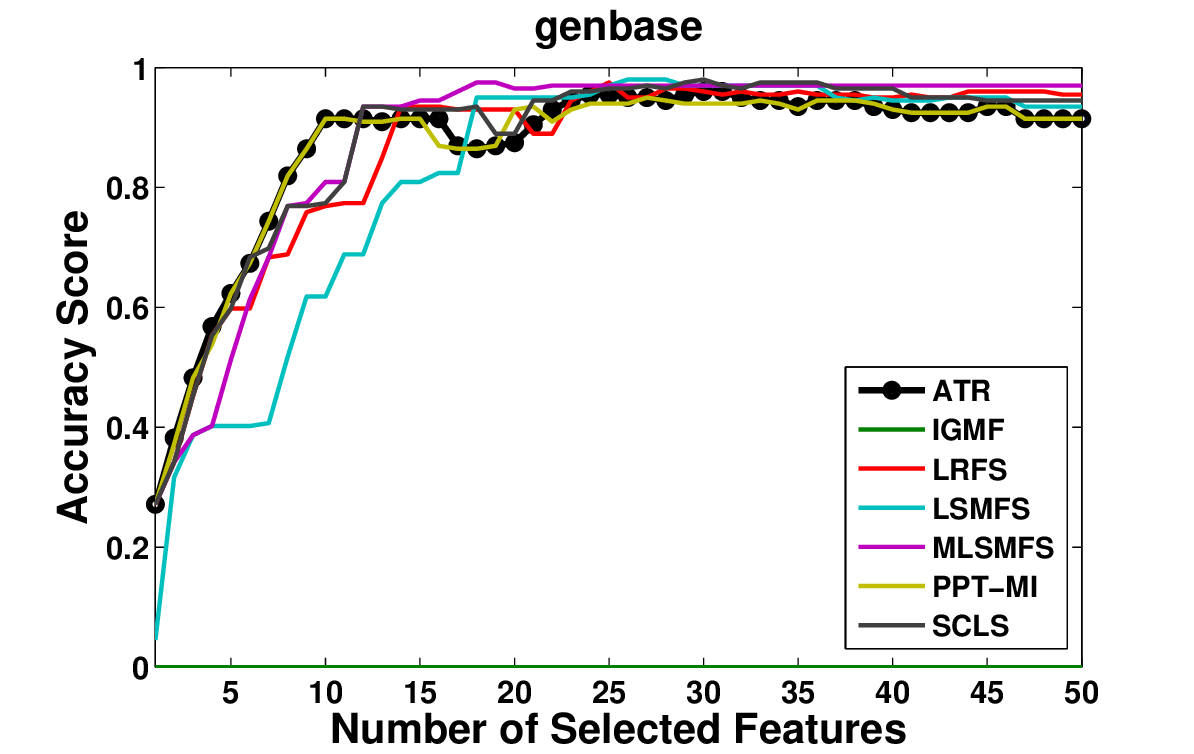}}
			\subfloat{\includegraphics[width=2.2in,height = 1.6in ]{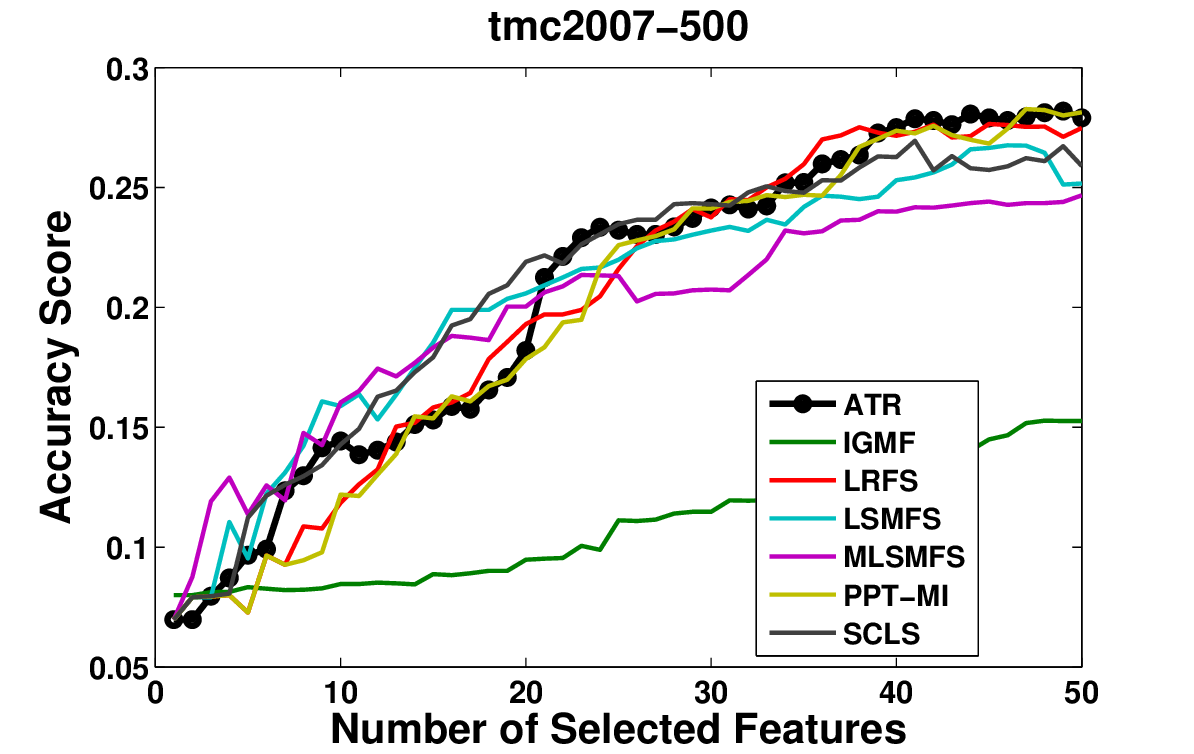}}
			\subfloat{\includegraphics[width=2.2in,height = 1.6in ]{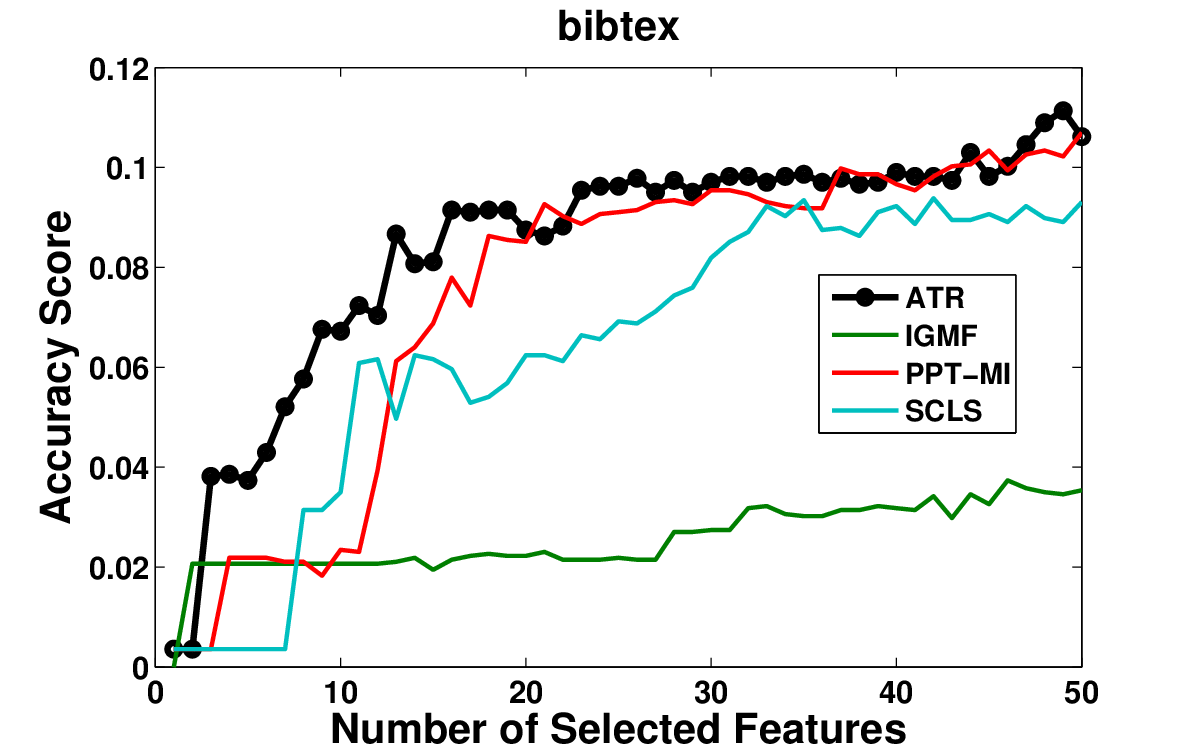}}\\
			\subfloat{\includegraphics[width=2.2in,height = 1.6in ]{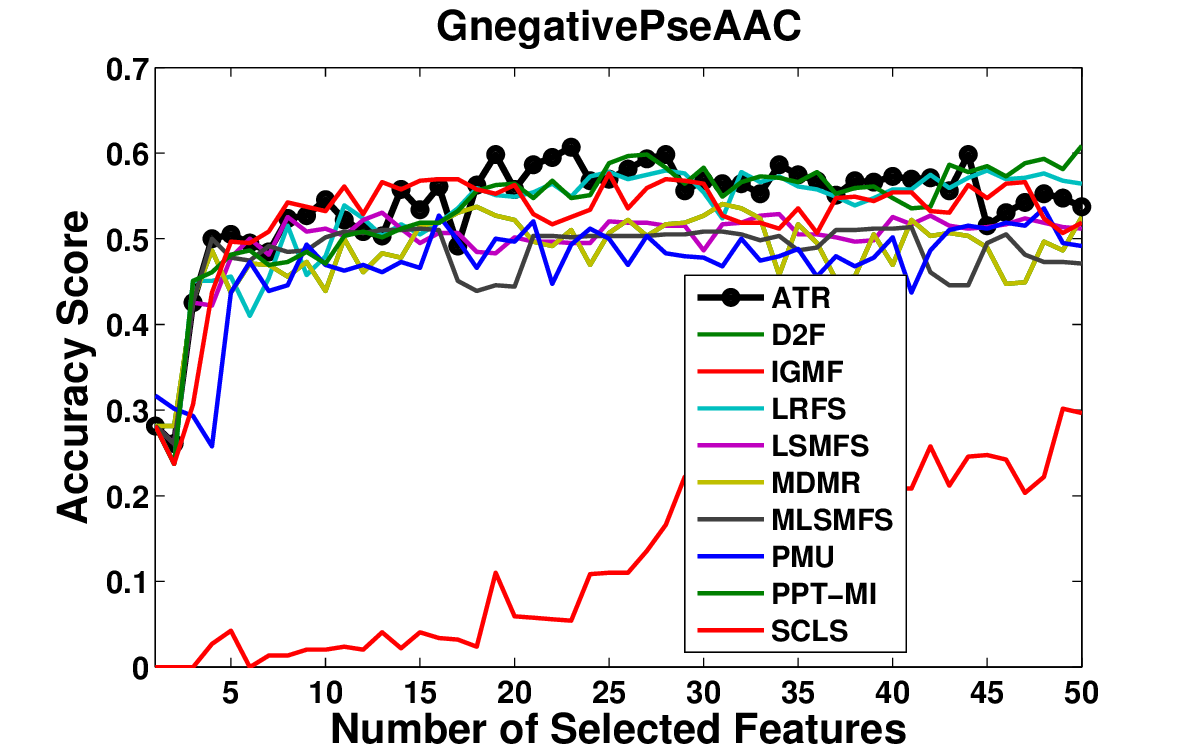}}
			\subfloat{\includegraphics[width=2.2in,height = 1.6in ]{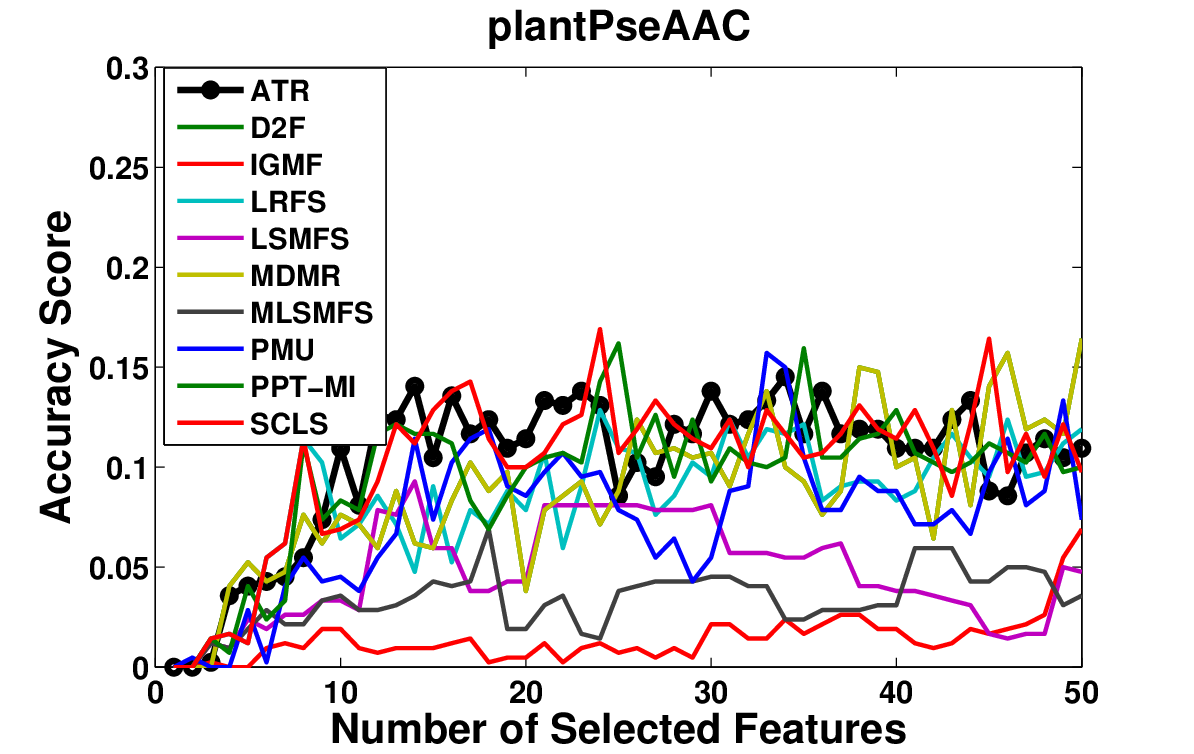}}			
			\subfloat{\includegraphics[width=2.2in,height = 1.6in ]{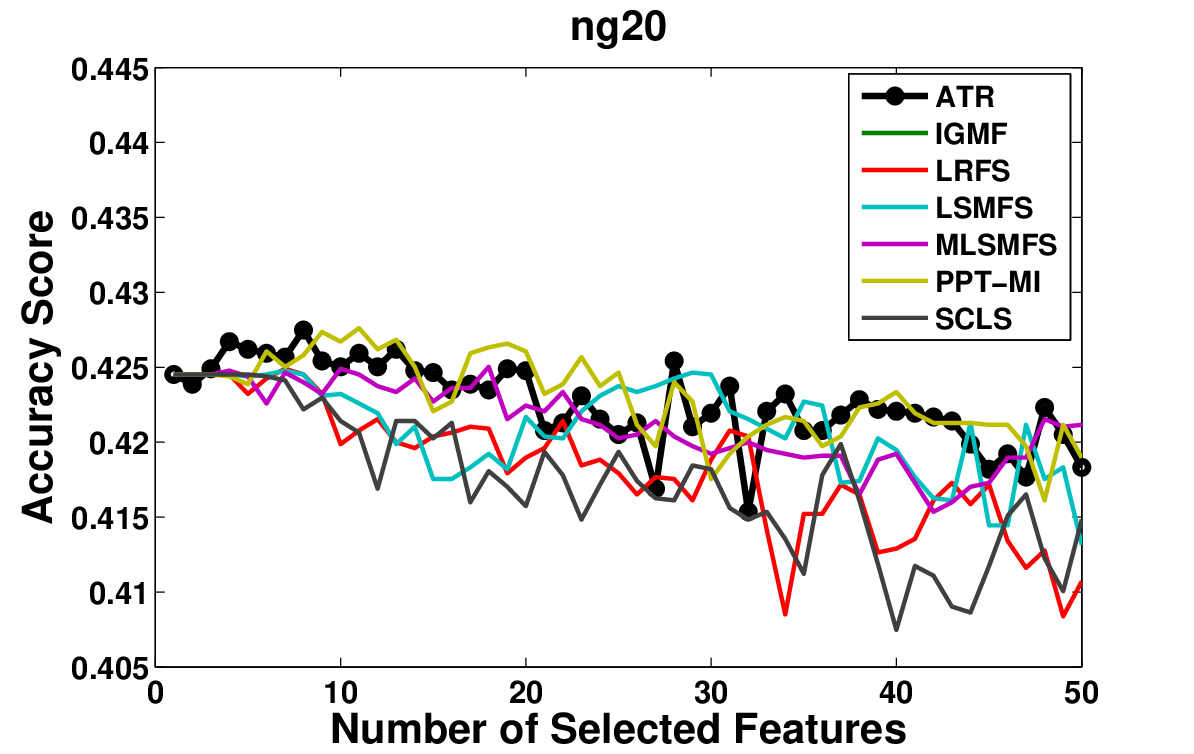}}
			
			\caption{Comparison of MLKNN Accuracy Scores for $N=1,2, \dots 50$ using the ten MLFS algorithms}
			\label{fig:mlknn_AS}
\end{figure*}

Tables \ref{table:accuracy_hamming-loss}- \ref{table:accuracy_accuracy-score} report the classification performance of the ten compared algorithms based on the six evaluation metrics. The results are averaged over the top 1 to top 50 ranked features for each algorithm. it is evident that ATR performs remarkably well, exhibiting an improvement in classification performance across most of the tests. 
It is essential to acknowledge the challenges faced by D2F, PMU, and MDMR when applied to the six datasets \textit{enron}, \textit{medical}, \textit{genbase}, \textit{tmc2007-500}, \textit{bibtex}, and \textit{ng20}. These challenges are mainly attributed to the term $|L||F|^3$ in their computational complexity (as shown in Table \ref{Table: MLFS-Algorithms}), which renders them inefficient for high-dimensional datasets with large values of $|F|$. Similarly, the LRFS, LSMFS, and MLSMFS encountered a similar challenge on the \textit{bibtex} dataset due to the term $|L|^2|F|^2$ in their computational complexity that makes them inefficient on datasets with large label space (large values of $|L|$). The following results can be derived
from these tables:

\begin{enumerate}
    \item In terms of Hamming Loss (Table \ref{table:accuracy_hamming-loss}), ATR achieves the lowest losses for seven datasets: \textit{emotions}, \textit{birds}, \textit{yeast}, \textit{genbase}, \textit{bibtex}, \textit{GnegativePseAAC}, and \textit{ng20}. Additionally, our proposed algorithm secures the second rank for three datasets: \textit{enron}, \textit{scene}, and \textit{tmc2007-500}. Comparing with PMU, D2F, and MDMR within the subset of six datasets for which they were able to find results, our algorithm consistently demonstrates superior performance in all cases. When compared with IGMF, our algorithm is inferior for only one dataset, \textit{GnegativePseAAC}. However, for all other cases, ATR shows a significant improvement over IGMF. In comparison to LRFS, our algorithm proves to be superior for all the cases. Among the existing algorithms, MLSMFS demonstrates the best performance and obtains the top results for three datasets: \textit{enron}, \textit{medical}, and \textit{plantPseAAC}. However, in the remaining cases, our algorithm exhibits superior performance. Commparing with LSMFS, PPT-MI, and SCLS, our algorithm is inferior in only two cases, while achieving victory in all the other cases. 

    \item In terms of Label Ranking Loss (Table \ref{table:accuracy_label-ranking-loss}), ATR secures the first rank for six datasets, including \textit{yeast}, \textit{genbase}, \textit{bibtex}, \textit{GnegativePseAAC}, \textit{plantPseAAC}, and \textit{ng20}, and achieves the second rank for \textit{emotions} and \textit{scene}. When compared with PMU, D2F, MDMR, and IGMF, ATR consistently demonstrates superior performance. Furthermore, ATR outperforms LRFS in ten cases. Among existing algorithms, MLSMFS again performs best, obtaining top results for \textit{enron}, \textit{medical}, and \textit{tmc2007-500}, while ATR excels in the remaining cases. SCLS shows top results for \textit{scene} and second rank for \textit{birds} and \textit{tmc2007-500}, but ATR outperforms it in all other cases. Comparing with LSMFS, ATR is inferior for only \textit{enron} and \textit{medical}, and in comparison with PPT-MI, ATR is inferior only for \textit{tmc2007-500}, achieving victory in all other cases. 
    
    \item Considering (Table \ref{table:accuracy_coverage-error})  ATR achieves the lowest Coverage Error for four datasets: \textit{genbase}, \textit{bibtex}, \textit{GnegativePseAAC}, and \textit{plantPseAAC}. It also secures the second rank for five datasets: \textit{emotions}, \textit{birds}, \textit{scene}, \textit{yeast}, and \textit{ng20}. comparing with PMU,  ATR demonstrates superior performance in all cases.  Comparing with D2F, MDMR, and IGMF, ATR is inferior for only one dataset.  LRFS demonstrates the best performance for two datasets: \textit{emotions } and  \textit{birds }. Yet, when it comes to the remaining cases, ATR continues to showcase its superiority. MLSMFS obtains the top results for three datasets: \textit{enron}, \textit{medical}, and \textit{ tmc2007-500 }. However, our algorithm exhibits superior performance in other cases.  Additionally, PT-MI and SCLS achieve top results for \textit{ng20} and \textit{scene}, respectively. Our algorithm is inferior in only one dataset, \textit{tmc2007-500}, in comparison with SCLS. However, for all other cases, ATR shows a significant improvement over PT-MI and SCLS. Comparing with LSMFS, our algorithm is inferior in only one case: \textit{enron}, while achieving victory in all the other cases.

    \item In terms of F1-Score (Table \ref{table:accuracy_f1-score}), ATR attains the highest score for five datasets including \textit{emotions}, \textit{yeast}, \textit{ GnegativePseAAC }, \textit{ plantPseAAC }, and \textit{ng20}. Moreover, our proposed algorithm obtains the second position for two datasets: \textit{ scene } and \textit{ bibtex }. When comparing with PMU, D2F, MDMR, IGMF, and PPT-MI, our algoirhm  exhibits superior performance in all cases.  MLSMFS obtains the top results for three datasets: \textit{enron}, \textit{medical}, and \textit{ genbase}. For the remaining cases, except \textit{tmc2007-500},  our algorithm showcases superior performance.  Also, Comparing  with SCLS, our algorithm is inferior in four cases: \textit{scene}, \textit{genbase},  \textit{tmc2007-500} and \textit{bibtex}, while achieving victory in all the other cases.  LRFS obtains a better score than ATR only  in \textit{ birds}. Finally, in comparison with LSMFS, our algorithm is inferior in two cases: \textit{enron} and \textit{medical}, and in all other cases, our algorithm emerges victorious.
    \item In terms of Jaccard Score (Table \ref{table:accuracy_jaccard-score}), ATR secures the highest score for four datasets, encompassing \textit{yeast}, \textit{ GnegativePseAAC}, \textit{ plantPseAAC }, and \textit{ ng20 }. Additionally, our proposed algorithm secures the second rank for three datasets: \textit{emotions}, \textit{scene}, and \textit{ bibtex}. Comparing with LRFS, our algorithm is inferior in two cases: \textit{ emotions } and \textit{birds}. For the other cases, our algorithm comes out on top. The resutls for PMU, D2F, MDMR, IGMF, PPT-MI, MLSMFS, SCLS, and  LSMFS are the same as F1-Score. 
    
    \item In terms of Accuracy Score (Table \ref{table:accuracy_accuracy-score}), ATR attains the highest score for five datasets including \textit{emotions}, \textit{ yeast},\textit{ bibtex}, \textit{GnegativePseAAC}, and\textit{ plantPseAAC }. Moreover, our proposed algorithm obtains the second position for three datasets: \textit{ scene }, \textit{tmc2007-500}, and \textit{ng20}. When comparing with D2F and MDMR, ATR is inferior for only one dataset, \textit{ birds} and in comparison with PMU, IGMF, and LRFS,  ATR demonstrates superior performance in all cases.  PPT-MI obtains higher score than ATR for \textit{birds} and \textit{ ng20} and SCLS achieves better scores for \textit{scene}, \textit{ genbase}, and \textit{ tmc2007-500}. However, for the remaining  cases, ATR is more accurate.   MLSMFS demonstrates the best performance for three datasets: \textit{enron}, \textit{medical}, and \textit{ genbase }. Yet, in the remaining datasets, our algorithm showcases superior performance. 
    
\end{enumerate}
Figures \ref{fig:mlknn_HL}- \ref{fig:mlknn_AS} offer a detailed understanding of how the ten compared algorithms perform in the classification process while incorporating the top 50 features into the chosen feature set, as evaluated by six metrics. Notably, ATR stands out with impressive performance, showcasing significant enhancements across the majority of the tests.

\begin{table}[H]

	\renewcommand{\arraystretch}{1.3}
	\caption{Comparison of MLKNN Hamming loss for the ten algorithms (the smaller is the better). The results are averaged over the top 1 to top 50 ranked features for each algorithm. The best algorithm for each dataset is indicated in bold, and the second best algorithm is shown in red color.  Moreover, '-' indicates that the algorithm couldn't find a result within 4 computational hours.  }
	\label{table:accuracy_hamming-loss}
	\centering
	\footnotesize
	\begin{adjustbox}{width=1.\textwidth}
		\small
		\begin{tabular}{ lcccccccccc }
			
			\textbf{Dataset} & \textbf{ATR} & 	\textbf{D2F}  & \textbf{IGMF} & \textbf{LRFS} & \textbf{LSMFS} & \textbf{MDMR} & 	\textbf{MLSMFS}  & \textbf{PMU} & \textbf{PPT\_MI} & \textbf{SCLS} \\
			\hline
			emotions & $ \mathbf{0.2419 \pm0.0151}$ &$0.2519 \pm0.0147$ &$0.2840 \pm0.0417$ &\textcolor{red}{$0.2439 \pm0.0143$} &$0.2599 \pm0.0091$ &$0.2519 \pm0.0151$ &$0.2540 \pm0.0112$ &$0.2782 \pm0.0200$ &$0.2465 \pm0.0178$ &$0.2454 \pm0.0112$\\
			birds & $ \mathbf{0.0508 \pm0.0015}$ &$0.0516 \pm0.0014$ &$0.0521 \pm0.0008$ &\textcolor{red}{$0.0508 \pm0.0008$ }&$0.0526 \pm0.0014$ &$0.0516 \pm0.0015$ &$0.0517 \pm0.0016$ &$0.0509 \pm0.0014$ &$0.0528 \pm0.0016$ &$0.0523 \pm0.0011$\\
			enron &\textcolor{red}{$0.0526 \pm0.0017$} & -- &$0.0585 \pm0.0011$ &$0.0586 \pm0.0006$ &$0.0529 \pm0.0030$ & -- & $ \mathbf{0.0515 \pm0.0023}$ & -- &$0.0542 \pm0.0014$ &$0.0535 \pm0.0038$\\
			medical &$0.0160 \pm0.0021$ & -- &$0.0276 \pm0.0000$ &$0.0168 \pm0.0017$ &\textcolor{red}{$0.0150 \pm0.0024$} & -- & $ \mathbf{0.0150 \pm0.0020}$ & -- &$0.0162 \pm0.0021$ &$0.0162 \pm0.0020$\\
			scene &\textcolor{red}{$0.1459 \pm0.0089$} &$0.1497 \pm0.0066$ &$0.1788 \pm0.0025$ &$0.1632 \pm0.0064$ &$0.1486 \pm0.0084$ &$0.1497 \pm0.0066$ &$0.1530 \pm0.0058$ &$0.1712 \pm0.0053$ &$0.1683 \pm0.0054$ & $ \mathbf{0.1335 \pm0.0127}$\\
			yeast & $ \mathbf{0.2204 \pm0.0045}$ &$0.2215 \pm0.0049$ &$0.2345 \pm0.0039$ &$0.2214 \pm0.0040$ &$0.2214 \pm0.0043$ &$0.2215 \pm0.0049$ &$0.2226 \pm0.0048$ &$0.2241 \pm0.0046$ &\textcolor{red}{$0.2208 \pm0.0054$} &$0.2225 \pm0.0049$\\	
			genbase & $ \mathbf{0.0069 \pm0.0080}$ & -- &$0.0456 \pm0.0000$ &$0.0155 \pm0.0077$ &$0.0159 \pm0.0085$ & -- &$0.0164 \pm0.0076$ & -- &$0.0071 \pm0.0082$ &\textcolor{red}{$0.0174 \pm0.0064$}\\
			tmc2007-500 &\textcolor{red}{$0.0725 \pm0.0102$} & -- &$0.0873 \pm0.0036$ &$0.0727 \pm0.0105$ &$0.0731 \pm0.0080$ & -- &$0.0737 \pm0.0071$ & -- &$0.0732 \pm0.0106$ & $ \mathbf{0.0718 \pm0.0092}$\\
			bibtex & $ \mathbf{0.0139 \pm0.0002}$ & -- &$0.0152 \pm0.0001$ & -- & -- & -- & -- & -- &\textcolor{red}{$0.0140 \pm0.0002$} &$0.0142 \pm0.0002$\\
			GnegativePseAAC & $ \mathbf{0.0823 \pm0.0051}$ &$0.0893 \pm0.0042$ &$0.1254 \pm0.0039$ &$0.0837 \pm0.0064$ &$0.0912 \pm0.0035$ &$0.0893 \pm0.0042$ &$0.0926 \pm0.0029$ &$0.0921 \pm0.0069$ &\textcolor{red}{$0.0832 \pm0.0058$} &$0.0846 \pm0.0044$\\
			plantPseAAC &$0.0940 \pm0.0028$ &$0.0946 \pm0.0029$ &\textcolor{red}{$0.0927 \pm0.0025$} &$0.0952 \pm0.0041$ &$0.0934 \pm0.0024$ &$0.0946 \pm0.0029$ & $ \mathbf{0.0920 \pm0.0011}$ &$0.0950 \pm0.0029$ &$0.0931 \pm0.0032$ &$0.0944 \pm0.0032$\\
			ng20 &$\mathbf{0.0524 \pm0.0003}$ & -- &$0.0644 \pm0.0014$ &$0.0530 \pm0.0004$ &$0.0527 \pm0.0003$ & -- &$0.0525 \pm0.0003$ & -- & $ \mathbf{0.0524 \pm0.0003}$ &$0.0532 \pm0.0005$\\
			\hline

		\end{tabular}
	\end{adjustbox}
\end{table}

\begin{table}[H]
	\renewcommand{\arraystretch}{1.3}
	\caption{Comparison of MLKNN Label Ranking Loss for the ten algorithms (the smaller is the better) }
	\label{table:accuracy_label-ranking-loss}
	\centering
	\footnotesize
	\begin{adjustbox}{width=1.\textwidth}
		\small
		\begin{tabular}{ lcccccccccc }
			
			\textbf{Dataset} & \textbf{ATR} & 	\textbf{D2F}  & \textbf{IGMF} & \textbf{LRFS} & \textbf{LSMFS} & \textbf{MDMR} & 	\textbf{MLSMFS}  & \textbf{PMU} & \textbf{PPT\_MI} & \textbf{SCLS} \\
			\hline
			emotions &\textcolor{red}{$0.5220 \pm0.0346$} &$0.5500 \pm0.0403$ &$0.6149 \pm0.1108$ & $ \mathbf{0.5199 \pm0.0392}$ &$0.5553 \pm0.0366$ &$0.5500 \pm0.0346$ &$0.5459 \pm0.0385$ &$0.6096 \pm0.0852$ &$0.5229 \pm0.0480$ &$0.5267 \pm0.0428$\\
			birds &$0.4778 \pm0.0195$ &$0.4770 \pm0.0139$ &$0.5069 \pm0.0134$ & $ \mathbf{0.4648 \pm0.0209}$ &$0.4794 \pm0.0218$ &$0.4770 \pm0.0195$ &$0.4879 \pm0.0173$ &$0.4958 \pm0.0161$ &$0.4877 \pm0.0154$ &\textcolor{red}{$0.4753 \pm0.0212$}\\
			enron &$0.6172 \pm0.0730$ & -- &$0.7678 \pm0.0647$ &$0.7537 \pm0.0153$ &\textcolor{red}{$0.6081 \pm0.0891$} & -- & $ \mathbf{0.5725 \pm0.0681}$ & -- &$0.6904 \pm0.0539$ &$0.6459 \pm0.1037$\\
			medical &$0.4589 \pm0.0926$ & -- &$1.0000 \pm0.0000$ &$0.4874 \pm0.0812$ &\textcolor{red}{$0.4189 \pm0.1059$} & -- & $ \mathbf{0.4055 \pm0.0930}$ & -- &$0.4672 \pm0.0964$ &$0.4747 \pm0.0870$\\
			scene &\textcolor{red}{$0.5696 \pm0.0894$} &$0.5722 \pm0.0808$ &$0.8436 \pm0.0585$ &$0.7375 \pm0.0812$ &$0.5791 \pm0.0854$ &$0.5722 \pm0.0808$ &$0.6141 \pm0.0696$ &$0.7133 \pm0.0854$ &$0.7852 \pm0.0861$ & $ \mathbf{0.5076 \pm0.1122}$\\
			yeast & $ \mathbf{0.5015 \pm0.0331}$ &$0.5060 \pm0.0302$ &$0.5504 \pm0.0374$ &$0.5078 \pm0.0270$ &$0.5061 \pm0.0366$ &$0.5060 \pm0.0302$ &$0.5068 \pm0.0296$ &$0.5085 \pm0.0372$ &\textcolor{red}{$0.5035 \pm0.0331$} &$0.5117 \pm0.0349$\\
			genbase & $ \mathbf{0.0944 \pm0.1602}$ & -- &$1.0000 \pm0.0000$ &$0.1115 \pm0.1732$ &$0.1595 \pm0.2311$ & -- &$0.1012 \pm0.1876$ & -- &\textcolor{red}{$0.0958 \pm0.1623$} &$0.0993 \pm0.1690$\\
			tmc2007-500 &$0.4833 \pm0.0888$ & -- &$0.6440 \pm0.0474$ &$0.4949 \pm0.1062$ &$0.4923 \pm0.0893$ & -- & $ \mathbf{0.4653 \pm0.0906}$ & -- &\textcolor{red}{$0.4951 \pm0.1020$} &\textcolor{red}{$0.4784 \pm0.0944$}\\
			bibtex & $ \mathbf{0.8473 \pm0.0381}$ & -- &$0.9603 \pm0.0149$ & -- & -- & -- & -- & -- &$0.8591 \pm0.0471$ &$0.8556 \pm0.0503$\\
			GnegativePseAAC & $ \mathbf{0.4486 \pm0.0696}$ &$0.5034 \pm0.0556$ &$0.8688 \pm0.1029$ &$0.4564 \pm0.0749$ &$0.4927 \pm0.0545$ &$0.5034 \pm0.0556$ &$0.5100 \pm0.0514$ &$0.5213 \pm0.0611$ &\textcolor{red}{$0.4504 \pm0.0761$} &$0.4588 \pm0.0753$\\
			plantPseAAC & $ \mathbf{0.8865 \pm0.0429}$ &$0.9056 \pm0.0402$ &$0.9853 \pm0.0127$ &$0.9056 \pm0.0378$ &$0.9498 \pm0.0263$ &$0.9056 \pm0.0402$ &$0.9645 \pm0.0153$ &$0.9203 \pm0.0383$ &\textcolor{red}{$0.8957 \pm0.0416$} &$0.8890 \pm0.0428$\\
			ng20 & $ \mathbf{0.4465 \pm0.0104}$ & -- &$0.5403 \pm0.0527$ &$0.4524 \pm0.0079$ &$0.4547 \pm0.0077$ & -- &$0.4492 \pm0.0082$ & -- &\textcolor{red}{$0.4469 \pm0.0097$} &$0.4512 \pm0.0097$\\
					
			\hline

		\end{tabular}
	\end{adjustbox}
\end{table}

\begin{table*}
	\renewcommand{\arraystretch}{1.3}
	\caption{Comparison of MLKNN Coverage Error for the ten algorithms (the smaller is the better) }
	\label{table:accuracy_coverage-error}
	\centering
	\footnotesize
	\begin{adjustbox}{width=1.\textwidth}
		\small
		\begin{tabular}{ lcccccccccc }
			
			\textbf{Dataset} & \textbf{ATR} & 	\textbf{D2F}  & \textbf{IGMF} & \textbf{LRFS} & \textbf{LSMFS} & \textbf{MDMR} & 	\textbf{MLSMFS}  & \textbf{PMU} & \textbf{PPT\_MI} & \textbf{SCLS} \\
			\hline
			emotions &\textcolor{red}{$4.7298 \pm0.1404$} &$4.8460 \pm0.1693$ &$5.0702 \pm0.3697$ & $ \mathbf{4.7227 \pm0.1682}$ &$4.8479 \pm0.1612$ &$4.8460 \pm0.1404$ &$4.8709 \pm0.1609$ &$5.0359 \pm0.2854$ &$4.7334 \pm0.2018$ &$4.7490 \pm0.1774$\\
			birds &\textcolor{red}{$9.6344 \pm0.1920$} &$9.6370 \pm0.1506$ &$9.8225 \pm0.1041$ & $ \mathbf{9.5949 \pm0.1704}$ &$9.6598 \pm0.2137$ &$9.6370 \pm0.1920$ &$9.7692 \pm0.1623$ &$9.8213 \pm0.1553$ &$9.7777 \pm0.1501$ &$9.6867 \pm0.2164$\\
			enron &$48.0848 \pm2.7240$ & -- &$51.3798 \pm1.1631$ &$52.0482 \pm0.2787$ &\textcolor{red}{$47.2323 \pm2.7319$} & -- & $ \mathbf{46.2827 \pm2.0768}$ & -- &$51.2714 \pm1.5373$ &$48.7331 \pm2.6974$\\
			medical &$0.0498 \pm0.0103$ & -- &\textcolor{red}{$0.0276 \pm0.0000$} &$0.0505 \pm0.0108$ &$0.0315 \pm0.0037$ & -- & $ \mathbf{0.0196 \pm0.0024}$ & -- &$0.0510 \pm0.0098$ &$0.0505 \pm0.0108$\\
			scene &\textcolor{red}{$3.9677 \pm0.4257$} &$3.9755 \pm0.3833$ &$5.2427 \pm0.2799$ &$4.7538 \pm0.3797$ &$4.0156 \pm0.4046$ &$3.9755 \pm0.3833$ &$4.1872 \pm0.3260$ &$4.6353 \pm0.3995$ &$4.9778 \pm0.4018$ & $ \mathbf{3.6759 \pm0.5313}$\\
			yeast &\textcolor{red}{$11.7049 \pm0.4356$} & $ \mathbf{11.6963 \pm0.4350}$ &$12.0899 \pm0.5879$ &$11.7296 \pm0.4176$ &$11.8083 \pm0.3960$ & $ \mathbf{11.6963 \pm0.4350}$ &$11.8425 \pm0.3994$ &$11.7509 \pm0.4662$ &$11.7157 \pm0.4307$ &$11.7828 \pm0.4297$\\
			genbase & $ \mathbf{4.1851 \pm3.9560}$ & -- &$27.0000 \pm0.0000$ &$4.5887 \pm4.3908$ &$5.7964 \pm5.9119$ & -- &$4.2618 \pm4.8021$ & -- &\textcolor{red}{$4.2194 \pm3.9987$} &$4.2749 \pm4.3026$\\
			tmc2007-500 &$15.7850 \pm1.8418$ & -- &$18.8855 \pm0.8871$ &$15.9755 \pm2.1417$ &$15.8772 \pm1.7724$ & -- & $ \mathbf{15.4484 \pm1.7500}$ & -- &$16.0186 \pm2.0598$ &\textcolor{red}{$15.5638 \pm1.8815$}\\
			bibtex & $ \mathbf{145.4108 \pm4.1489}$ & -- &$154.7215 \pm1.3006$ & -- & -- & -- & -- & -- &$146.9469 \pm5.4795$ &$147.9672 \pm5.3162$\\
			GnegativePseAAC & $ \mathbf{4.1890 \pm0.4775}$ &$4.5728 \pm0.3776$ &$7.0999 \pm0.7093$ &$4.2466 \pm0.5152$ &$4.4968 \pm0.3711$ &$4.5728 \pm0.3776$ &$4.6272 \pm0.3461$ &$4.6920 \pm0.4192$ &\textcolor{red}{$4.2061 \pm0.5215$} &$4.2617 \pm0.5144$\\
			plantPseAAC & $ \mathbf{10.8073 \pm0.4450}$ &$11.0012 \pm0.4349$ &$11.8410 \pm0.1364$ &$11.0048 \pm0.4001$ &$11.4718 \pm0.2783$ &$11.0012 \pm0.4349$ &$11.6234 \pm0.1642$ &$11.1543 \pm0.4061$ &$10.9081 \pm0.4327$ &\textcolor{red}{$10.8328 \pm0.4524$}\\
			ng20 &\textcolor{red}{$11.1944 \pm0.1128$} & -- &$12.4358 \pm0.7061$ &$11.2890 \pm0.0676$ &$11.2884 \pm0.0862$ & -- &$11.2299 \pm0.0885$ & -- & $ \mathbf{11.1940 \pm0.1038}$ &$11.2684 \pm0.0938$\\

			\hline

		\end{tabular}
	\end{adjustbox}
\end{table*}

\begin{table*}
	\renewcommand{\arraystretch}{1.3}
	\caption{Comparison of MLKNN F1-Score for the ten algorithms (the larger is the better) }
	\label{table:accuracy_f1-score}
	\centering
	\footnotesize
	\begin{adjustbox}{width=1.\textwidth}
		\small
		\begin{tabular}{ lcccccccccc }
			
			\textbf{Dataset} & \textbf{ATR} & 	\textbf{D2F}  & \textbf{IGMF} & \textbf{LRFS} & \textbf{LSMFS} & \textbf{MDMR} & 	\textbf{MLSMFS}  & \textbf{PMU} & \textbf{PPT\_MI} & \textbf{SCLS} \\
			\hline
			emotions & $ \mathbf{0.5754 \pm0.0565}$ &$0.5523 \pm0.0517$ &$0.4442 \pm0.1460$ &\textcolor{red}{$0.5743 \pm0.0543$} &$0.5393 \pm0.0465$ &$0.5523 \pm0.0517$ &$0.5477 \pm0.0496$ &$0.4878 \pm0.1010$ &$0.5702 \pm0.0601$ &$0.5701 \pm0.0530$\\	
			birds &$0.1486 \pm0.0355$ &$0.1501 \pm0.0454$ &$0.0604 \pm0.0333$ & $ \mathbf{0.1880 \pm0.0529}$ &$0.1471 \pm0.0562$ &$0.1501 \pm0.0454$ &$0.1393 \pm0.0487$ &$0.1141 \pm0.0371$ &$0.1268 \pm0.0406$ &\textcolor{red}{$0.1553 \pm0.0462$}\\
			enron &$0.3760 \pm0.0552$ & -- &$0.2768 \pm0.0541$ &$0.2660 \pm0.0219$ &\textcolor{red}{$0.3890 \pm0.0837$} & -- & $ \mathbf{0.4150 \pm0.0651}$ & -- &$0.3291 \pm0.0416$ &$0.3760 \pm0.0918$\\
			medical &$0.5537 \pm0.1017$ & -- &$0.0000 \pm0.0000$ &$0.5278 \pm0.0913$ &\textcolor{red}{$0.5874 \pm0.1120$} & -- & $ \mathbf{0.5884 \pm0.0971}$ & -- &$0.5463 \pm0.1055$ &$0.5405 \pm0.0974$\\
			scene &\textcolor{red}{$0.4920 \pm0.0938$} &$0.4889 \pm0.0856$ &$0.1943 \pm0.0678$ &$0.3193 \pm0.0942$ &$0.4880 \pm0.0912$ &$0.4889 \pm0.0856$ &$0.4572 \pm0.0770$ &$0.3549 \pm0.0964$ &$0.2648 \pm0.0984$ & $ \mathbf{0.5481 \pm0.1175}$\\
			yeast & $ \mathbf{0.5333 \pm0.0440}$ &$0.5286 \pm0.0428$ &$0.4757 \pm0.0510$ &\textcolor{red}{$0.5299 \pm0.0401$} &$0.5289 \pm0.0492$ &$0.5286 \pm0.0428$ &$0.5248 \pm0.0407$ &$0.5268 \pm0.0465$ &$0.5295 \pm0.0435$ &$0.5286 \pm0.0501$\\
			genbase &$0.8673 \pm0.1762$ & -- &$0.0000 \pm0.0000$ &$0.8703 \pm0.1804$ &$0.8420 \pm0.2092$ & -- & $ \mathbf{0.8863 \pm0.1822}$ & -- &$0.8646 \pm0.1775$ &\textcolor{red}{$0.8814 \pm0.1762$}\\
			tmc2007-500 &$0.5032 \pm0.1193$ & -- &$0.3134 \pm0.0615$ &$0.4919 \pm0.1328$ &$0.5067 \pm0.1123$ & -- &\textcolor{red}{$0.5166 \pm0.1054$} & -- &$0.4886 \pm0.1318$ & $ \mathbf{0.5184 \pm0.1226}$\\
			bibtex &\textcolor{red}{$0.1292 \pm0.0307$} & -- &$0.0387 \pm0.0168$ & -- & -- & -- & -- & -- &$0.1216 \pm0.0328$ & $ \mathbf{0.1434 \pm0.0459}$\\
			GnegativePseAAC & $ \mathbf{0.5848 \pm0.0577}$ &$0.5369 \pm0.0435$ &$0.1745 \pm0.1237$ &$0.5726 \pm0.0656$ &$0.5422 \pm0.0443$ &$0.5369 \pm0.0435$ &$0.5304 \pm0.0406$ &$0.5146 \pm0.0713$ &\textcolor{red}{$0.5783 \pm0.0657$} &$0.5769 \pm0.0668$\\
			plantPseAAC & $ \mathbf{0.1500 \pm0.0539}$ &$0.1284 \pm0.0449$ &$0.0239 \pm0.0181$ &$0.1286 \pm0.0465$ &$0.0734 \pm0.0333$ &$0.1284 \pm0.0449$ &$0.0580 \pm0.0234$ &$0.1105 \pm0.0476$ &$0.1408 \pm0.0513$ &\textcolor{red}{$0.1435 \pm0.0496$}\\
			ng20 & $ \mathbf{0.3735 \pm0.0300}$ & -- &$0.2435 \pm0.0186$ &$0.3619 \pm0.0274$ &$0.3635 \pm0.0256$ & -- &$0.3678 \pm0.0223$ & -- &\textcolor{red}{$0.3683 \pm0.0282$ }&$0.3681 \pm0.0322$\\

			\hline

		\end{tabular}
	\end{adjustbox}
\end{table*}
 
\begin{table*}
	\renewcommand{\arraystretch}{1.3}
	\caption{Comparison of MLKNN Jaccard Score for the ten algorithms (the larger is the better) }
	\label{table:accuracy_jaccard-score}

	\footnotesize
	\begin{adjustbox}{width=1.\textwidth}
		\small
		\begin{tabular}{ lcccccccccc }
			
			\textbf{Dataset} & \textbf{ATR} & 	\textbf{D2F}  & \textbf{IGMF} & \textbf{LRFS} & \textbf{LSMFS} & \textbf{MDMR} & 	\textbf{MLSMFS}  & \textbf{PMU} & \textbf{PPT\_MI} & \textbf{SCLS} \\
			\hline
			emotions &\textcolor{red}{$0.4144 \pm0.0437$} &$0.3897 \pm0.0418$ &$0.3088 \pm0.1130$ & $ \mathbf{0.4154 \pm0.0421}$ &$0.3839 \pm0.0348$ &$0.3897 \pm0.0418$ &$0.3904 \pm0.0388$ &$0.3322 \pm0.0763$ &$0.4103 \pm0.0500$ &$0.4112 \pm0.0411$\\		
			emotions &$0.0904 \pm0.0219$ &$0.0902 \pm0.0291$ &$0.0348 \pm0.0206$ & $ \mathbf{0.1161 \pm0.0335}$ &$0.0883 \pm0.0346$ &$0.0902 \pm0.0291$ &$0.0839 \pm0.0297$ &$0.0670 \pm0.0225$ &$0.0742 \pm0.0245$ &\textcolor{red}{$0.0942 \pm0.0297$}\\
			enron &$0.2822 \pm0.0416$ & -- &$0.1880 \pm0.0412$ &$0.1875 \pm0.0144$ &\textcolor{red}{$0.2841 \pm0.0631$} & -- & $ \mathbf{0.3069 \pm0.0495}$ & -- &$0.2470 \pm0.0338$ &$0.2742 \pm0.0740$\\
			medical &$0.4636 \pm0.0808$ & -- &$0.0000 \pm0.0000$ &$0.4385 \pm0.0701$ &\textcolor{red}{$0.4957 \pm0.0914$} & -- & $ \mathbf{0.4992 \pm0.0780}$ & -- &$0.4575 \pm0.0840$ &$0.4521 \pm0.0769$\\
			scene &\textcolor{red}{$0.3463 \pm0.0704$} &$0.3371 \pm0.0611$ &$0.1212 \pm0.0433$ &$0.2128 \pm0.0670$ &$0.3362 \pm0.0662$ &$0.3371 \pm0.0611$ &$0.3117 \pm0.0553$ &$0.2229 \pm0.0643$ &$0.1728 \pm0.0681$ & $ \mathbf{0.3965 \pm0.0900}$\\
			yeast & $ \mathbf{0.4103 \pm0.0294}$ &$0.4056 \pm0.0279$ &$0.3679 \pm0.0322$ &$0.4060 \pm0.0256$ &$0.4061 \pm0.0322$ &$0.4056 \pm0.0279$ &$0.4042 \pm0.0270$ &$0.4048 \pm0.0318$ &\textcolor{red}{$0.4073 \pm0.0291$} &$0.4047 \pm0.0326$ \\	
			genbase &$0.8591 \pm0.1764$ & -- &$0.0000 \pm0.0000$ &$0.8654 \pm0.1795$ &$0.8359 \pm0.2101$ & -- & $ \mathbf{0.8853 \pm0.1832}$ & -- &$0.8561 \pm0.1788$ &\textcolor{red}{$0.8762 \pm0.1748$}\\
			tmc2007-500 &$0.3792 \pm0.0923$ & -- &$0.2273 \pm0.0418$ &$0.3715 \pm0.1029$ &$0.3768 \pm0.0851$ & -- &\textcolor{red}{$0.3837 \pm0.0801$} & -- &$0.3689 \pm0.1011$ & $ \mathbf{0.3886 \pm0.0934}$\\
			bibtex &\textcolor{red}{$0.1139 \pm0.0234$} & -- &$0.0249 \pm0.0113$ & -- & -- & -- & -- & -- &$0.1076 \pm0.0264$ & $ \mathbf{0.1159 \pm0.0317}$\\
			GnegativePseAAC & $ \mathbf{0.4656 \pm0.0532}$ &$0.4174 \pm0.0395$ &$0.1083 \pm0.0812$ &$0.4564 \pm0.0595$ &$0.4203 \pm0.0387$ &$0.4174 \pm0.0395$ &$0.4013 \pm0.0335$ &$0.3968 \pm0.0585$ &\textcolor{red}{$0.4609 \pm0.0583$} &$0.4554 \pm0.0558$\\
			plantPseAAC & $ \mathbf{0.0871 \pm0.0321}$ &$0.0730 \pm0.0268$ &$0.0125 \pm0.0099$ &$0.0733 \pm0.0275$ &$0.0401 \pm0.0189$ &$0.0730 \pm0.0268$ &$0.0309 \pm0.0128$ &$0.0621 \pm0.0277$ &$0.0817 \pm0.0306$ &\textcolor{red}{$0.0837 \pm0.0299$}\\
			ng20 & $ \mathbf{0.3025 \pm0.0172}$ & -- &$0.1701 \pm0.0167$ &$0.2940 \pm0.0153$ &$0.2951 \pm0.0146$ & -- &$0.2986 \pm0.0129$ & -- &\textcolor{red}{$0.2998 \pm0.0161$} &$0.2978 \pm0.0184$\\		
			\hline

		\end{tabular}
	\end{adjustbox}
\end{table*}

\begin{table*}
	\renewcommand{\arraystretch}{1.3}
	\caption{Comparison of MLKNN Accuracy Score for the ten algorithms (the larger is the better  }
	\label{table:accuracy_accuracy-score}
	\centering
	\footnotesize
	\begin{adjustbox}{width=1.\textwidth}
		\small
		\begin{tabular}{ lcccccccccc }
			
			\textbf{Dataset} & \textbf{ATR} & 	\textbf{D2F}  & \textbf{IGMF} & \textbf{LRFS} & \textbf{LSMFS} & \textbf{MDMR} & 	\textbf{MLSMFS}  & \textbf{PMU} & \textbf{PPT\_MI} & \textbf{SCLS} \\
			\hline
			emotions & $ \mathbf{0.2245 \pm0.0337}$ &$0.2032 \pm0.0270$ &$0.1605 \pm0.0633$ &\textcolor{red}{$0.2206 \pm0.0298$} &$0.1738 \pm0.0264$ &$0.2032 \pm0.0270$ &$0.1767 \pm0.0280$ &$0.1568 \pm0.0430$ &$0.2179 \pm0.0356$ &$0.2077 \pm0.0285$\\	
			emotions &$0.4664 \pm0.0127$ & $ \mathbf{0.4741 \pm0.0070}$ &$0.4624 \pm0.0056$ &$0.4623 \pm0.0121$ &$0.4546 \pm0.0128$ & $ \mathbf{0.4741 \pm0.0070}$ &$0.4614 \pm0.0155$ &$0.4631 \pm0.0158$ &\textcolor{red}{$0.4466 \pm0.0179$} &$0.4599 \pm0.0149$\\
			enron &$0.0808 \pm0.0474$ & -- &$0.0237 \pm0.0191$ &$0.0146 \pm0.0031$ &\textcolor{red}{$0.0910 \pm0.0424$} & -- & $ \mathbf{0.1060 \pm0.0298}$ & -- &$0.0228 \pm0.0251$ &$0.0627 \pm0.0412$\\
			medical &$0.4675 \pm0.0830$ & -- &$0.0000 \pm0.0000$ &$0.4401 \pm0.0711$ &\textcolor{red}{$0.5045 \pm0.0963$} & -- & $ \mathbf{0.5115 \pm0.0825}$ & -- &$0.4585 \pm0.0850$ &$0.4528 \pm0.0767$\\
			scene &\textcolor{red}{$0.3894 \pm0.0780$} &$0.3863 \pm0.0693$ &$0.1480 \pm0.0527$ &$0.2395 \pm0.0703$ &$0.3816 \pm0.0744$ &$0.3863 \pm0.0693$ &$0.3491 \pm0.0593$ &$0.2618 \pm0.0736$ &$0.2010 \pm0.0779$ & $ \mathbf{0.4416 \pm0.0968}$\\
			yeast & $ \mathbf{0.1367 \pm0.0314}$ &$0.1307 \pm0.0266$ &$0.0948 \pm0.0311$ &$0.1331 \pm0.0235$ &$0.1259 \pm0.0258$ &$0.1307 \pm0.0266$ &$0.1211 \pm0.0223$ &$0.1267 \pm0.0313$ &\textcolor{red}{$0.1350 \pm0.0294$} &$0.1335 \pm0.0296$\\
			genbase &$0.8681 \pm0.1505$ & -- &$0.0000 \pm0.0000$ &$0.8638 \pm0.1694$ &$0.8214 \pm0.2286$ & -- & $ \mathbf{0.8806 \pm0.1863}$ & -- &$0.8650 \pm0.1513$ &\textcolor{red}{$0.8762 \pm0.1663$}\\
			tmc2007-500 &\textcolor{red}{$0.2052 \pm0.0677$} & -- &$0.1090 \pm0.0240$ &$0.2010 \pm0.0704$ &$0.2047 \pm0.0548$ & -- &$0.1971 \pm0.0454$ & -- &$0.1982 \pm0.0709$ & $ \mathbf{0.2080 \pm0.0595}$\\
			bibtex & $ \mathbf{0.0840 \pm0.0250}$ & -- &$0.0257 \pm0.0067$ & -- & -- & -- & -- & -- &\textcolor{red}{$0.0742 \pm0.0332$} &$0.0640 \pm0.0297$\\
			GnegativePseAAC & $ \mathbf{0.5391 \pm0.0656}$ &$0.4848 \pm0.0512$ &$0.1273 \pm0.0994$ &$0.5282 \pm0.0695$ &$0.4958 \pm0.0510$ &$0.4848 \pm0.0512$ &$0.4813 \pm0.0492$ &$0.4689 \pm0.0579$ &\textcolor{red}{$0.5338 \pm0.0700$} &$0.5233 \pm0.0688$\\
			plantPseAAC & $ \mathbf{0.1020 \pm0.0373}$ &$0.0883 \pm0.0378$ &$0.0142 \pm0.0123$ &$0.0845 \pm0.0324$ &$0.0473 \pm0.0251$ &$0.0883 \pm0.0378$ &$0.0338 \pm0.0146$ &$0.0748 \pm0.0365$ &$0.0948 \pm0.0366$ &\textcolor{red}{$0.1012 \pm0.0382$}\\
			ng20 &\textcolor{red}{$0.4227 \pm0.0027$} & -- &$0.3024 \pm0.0259$ &$0.4183 \pm0.0043$ &$0.4209 \pm0.0031$ & -- &$0.4212 \pm0.0027$ & -- & $ \mathbf{0.4230 \pm0.0027}$ &$0.4173 \pm0.0046$\\

			\hline

		\end{tabular}
	\end{adjustbox}
\end{table*}



\subsection{Running Times}
Table \ref{table:running-times} presents the running times (in seconds) for ranking the whole feature space of each dataset using different MLFS algorithms. For each algorithm and dataset, the reported time is the mean of three runs to ensure accurate and reliable measurements.
It is important to note that a limit of four hours (14400 seconds) is set for each run. If an algorithm couldn't find a result within this time limit, it is indicated by a '-'. 

As it can be seen, PPT-MI is the fastest method among the compared  algorithms. It achieves this speed advantage by evaluating only a single mutual information value for each feature. IGMF calculates two entropies of order $|L|$ for each feature. For datasets with small label spaces, IGMF runs highly efficient. However, it encounters challenges when applied to datasets with large label spaces, such as \textit{tmc2007-500} and \textit{bibtex}.  Both SCLS and ATR showcase scalability for datasets with large feature and label spaces, thanks to their low computational complexity.  The marginal distinction between them lies in the PPT transform used in ATR and the calculation of mutual information after transformation. LRFS, LSMFS, and MLSMFS utilize first-order redundancy calculations and second-order relevance evaluations, making them computationally inefficient for datasets with large label spaces, such as \textit{bibtex}, \textit{ng20}, \textit{enron}, and \textit{medical}. Conversly, PMU, D2F, and MDMR utilize second-order redundancy calculations and first-order relevance evaluations, resulting in computational inefficiency for datasets with large feature spaces, encompassing almost all the datasets used in the experiments.

\begin{table*}
    
	\renewcommand{\arraystretch}{1.3}
	    \caption{Running times of the comparing algorithms (in seconds). To avoiding redundant computations, The 'pre-eval' option in the PyIT-MLFS library was adopted to calculate all the pairwise mutual informations in advance.}

	\label{table:running-times}
	\centering
	\footnotesize
	\begin{adjustbox}{width=1.\textwidth}
		\small
		\begin{tabular}{ lcccccccccc }
			
			\textbf{Dataset} & \textbf{ATR} & 	\textbf{D2F}  & \textbf{IGMF} & \textbf{LRFS} & \textbf{LSMFS} & \textbf{MDMR} & 	\textbf{MLSMFS}  & \textbf{PMU} & \textbf{PPT\_MI} & \textbf{SCLS} \\
			\hline
			emotions & $ {8}$ & 
               $ {107}$ & 
               $ {4}$   & 
               $ {14}$  & 
               $ {17}$  & 
               $ {107 }$& 
               $ {17}$  & 
               $ {115}$ &
               $ {0}$   &  
               $ {8}$   \\
                birds &$ {115}$&
                $ {4279}$&
                $ {13}$ & 
                $ {344}$& 
                $ {430}$ & $ {4253}$ & $ {430}$ & $ {4577}$& 
                $ {0}$ & $ {115}$  \\   
			enron & $ {2149}$&
                  -- &   
                  $ {212}$ &
                  $ {12998}$ & 
                  $ {12212}$ & 
                  -- & 
                  $ {12202}$ &
                  -- &
                  $ {1}$ &
                  $ {2291}$ \\
                      
		    medical &$ {3905}$ &
                           --&
                        $ {90}$&
                        --&
                        --&
                           --&
                        --&
                           --&
                        $ {2}$&
                        $ {3869}$ \\
   
			scene &$ {172}$ & 
                       $ {2215}$ &
                       $ {57}$&
                       $ {202}$&
                       $ {215}$&
                       $ {2217}$&
                       $ {214}$&
                       $ {2273}$&
                       $ {0}$&
                       $ {170}$
                       \\
			yeast&$ {24}$ & 
                       $ {701}$ &
                       $ {28}$&
                       $ {106}$&
                       $ {139}$&
                       $ {721}$&
                       $ {107}$&
                       $ {818}$&
                       $ {0}$&
                       $ {29}$ \\
			genbase&$ {3294}$ & 
                       -- &
                       $ {93}$&
                       $ {5200}$&
                       $ {5791}$&
                       --&
                       $ {6194}$&
                       --&
                       $ {1}$&
                       $ {3472}$ \\
			tmc2007-500&$ {2799}$ & 
                       -- &
                       $ {12971}$&
                       $ {7364}$&
                       $ {8823}$&
                       --&
                       $ {8831}$&
                       --&
                       $ {16}$&
                       $ {2861}$ \\
			bibtex&$ {14331}$ & 
                       -- &
                       --&
                       -- &
                       -- &
                       -- &
                       -- &
                       -- &
                       $ {5}$&
                       $ {14343}$\\
			GnegativePseAAC&$ {363}$ & 
                       $ {6336}$ &
                       $ {63}$&
                       $ {445}$&
                       $ {477}$&
                       $ {6055}$&
                       $ {475}$&
                       $ {6365}$&
                       $ {0}$&
                       $ {362}$ \\
			plantPseAAC&$ {359}$ & 
                       $ {8741}$ &
                       $ {47}$&
                       $ {521}$&
                       $ {581}$&
                       $ {8594}$&
                       $ {580}$&
                       --&
                       $ {0}$&
                       $ {354}$ \\
			ng20&$ {7049}$ & 
                       -- &
                       $ {2389}$&
                       $ {11819}$&
                       $ {13424}$&
                       --&
                       $ {13399}$&
                       --&
                       $ {8}$&
                       $ {7085}$ \\

			\hline

		\end{tabular}
	\end{adjustbox}
\end{table*}

\section{Conclusions}
Multi-label learning has emerged as a fundamental paradigm, tackling situations where instances are associated with multiple class labels simultaneously. MLFS is the task of selecting most important and discriminative features in multi-label data with a vast feature space encompassing a multitude of irrelevant or redundant features. In this paper, we proposed an information-theoretical filter-based MLFS method, called ATR.  By synergizing algorithm adaptation and problem transformation strategies, ATR effectively ranks features, taking into consideration individual label influences as well as the discriminative potentials within the abstract label space.

We conducted a comprehensive series of experiments to evaluate the accuracy and efficiency of ATR. Our assessment involved a comparative analysis against ten existing  information-theoretic filter-based MLFS algorithms, using twelve diverse real-world datasets spanning various domains. The results demonstrate ATR's superiority across six  evaluation metrics, Hamming Loss, Label Ranking Loss, Coverage Error, F1 Score, Jaccard Score, and Accuracy Score of the MLKNN classifier. Notably, ATR exhibits computational efficiency comparable to that of SCLS, aligning it effectively for handling large-scale problems.

The authors would like to propose the following subjects for
future works:
\begin{itemize}
    \item \textbf{Online Streaming MLFS}: In an online streaming scenario, the feature/label space expands continuously while the number of instances remains constant. This situation arises primarily due to the potential unknown nature of the feature/label space. For instance, certain clinical tests may take longer to yield results than others, necessitating the gradual emergence of features/labels relevant to such tasks over time.

     \item \textbf{ATR with Different Levels of Abstraction}: An interesting direction to extend ATR is to selectively applying the PPT method to different subsets of labels, allowing for varying levels of abstraction. This would enable the algorithm to capitalize on specific label correlations and optimize feature relevance estimation for different subsets of labels. 
\end{itemize}

\section*{Declaration of generative AI and AI-assisted technologies in the writing process}

During the preparation of this work the authors used ChatGPT in order to improve language and readability. After using this tool, the authors reviewed and edited the content as needed and take full responsibility for the content of the publication.

\section*{Declaration of Competing Interest}

The authors declare that they have no known competing financial interests or personal relationships that could have appeared to influence the work reported in this paper.

 \bibliographystyle{elsarticle-num} 
 \bibliography{cas-refs}

\end{document}